\documentclass[10pt,twocolumn,letterpaper]{article}

\usepackage[pagenumbers]{cvpr} 

\usepackage{dsfont}

\definecolor{cvprblue}{rgb}{0.21,0.49,0.74}
\definecolor{tabfirst}{rgb}{0.4, 0.65, 0.3} 
\definecolor{tabsecond}{rgb}{0.7, 0.8, 0.65} 

\usepackage{bm}
\usepackage[table]{xcolor}
\usepackage[pagebackref,breaklinks,colorlinks,allcolors=cvprblue]{hyperref}

\def\paperID{*****} 
\def\confName{CVPR}
\def\confYear{2026}

\title{Global Structure-from-Motion Meets Feedforward Reconstruction}

\author{Linfei Pan\\
ETH Zurich\\
\and
Johannes Schönberger\\
Meta Reality Labs\\
\and
Marc Pollefeys\\
ETH Zurich, Microsoft\\
}

\begin{document}
\maketitle
\begin{abstract}
Structure-from-Motion -- the process of simultaneously estimating camera poses and 3D scene structure from a collection of images -- remains a central challenge in computer vision, with many open problems yet to be solved.
Recent advances in feedforward 3D reconstruction have made significant strides in overcoming persistent failure cases of classical SfM methods, particularly in scenarios characterized by low texture, limited overlap, and symmetries.
However, while feedforward approaches excel in these challenging conditions, they often face limitations regarding scalability, accuracy, or robustness, and typically fall short of classical methods in standard reconstruction settings.
In this work, we systematically analyze these limitations and propose a new Structure-from-Motion pipeline by combining the respective strengths of classical and feedforward methods.
Extensive experiments across multiple datasets show the benefits of our approach, achieving state-of-the-art results across a wide range of scenarios.
We share our system as an open-source implementation at \url{https://github.com/colmap/gluemap}.
\end{abstract}    

\section{Introduction}
\label{sec:intro}


Structure-from-Motion (SfM) tackles the problem of reconstructing 3D scene structure and cameras given a set of images.
It is a fundamental technique in computer vision and serves as a critical building block for numerous applications like localization~\cite{sattler2011fast}, multi-view stereo~\cite{schoenberger2016mvs}, novel-view-synthesis~\cite{kerbl2023gaussian}, or 3D training data generation~\cite{wang2025vggt}.

Throughout its long history~\cite{ullman1979interpretation}, progress in the field has been primarily driven by optimization-based algorithms~\cite{wu2013towards, snavely2006photo, moulon2016openmvg, schonberger2016structure, pan2024global, liu2024hybrid}, which we refer to as \textit{classical} methods in the remainder of this text.
While these approaches differ in their specific formulations, they generally share a common structure: correspondence search using pairwise (local) feature matching and (global or incremental) reconstruction using robust optimization.
To this day, SIFT~\cite{lowe2004distinctive} remains the standard choice for correspondence search.
The global reconstruction paradigm~\cite{pan2024global} sets itself apart from the incremental~\cite{schonberger2016structure} one by estimating camera poses and scene structure in a single, unified global optimization step, rather than incrementally building up a partial reconstruction using repeated local and global optimizations.
The global approach generally results in improved runtime scalability and better robustness against symmetry issues.
The state-of-the-art classical SfM systems~\cite{schonberger2016structure, pan2024global} achieve high levels of reliability~\cite{pataki2025mp} for sufficiently overlapping input images with enough parallax and discriminative scene texture.
In contrast, these methods frequently struggle when the input images are sparse, have limited parallax, or do not contain enough distinctive features to match.

\begin{figure}[t]
\centering
\includegraphics[width=0.95\columnwidth]{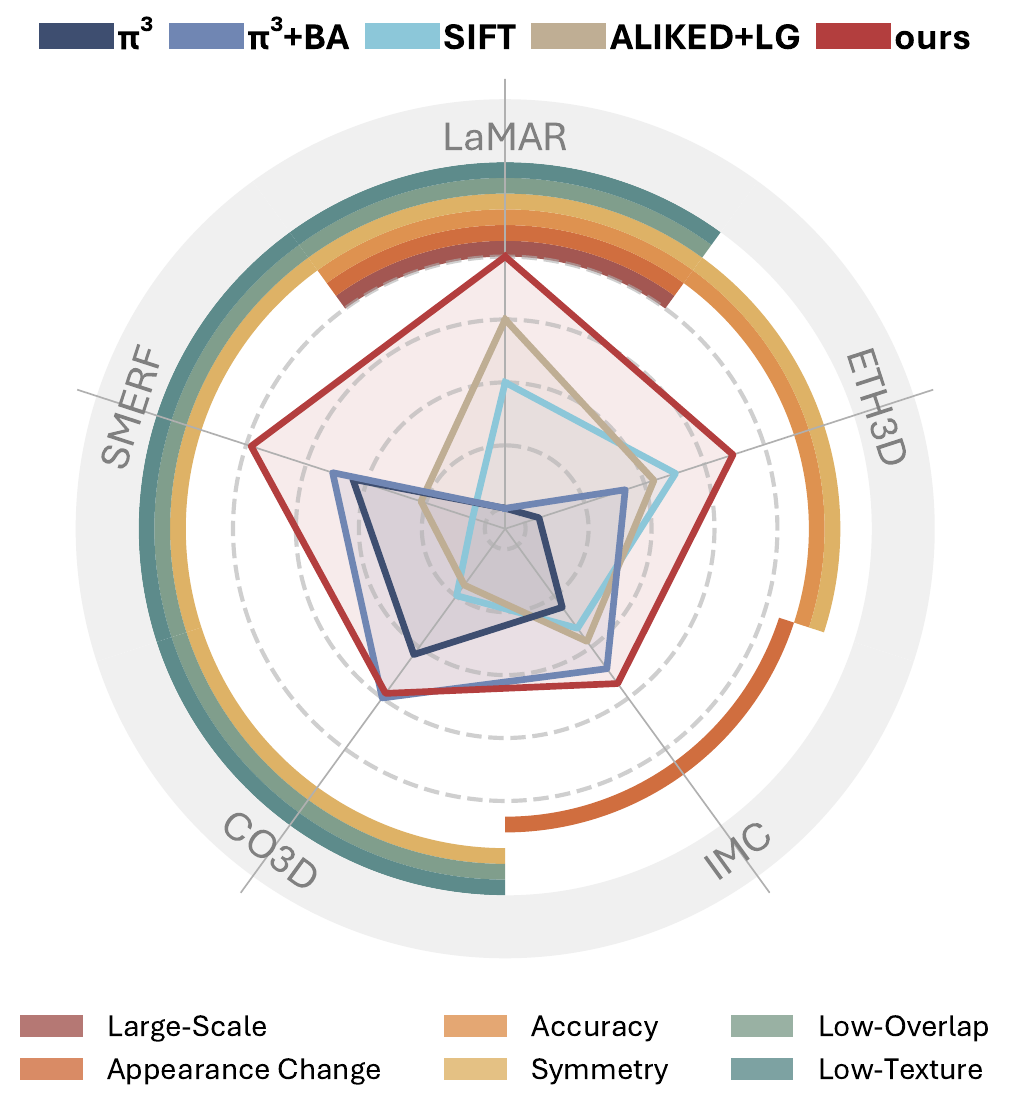}
\caption{We evaluate on 5 datasets, featuring different challenges. The average rank on each dataset is reported. Classical and feedforward methods struggle under different settings, while ours consistently maintains best performance across the board.}
\label{fig:teaser}
\end{figure}

To address the limitations of classical approaches, the research community has increasingly focused on replacing hand-crafted modules in the 3D reconstruction pipeline with learned feedforward components, while still leveraging classical optimization techniques.
For example, learned features like ALIKED~\cite{zhao2023aliked} and learned matchers like LightGlue~\cite{lindenberger2023lightglue} enable more reliable matching under challenging appearance changes.
More recently, fueled by significant advances in large-scale model training, the community has achieved breakthrough results by solving the multi-view 3D reconstruction problem end-to-end with a single feedforward architecture, effectively eliminating the need for explicit geometric solvers and optimization.
After a long period of incremental progress, these methods have enabled a substantial leap forward, resolving several persistent failure cases that challenged traditional approaches.

However, these methods are no silver bullet. In many cases, they still lag behind classical techniques in terms of scalability, accuracy, and robustness.
The reliance on heavy transformer-based architectures leads to high memory requirements, limiting scalability to only several hundred input views at relatively low image resolutions, thus limiting accuracy.
Recent efforts~\cite{yang2025fast3r, deng2025vggt} have sought to address these scalability issues, but they still struggle to process beyond a thousand images.
Moreover, as we will demonstrate later, these approaches can exhibit counterintuitive behavior in which adding more input images does not necessarily improve the outputs.
For scenarios where classical SfM methods succeed, feedforward approaches typically lag significantly behind in terms of accuracy.
When feedforward methods do approach classical performance in these nominal cases, they often do so by incorporating optimization-based bundle adjustment (BA), which only partially bridges the gap.
In terms of robustness, none of the existing methods can reliably handle multiple connected components, resolve symmetric structures, or systematically reject outliers such as irrelevant input images.


Our key contributions are an analysis of the limitations of classical and feedforward 3D reconstruction. We then use these insights to propose a novel pipeline by combining their respective strengths.
Extensive experiments across multiple datasets show the benefits of our approach, achieving state-of-the-art results across a wide range of scenarios.





\section{Related Work}
\subsection{Classical Methods\label{sec:related_classical}}

Classical SfM algorithms broadly categorize into incremental and global methods.
Both types begin with correspondence search, which typically involves extracting local image features~\cite{lowe2004distinctive, detone2018superpoint, zhao2023aliked} and matching them across images~\cite{lindenberger2023lightglue}. To establish the view graph, images can be paired exhaustively or, for greater scalability, by employing image retrieval techniques~\cite{arandjelovic2016netvlad, schoenberger2016vote} to identify overlapping image pairs.
Correspondence search concludes by estimating two-view geometries~\cite{hartley2003multiple} using robust estimation techniques.

The primary distinction between incremental and global pipelines lies in how they estimate cameras and structure.
Incremental SfM pipelines -- such as Bundler~\cite{snavely2006photo} or COLMAP~\cite{schonberger2016structure} -- build the reconstruction by adding images one at a time. At each step, they perform robust BA to jointly refine cameras and 3D points, interleaving this optimization as new images are incorporated.
Global SfM pipelines, in contrast, estimate all cameras and 3D points simultaneously from the entire set of images.
Historically, incremental methods have been regarded as more robust, particularly in challenging scenarios. However, recent methods like GLOMAP~\cite{pan2024global} demonstrate that global SfM pipelines can now achieve comparable robustness and accuracy but with better efficiency and scalability characteristics.
In our work, we build upon the global SfM paradigm.


In global SfM, camera intrinsics, rotations, and translations are usually recovered in different stages.
For camera intrinsics, Sweeney~\etal~\cite{sweeney2015optimizing} proposed to perform view-graph calibration.
The estimated intrinsics are then used to decompose two-view geometries into relative camera poses before estimating global camera rotations using rotation averaging~\cite{govindu2001combining, martinec2007robust, chatterjee2013efficient}.
Next, many systems use translation averaging~\cite{zhuang2018baseline, ozyesil2015robust, arrigoni2018bearing} to solve for global camera translations, but since pairwise translations are only up-to-scale, the formulation is often ill-posed.
GLOMAP~\cite{pan2024global} addresses this with global positioning to simultaneously recover camera poses and 3D points, but it can get stuck in local minima when tracks are insufficient.
Similarity averaging~\cite{cui2015global} takes a different approach by using depth-derived scale constraints but remains sensitive to noise.
All approaches finalize the reconstruction with global BA to improve accuracy.

Classical SfM methods face significant challenges in certain types of scenes, particularly those that are texture-less, have low image overlap, or exhibit low parallax and symmetries.
In areas with \textbf{low texture}, local feature matching becomes unreliable or even impossible. Without sufficient correspondences, accurate two-view geometry estimation is not feasible. This, in turn, leaves BA under-constrained, often resulting in reconstruction failures.
When images have \textbf{low overlap}, it becomes difficult to constrain the scale of the reconstruction. Since relative camera poses estimated from two views are only determined up to an unknown scale, at least three-view overlap is necessary to achieve consistent and accurate global scale.
Furthermore, relative pose estimation becomes degenerate when image pairs exhibit \textbf{low parallax} (\ie, when the camera motion is mostly rotational or the scene is far away). This degeneracy can cause the entire pipeline to fail.
Last but not least, the presence of visually similar or \textbf{symmetric scene structure} (\ie Doppelgangers~\cite{zach2010disambiguating,cai2023doppelgangers}) leads to unresolved ambiguities and most often results in collapsed 3D reconstructions.

\subsection{Feedforward Methods\label{sec:related_ff}}
Recent feedforward methods approach 3D reconstruction by learning to infer scene geometry and cameras in an end-to-end fashion, leveraging large-scale training datasets that typically include both synthetics and 3D models generated by classical techniques.
This allows them to learn complex priors to solve some of the failure cases of classical methods.
Depending on their network architecture, feedforward methods can be broadly categorized into three groups: diffusion, recurrent networks, and transformers.

CameraAsRays~\cite{zhang2024cameras} and PoseDiffusion~\cite{wang2023posediffusion} are two examples from the first category.
Initialized from random positions, they adopt diffusion processes to obtain the final pose estimation.
CUT3R~\cite{wang2025continuous} and its follow-ups are recurrent methods.
It maintains a scene state and incrementally reconstructs each incoming image.
DUSt3R~\cite{wang2024dust3r} and MASt3R~\cite{leroy2024grounding} represent two-view transformer-based methods.
Images are patchified into a set of tokens and then fed through a set of transformation layers.
These networks then directly regress 3D points in both images, and estimate camera poses and calibrations via RANSAC.
However, since it only receives two-view input, the scale between image pairs is estimated by non-metric depth, which is often unstable.
Methods like VGGT~\cite{wang2025vggt}, MV-DUSt3R~\cite{tang2025mv}, and MapAnything~\cite{keetha2025mapanything} took a step further to directly estimate multi-view reconstructions end-to-end.
$\pi^3$~\cite{wang2025pi} currently represents the state of the art by improving upon VGGT using a permutation invariant loss formulation.
However, these methods have their intrinsic limitations, as we show later.

In terms of \textbf{scalability}, diffusion- and transformer-based methods are inherently bound by GPU memory.
Even though Fast3R~\cite{yang2025fast3r}, FastVGGT~\cite{shen2025fastvggt}, StreamVGGT~\cite{zhuo2025streaming}, and SAIL-Recon~\cite{deng2025sail} propose mechanisms to reduce memory usage, they are still limited to several hundreds of images.
For recurrent networks, though theoretically scalable, they maintain a fixed state or they suffer from the same limitation as incremental classical approaches, where a single bad decision can lead to an unrecoverable failure.

Transformer-based models achieve the best \textbf{accuracy}~\cite{wang2025vggt, wang2025pi} across all feedforward approaches.
Yet, in scenes where classical methods work well, transformer-based models still lag significantly behind in terms of camera pose accuracy -- a limitation that is often underreported in prior literature.
To better understand these challenges, we conduct an extensive analysis on datasets that are difficult for both classical and feedforward methods.

In terms of \textbf{robustness}, diffusion- and transformer-based models for multi-view estimation treat all images equally, enabling every image to interact with another. This full connectivity works well for object-centric scenes.
However, in complex, large-scale scenes -- characterized by a large view graph radius -- attending information globally becomes problematic.
The resulting quadratic increase in possible connections makes it difficult to distinguish relevant from irrelevant information, leading to significant performance drops.
This problem is exacerbated in the presence of symmetric scene structure.
In contrast, recurrent network models depend on a specific, typically sequential, input order. If images are provided in random order, or if an image lacks visual overlap with previously processed images, pose estimation can fail entirely.


\subsection{Hybrid Methods}

There are some existing attempts in leveraging the merits of both categories to improve system performance. 

Early feedforward methods primarily aimed to enhance individual components of the classical SfM pipeline, such as image retrieval, pair filtering, feature extraction, feature matching, and optimization.
For instance, NetVLAD~\cite{arandjelovic2016netvlad} and more recent methods like SALAD~\cite{izquierdo2024optimal} or MegaLoc~\cite{berton2025megaloc} showcased strong improvements over traditional bag-of-words approaches.
Doppelgangers~\cite{cai2023doppelgangers} and its successor Doppelgangers++~\cite{xiangli2025doppelgangers++} developed feedforward networks to identify image pairs with symmetry issues.
SuperPoint~\cite{detone2018superpoint} and ALIKED~\cite{zhao2023aliked} are notable learned feature extractors, while SuperGlue~\cite{sarlin2020superglue} and LightGlue~\cite{lindenberger2023lightglue} proposed feedforward methods for feature matching.
These methods serve as an add-on to classical SfM, leaving the optimization formulation unchanged.
Meanwhile, PixSfM~\cite{lindenberger2021pixel} performs a joint refinement over learned features and structure.
Liu~\etal~\cite{liu2024hybrid} developed a hybrid SfM system to jointly reconstruct points and lines.
MP-SFM~\cite{pataki2025mp} addresses some limitations of classical SfM by incorporating learned monocular priors into an incremental pipeline. However, it struggles to scale to large problem instances, due to the incremental nature of the pipeline and the computational cost of depth and normal optimization.

On the feedforward side, efforts have been made in improving scalability and accuracy.
MASt3R-SfM~\cite{duisterhof2025mast3r} performs two-view inference on the view graph and obtains a multi-view reconstruction by minimizing the 3D point cloud alignment and reprojection errors.
VGGT-Long~\cite{deng2025vggt} and VGGT-SLAM~\cite{maggio2025vggt} propose to divide a long sequential input into small segments and apply factor graph optimization to align them while 
VGGT~\cite{wang2025vggt} and VGGSfM~\cite{wang2024vggsfm} produce tracks as part of outputs and inject these into bundle adjustment for improving the final pose accuracy.
However, these methods are often still less accurate than the classical SfM systems.
In contrast, our method achieves state-of-the-art performance under a wide range of input scenarios and scales well to tens of thousands of images.

\begin{figure*}[t]
\centering
\vspace{-10px}
\includegraphics[width=\linewidth]{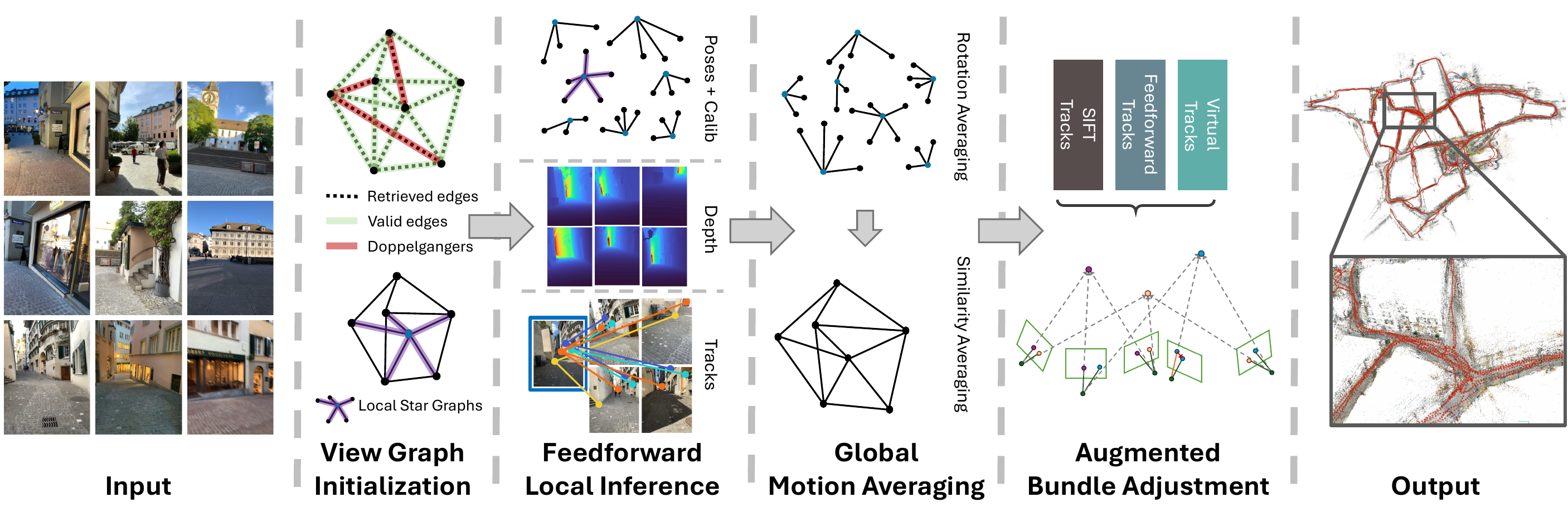}
\vspace{-20px}
\caption{Illustration of our proposed GLUEMAP pipeline consisting of four major steps: view graph initialization, feedforward local inference, global motion averaging, and augmented bundle adjustment. 
The proposed method combines the advantages of both feedforward and classical methods and can efficiently scale to tens of thousands of images while remaining robust.
}
\label{fig:pipeline}
\end{figure*}

\section{Method}

In this section, we introduce our approach for integrating classical and feedforward techniques into a unified, end-to-end reconstruction system.
The overall pipeline consists of four key stages: (1) \textit{view graph initialization} selects tentatively overlapping view pairs using scalable image retrieval and Doppelganger filtering, (2) from the resulting local star graphs centered on each input view, \textit{feedforward local inference} then estimates local reconstructions, (3) \textit{global motation averaging} performs rotation and similarity averaging to initialize the global reconstruction from the local stars, and (4) \textit{augmented bundle adjustment} jointly refines camera poses and structure using a combination of classical SIFT and virtual feedforward tracks.
An overview of our GLUEMAP system is shown in Figure~\ref{fig:pipeline}.

Our system takes as input a set of $i = 1...n$ unordered images $\mathcal{I} = \{ I_i \in \mathbb{R}^{H_i \times W_i \times 3} \}$ and estimates $k = 1...m$ scene points $\mathcal{X} = \{ X_k \in \mathbb{R}^3 \}$ as well as camera poses $\mathcal{P} = \{ P_i = (R_i, t_i) \in \mathrm{SE}(3)\ |\ i \in \mathcal{I}^*\}$ and intrinsics $\mathcal{K} = \{ \pi_i \in \mathbb{R}^3 \xrightarrow{} \mathbb{R}^2 |\ i \in \mathcal{I}^*\}$ for 
a subset $\mathcal{I}^*$ of confidently registered images.
Further technical details for each stage are provided in the following sections.


\subsection{View Graph Initialization}

In this stage, we begin by tentatively selecting overlapping image pairs using scalable image retrieval techniques combined with Doppelganger filtering.
Instead of the standard feedforward approach of globally attending over all images, we use the resulting sparse view graph $G(\mathcal{I}, \mathcal{E})$ to only locally attend feedforward reconstruction.
This allows our approach to more efficiently scale to an arbitrary number of input images, in particular by avoiding out-of-memory issues and by batch-reconstructing many local problems in parallel.
Furthermore, by restricting attention to only the most relevant information and by explicit Doppelganger filtering, feedforward inference becomes significantly more accurate and robust against symmetry issues.


More specifically, for each input image $I_i$, we retrieve a fixed number of $c$ candidate neighbors $\mathcal{C}_i$ using SALAD~\cite{izquierdo2024optimal}, resulting in a total of $O(c\cdot n)$ candidate pairs.
Then, for each of the pairs $(i, j), j\in\mathcal{C}_i$, we identify potentially non-overlapping or Doppelganger edges as
\begin{equation}
    \alpha_{ij} = \text{DG}(I_i, I_j)
\end{equation}
where $\alpha_{i, j}$ are Doppelgangers++ ($\text{DG}$)~\cite{xiangli2025doppelgangers++} scores. To ensure local connectivity, we apply dynamic thresholding on the scores as follows.
Starting from an empty view graph $G_{t=0}$ with $\mathcal{E}_{t=0} = \emptyset $, we iteratively add edges between different connected components $\text{CC}(i)$ as
\begin{equation}
    \mathcal{E}_{t+1} = \mathcal{E}_{t} \cup \{(i,j) \; | \; \alpha_{ij} > \delta_t, \text{CC}(i) \neq \text{CC}(j) \}
\end{equation}
with an initial filtering threshold $\delta_{0} = 0.8$.
If the updated graph $G_{t+1}$ is connected, the iteration halts and the graph is accepted.
Otherwise, the threshold is lowered as $\delta_{t+1} = \delta_t - 0.1$.
The iteration halts if $\delta_t < 0.2$ and the largest connected component is kept.
The final view graph $G_T$ defines a local neighborhood for each image $l$ as $\mathcal{N}_l = \{l\} \cup \{m \; | \; (l,m) \in \mathcal{E}_T\}$.


\subsection{Feedforward Local Inference}
Next, feedforward reconstruction proceeds from the established view graph.
To this end, the view graph is decomposed into local star graphs $S_l = G(\mathcal{N}_l,\; \{(l, m) \; | \; m\in\mathcal{N}_l, l\not= m \})$ which we batch-reconstruct independently as
\begin{equation}
    (\mathcal{P}_l, \mathcal{F}_l, \mathcal{D}_l, \mathcal{T}_l) = \text{FF}(I_{\mathcal{N}_l})  \enspace ,
\end{equation}
where we use $\pi^3$~\cite{wang2025pi} ($\text{FF}$) to infer local poses $\mathcal{P}_{l} = \{ P_i |\ i \in  \mathcal{N}_l \}$, depth maps $\mathcal{D}_l = \{ D_i \in \mathbb{R}_0^{H_i \times W_i} |\ i \in  \mathcal{N}_l \}$, focal lengths $\mathcal{F}_l = \{ f_i \in \mathbb{R}^+ |\ i \in  \mathcal{N}_l \}$, and tracks $\mathcal{T}_l$~\cite{wang2024vggsfm}.
We keep the 25 frames with the highest DG scores if there are more neighbors than that.

Since each image is part of $\vert \mathcal{N}_l \vert$ local star graphs, we infer overlapping local reconstructions.
To merge the overlapping tracks across stars, we snap track positions to SIFT~\cite{lowe2004distinctive} keypoints within a radius $\beta = 1\text{px}$ and merge tracks snapping to the same keypoints. The global set of merged tracks across all stars is denoted as $\mathcal{T}$.

For each local star, we further apply a forward-backward depth consistency check to determine visual overlap. 
More specifically, we calculate the one-way reprojection error $\epsilon_{i\to j}$ from any image $i$ to $j$ in star $S_l$ as
\begin{align}
    X_i &= D_i(x,y) \cdot (u, v, 1)^\top \\
    (x', y')^\top &= \Pi_j(R_{ij}X_i + t_{ij}) \\
    X_j &= D_j(x',y') \cdot (u', v', 1)^\top \\
    (x'', y'')^\top &= \Pi_i(R_{ij}X_j + t_{ij}) \\
    \epsilon_{i\to j} &= \|(x, y)^\top - (x'', y'')^\top\|_2 \enspace ,
\end{align}
where $(x, y) \in \mathbb{R}^2$ are pixel image coordinates and $(u, v, 1) = \pi^{-1}(x, y)$ are normalized coordinates, respectively. 
For simplicity, $l$ is omitted in the above equations.
Using a reprojection threshold $\tau$, the raw overlap ratio $\tilde{o}_{ij}^l$ between $i$ and $j$ in star $S_l$ is defined as 
\begin{equation}
    \tilde{o}_{ij}^l = \frac{1}{W_i H_i}\cdot\sum_{(x,y)\in H_i\times W_i} \mathds{1}(\epsilon_{i\to j} < \tau) \enspace .
\end{equation}
To measure transitive co-visibility, we define
\begin{equation}
    o_{ij}^l = \max_{\tilde{\mathcal{O}}\in\mathcal{O}_{i,j}} \prod_{(p, q)\in \tilde{\mathcal{O}}} \tilde{o}_{pq}^l \enspace ,
\end{equation}
where $\mathcal{O}_{ij}$ represents all the paths between image $i$ and $j$ in the fully connected graph on $\mathcal{N}_l$.
Edges with small $o_{ij}^l$ are filtered unless removing them disconnects the view graph.





\subsection{Global Motion Averaging}
Using global SfM techniques, this stage merges the $n$ independent local reconstructions into a global one. 
More specifically, camera intrinsics, rotations, and centers are estimated with intrinsics averaging, rotation averaging, and similarity averaging, respectively. 

First, for intrinsics averaging, we simply calculate the median of all inferred focal lengths per physical camera.

Next, rotation averaging, also known as rotation synchronization, estimates global camera rotations $R_i$ from a set of relative rotations $R_{ij}$ by optimizing
\begin{equation}
    \min_{R} \sum_{(i,j)\in E} \rho\left(o^l_{ij} \cdot d(R^l_{ij}, R_j R_i^\top)\right) \enspace ,
\end{equation}
where $d$ is the geodesic error function, $\rho$ is the Huber loss as a robustifier, and $R^l_{ij}$ is the relative rotation derived from the locally inferred camera poses in star $S_l$.

After global rotation averaging, we use similarity averaging ~\cite{cui2015global} to infer camera centers $c_i \in \mathbb{R}^3$.
Relative camera translations can be calculated as
\begin{equation}
    t^l_{ij} = s^{l} \cdot R^l_{ij} (c_i - c_j) \enspace ,
\end{equation}
where $t^l_{ij}$ is the relative translation from locally inferred camera poses in star $S_l$.
Since relative translations within local star reconstruction are scale-consistent, only a single scale $s^l$ is needed for each star.
Note that this is different from the original formulation~\cite{cui2015global}, where relative scales between individual edges are estimated from noisy triangulations, which is more error-prone than our formulation.

The camera centers $c_i$ and star scales $s_l$ are estimated by
\begin{equation}
    \min_{c, s} \sum_{l, (i, j)\in S_l} o_{ij}\cdot d\left(R_{ij}^\top t_{ij} - s_{l}\cdot(c_i-c_j)\right), \; s_0 = 1
\label{eq:sa}
\end{equation}
We discuss an alternative formulation and its relation to translation averaging in the supplementary material.
We initialize this optimization using the maximum spanning tree, where the weight of each edge is the overlap ratio $o_{ij}^l$.

Finally, globally scale-consistent depth maps can be computed as $\tilde{D}^{l}_{i} = \tfrac{1}{s_l}D_i^l$
with $\tilde{D}_i^l$ as the depth map for image $i$ in $S_l$.




\subsection{Augmented Bundle Adjustment}
The accuracy of the final reconstruction is improved by a process referred to as augmented bundle adjustment (BA).
As discussed in Section~\ref{sec:related_classical}, standard BA formulations are only well-conditioned when there are sufficiently many tracks $\mathcal{T}$ with many-view overlap~\cite{pataki2025mp}.
Due to low-overlap view configurations, it can be impossible to establish such tracks even with theoretically perfect matching algorithms. Furthermore, even with sufficient overlap, low texture may prevent tracks with enough overlap in practice.
In contrast, multi-view feedforward models can overcome this drawback by leveraging sophisticated scene priors to infer accurate relative camera poses and consistent depth maps with two-view or sometimes with no view overlap at all.
We encode these scene priors by augmenting the standard BA formulation with \textit{virtual tracks} as follows.

We form two types of virtual tracks by reprojection of sampled pixels $(x, y)$ in each star's center image $l$ as
\begin{equation}
    \mathcal{V}^l_{l\to m} = \Pi_m \left( R_{lm}^l \left(\tilde{D}^l_l(x, y) \cdot \Pi_l^{-1}(x, y) - c_{lm}^l \right)\right)
\end{equation}
\begin{equation}
    \tilde{\mathcal{V}}^l_{l\to m} = \Pi_m \left(R_m \left({R_l}^\top\tilde{D}^l_l \cdot \Pi_l^{-1}(x, y) + c_l\right) - c_{m}\right) \enspace .
\end{equation}
The resulting $\vert \mathcal{N}_l \vert$-view tracks $\mathcal{V} = \{ \mathcal{V}^l_{l\to m} |\ m \in \mathcal{N}_l, l\not= m \}$ and equivalently $\tilde{\mathcal{V}}$ are conditioned on the respective poses from the feedforward local inference and the global motion averaging results.
In contrast to standard feature tracks, we allow virtual tracks to project outside of neighboring images, and the virtual 3D points may be observed behind the neighboring cameras.
For numerical stability, we ignore observations coinciding with the imaging plane during the optimization.
Empirically, we chose to sample $\approx 100$ virtual tracks with a ratio of $10\%$ of tracks being conditioned on global camera poses.
Intuitively, a higher ratio of tracks conditioned on global poses leads to a BA result closer to the output of global motion averaging.

For the final augmented BA problem, we use three types of tracks: tracks $\mathcal{T}$ from the feedforward network, virtual tracks $\mathcal{V}$ and $\tilde{\mathcal{V}}$ as described above, as well as classical SIFT tracks.
We use a standard reprojection cost function $\Pi$ over all images with Huber for SIFT and feedforward tracks, and with Arctan for virtual tracks as robustifiers.
For the snapping of $\mathcal{T}$, we obtain SIFT features as a free side product and found them to improve reconstruction accuracy.
3D positions of virtual tracks are known by construction, while those of the other tracks are obtained by triangulation.
\section{Experiments~\label{sec:experiments}}





In this section, we first empirically analyze the behavior of feedforward methods in terms of the most important structural properties of a 3D reconstruction problem.
We then compare the performance of classical, feedforward, and our proposed method in extensive experiments on datasets covering a wide range of 3D reconstruction challenges. Figure~\ref{fig:teaser} summarizes all results.

\subsection{Metrics}
Throughout all experiments, we follow standard practice~\cite{wang2025vggt, pan2024global, wang2025pi} and measure AUC@X (Area Under the recall Curve) scores calculated based on pose errors, which is defined as the maximum of relative rotation and translation errors between every possible image pair at different angular error thresholds $X$.
For tighter thresholds, the scores reflect accuracy of the reconstruction.
For looser thresholds, they reflect the completeness of the reconstruction.

\subsection{Feedforward Graph Analysis}

\begin{figure}[t]
\centering
\includegraphics[width=\columnwidth]{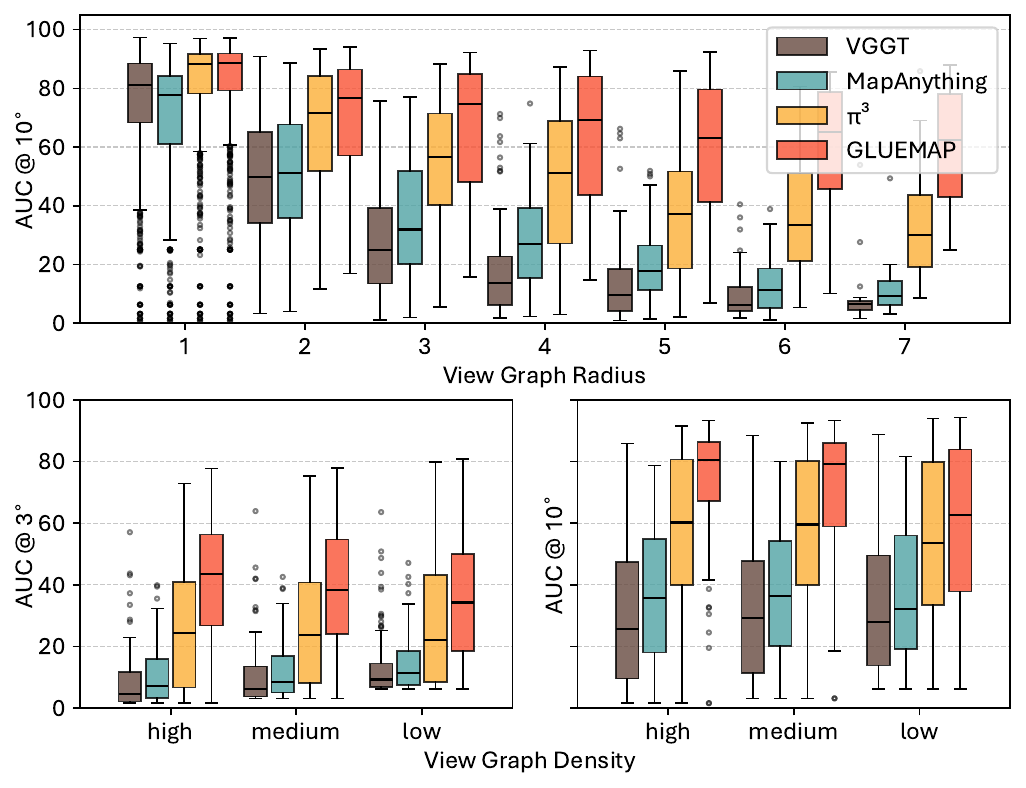}
\vspace{-20px}
\caption{Impact of view graph radius and density. The performance of feedforward methods drops with increasing radius and stays relatively stable with different densities. Our method is more robust to increasing radius and benefits from higher density.}
\label{fig:ff_analysis}
\end{figure}

In this experiment, we evaluate the performance of recent state-of-the-art feedforward methods with VGGT~\cite{wang2025vggt}, MapAnything~\cite{keetha2025mapanything} (\textit{MA}), and $\pi^3$~\cite{wang2025pi}, as well as the proposed motion averaging result, in terms of structural scene complexity.
To this end, we sample view graphs with varying radius $r = \min\max_{i, j \in \mathcal{I}} \delta(i, j)$ and 
density (\ie the Fiedler value / algebraic connectivity~\cite{fiedler1973algebraic} of the view graph $\mathcal{G}$, given by $\lambda_2$ of its graph Laplacian~\cite{chung1997spectral}).
For this evaluation, we chose LaMAR~\cite{sarlin2022lamar} as it contains a large number of sequences covering both indoor and outdoor scenes with a large variety in terms of view graph properties.
The sampling details can be found in the supplementary material.

The view graph radius -- defined as the minimum eccentricity among all vertices in a connected graph -- is used as a measure of scene complexity.
It determines the minimum number of passes required to exchange information between every pair of images in the graph, reflecting the intrinsic difficulty of the reconstruction problem.
We determine the ground truth view graph for each subsequence by rendering depth maps from the provided mesh. 
The results are summarized in the upper part in Figure~\ref{fig:ff_analysis}, where we observe a consistent performance drop with increasing radius, while $\pi^3$ consistently outperforms VGGT and MapAnything.
The proposed method achieves the best accuracy, especially for view graphs with a large radius.
This is because the proposed method inherits the classical SfM pipeline, thus it is more stable with increasing graph radius.

The Fiedler value $\lambda_2$ controls the convergence rate of iterative optimization on the view graph, including both motion averaging and bundle adjustment ~\cite{eriksson2018rotation}.
A lower value signals weakly connected components, which hinder information flow, impede error dissipation, and exacerbate noise accumulation.
To analyze its impact across methods, we compare the performance on the same segments with different frame sampling rates.
A higher sampling rate leads to sparser inputs.
Results are summarized in the lower part in Figure~\ref{fig:ff_analysis}.
For lower densities, counterintuitively, the accuracy of pose estimation improves.
This difference is especially observable for VGGT~\cite{wang2025vggt} and MapAnything~\cite{keetha2025mapanything}.
For higher densities, the performance stays at similar levels, with a slightly increased variance.
This behavior is further underlined by the results in Tables~\ref{tbl:imc}, where performance of feedforward methods degrades with more input views of the same scene.
Inversely, the proposed method behaves like other optimization-based pipelines, and the performance increases with the density of input, thanks to the additional redundancy of observations.





\subsection{Comparative Analysis on Real-World Datasets}
We conduct experiments on multiple datasets covering a wide range of challenges: ETH3D~\cite{schops2017multi}, IMC2021~\cite{jin2021image}, CO3Dv2~\cite{reizenstein2021common}, SMERF~\cite{duckworth2024smerf} and LaMAR~\cite{sarlin2022lamar}.

We evaluate state-of-the-art methods from both the classical and feedforward categories.
For classical ones, we chose GLOMAP + SIFT (\textit{SIFT}), GLOMAP + ALIKED + LightGlue (\textit{AL+LG}) as baselines.
To rule out the impact of image pairs with symmetry, for the classical baselines, we use the same input view graph using Doppelganger++~\cite{xiangli2025doppelgangers++} filtering.
This removes the potential performance differences caused by symmetry.
For feedforward methods, we first compare their performance on ETH3D~\cite{schops2017multi} and then select $\pi^3$ as the best performing method for all other experiments.
We also include results for $\pi^3$ with bundle adjustment (\textit{$\pi^3$ + BA}) and our results after global motion averaging (\textit{GLUEMAP$^\dagger$}) as additional baselines.
We additionally compare to MP-SFM with both sparse (\textit{MP-SFM (s)}) and dense (\textit{MP-SFM (d)}) tracks on ETH3D and SMERF.

Experiments are conducted on GH200 GPU with 96GB memory, but our method can fit on the RTX 4090 with 24GB memory.

\begin{table}[t]
    \centering
    \caption{Results on ETH3D~\cite{schops2017multi}. Results with * are with groundtruth calibration. GLUEMAP achieves best overall results while $\pi^3$ is the best feed-forward method.}
    \vspace{-5px}
    \setlength{\tabcolsep}{3pt} 
    \resizebox{\columnwidth}{!}{
    \begin{tabular}{l c c c c | c c c c c c
    } \toprule 
AUC@ & \small{SIFT} & \small{AL+LG} & \small{$\pi^3$} & \small{$\pi^3$ + BA} & \small{GLUEMAP$^\dagger$} & \small{GLUEMAP} & \small{GLUEMAP*} \\
\midrule
1 & 45.6 & 42.9 & 13.2 & 30.6 & 20.3 & 53.0 & \cellcolor{tabsecond}74.0 \\
3 & 62.2 & 62.1 & 36.1 & 55.1 & 49.0 & \cellcolor{tabsecond}76.9 & \cellcolor{tabfirst}85.9 \\
5 & 66.7 & 67.4 & 48.9 & 65.1 & 61.9 & 83.6 & \cellcolor{tabfirst}89.0 \\

    \end{tabular}
    }
    
    \resizebox{\columnwidth}{!}{
    \begin{tabular}{l c c c c  c c c c c c
    } \toprule 
AUC@ & \small{CUT3R} & \small{VGGT} & \small{MA} & \small{MASt3R-SfM} & \small{MP-SFM* (s)} & \small{MP-SFM* (d)} \\
\midrule
1 & 5.0 & 8.6 & 5.1 & 39.2 & \cellcolor{tabfirst}74.3 & 70.3 \\
3 & 11.4 & 24.0 & 11.1 & 55.6 &  -  &  -  \\
5 & 18.8 & 35.0 & 18.3 & 60.5 & \cellcolor{tabsecond}88.3 & 88.2 \\
      \bottomrule
    \end{tabular}
    }
    \label{tbl:eth3d}
\end{table}

\textbf{ETH3D}~\cite{schops2017multi} consists of an unordered collection of high-resolution images of both outdoor and indoor scenes with millimeter-accuracy groundtruth.
We use this dataset to analyze the performance of different methods with a focus on \textbf{accuracy}.
We adopt the same thresholds used in ~\cite{pan2024global}
and the results are summarized in Table~\ref{tbl:eth3d}.
To fairly compare with MP-SFM~\cite{pataki2025mp}, we also report results with ground truth intrinsics (marked with *).
Our method achieves the highest accuracy in both the calibrated and uncalibrated settings, with a large performance gap compared to feedforward methods.
The performance difference to classical methods is less pronounced. 
We attribute the slight performance gap to the local robustness provided by the feed-forward backbone, as well as some scenes exhibiting very low overlap.
Meanwhile, $\pi^3$ demonstrates a clear performance advantage over CUT3R~\cite{wang2025continuous}, VGGT~\cite{deng2025vggt}, and MapAnything~\cite{keetha2025mapanything}, supporting the choice of our backbone.

\textbf{IMC2021} (Image Matching Challenge 2021)~\cite{jin2021image} features unordered internet photo collections of outdoor landmarks captured by heterogeneous devices under drastic \textbf{appearance changes}.
The biggest challenge are extreme illumination and geometric changes.
We used the same thresholds used in ~\cite{wang2025vggt} and we report the results for different collections separately in Table~\ref{tbl:imc}.
From the table, when the number of images is small, with large appearance changes, classical methods struggle to establish enough matches, resulting in comparatively low accuracy.
However, when more images from the collection are included, low matching efficiency is compensated for by high density. 
As a result, GLOMAP + SIFT achieves the best performance.
Our method remains competitive across different input sizes, inheriting the benefits from both categories.

\begin{table}[t]
    \centering
    \caption{Results on IMC2021~\cite{jin2021image}. The relative performances of classical and feedforward methods changes with the number input views, while our method is competitive across all settings.}
    \vspace{-5px}
    \setlength{\tabcolsep}{3pt} 
    \resizebox{\columnwidth}{!}{
    \begin{tabular}{l l c c c l c c c l c c c l c c c
    } \toprule 
&& \multicolumn{3}{c}{bag 5} && \multicolumn{3}{c}{bag 10} && \multicolumn{3}{c}{bag 25} && \multicolumn{3}{c}{Full} \\

        \cmidrule{3-5} \cmidrule{7-9} \cmidrule{11-13} \cmidrule{15-17}
        AUC@&& 3 & 5 & 10 && 3 & 5 & 10 && 3 & 5 & 10 && 3 & 5 & 10 \\

\midrule
SIFT & & 39.6 & 47.3 & 57.0 & & 50.1 & 60.9 & 72.5 & & \cellcolor{tabfirst}64.4 & \cellcolor{tabfirst}74.3 & \cellcolor{tabsecond}84.2 & & \cellcolor{tabfirst}76.9 & \cellcolor{tabfirst}83.8 & \cellcolor{tabfirst}90.1 \\
AL+LG & & 48.4 & 58.5 & 70.6 & & 50.9 & 62.2 & 74.3 & & 54.7 & 64.9 & 75.6 & & 62.8 & 72.4 & 82.5 \\
$\pi^3$ & & 46.2 & 58.4 & 73.0 & & 39.7 & 53.4 & 69.7 & & 36.6 & 51.1 & 68.4 & & 35.2 & 49.2 & 66.2 \\
$\pi^3$ + BA & & \cellcolor{tabsecond}54.0 & \cellcolor{tabsecond}64.9 & \cellcolor{tabsecond}77.6 & & \cellcolor{tabsecond}54.1 & \cellcolor{tabsecond}65.3 & \cellcolor{tabsecond}77.5 & & 57.8 & 69.1 & 80.9 & & 45.8 & 54.8 & 65.9 \\
GLUEMAP$^\dagger$ & & 46.2 & 58.4 & 72.9 & & 39.8 & 53.5 & 69.9 & & 36.7 & 51.2 & 68.5 & & 36.7 & 51.7 & 69.1 \\
GLUEMAP & & \cellcolor{tabfirst}54.3 & \cellcolor{tabfirst}65.3 & \cellcolor{tabfirst}77.8 & & \cellcolor{tabfirst}58.0 & \cellcolor{tabfirst}69.3 & \cellcolor{tabfirst}81.2 & & \cellcolor{tabsecond}63.5 & \cellcolor{tabsecond}74.0 & \cellcolor{tabfirst}84.4 & & \cellcolor{tabsecond}73.0 & \cellcolor{tabsecond}81.3 & \cellcolor{tabsecond}89.1 \\

      \bottomrule
    \end{tabular}
    }
    \label{tbl:imc}
\end{table}

\textbf{CO3Dv2}~\cite{reizenstein2021common} features a large set of object-centric scenes presenting the methods with challenges in terms of \textbf{low-texture} and \textbf{low-overlap}.
Following the practice in ~\cite{wang2025vggt, wang2025pi}, we randomly sample 10 images from each test sequence, and report AUC scores there. 
We also sample at most 10 sequences from each category with 20 and 40 randomly sampled images.
Results are summarized in Table~\ref{tbl:co3d}.
On \textbf{sparse} sequences, a clear performance gap between classical and feedforward methods can be seen.
For denser sequences with more images, the gap diminishes.
Our method maintains high performance in both settings.

\begin{table}[t]
    \centering
    \caption{Results on CO3Dv2~\cite{reizenstein2021common}. Ours is on-par with feedforward approaches while classical ones fall far behind.}
\vspace{-5px}
    \resizebox{\columnwidth}{!}{
    \setlength{\tabcolsep}{3pt} 
    \begin{tabular}{l l c c c l c c c l c c c l c c c
    } \toprule 
&& \multicolumn{3}{c}{10 images} && \multicolumn{3}{c}{20 images} && \multicolumn{3}{c}{40 images} && \multicolumn{3}{c}{Average} \\

        \cmidrule{3-5} 
        \cmidrule{7-9} \cmidrule{11-13} \cmidrule{15-17}
        AUC@&& 3 & 10 & 30 && 3 & 10 & 30 && 3 & 10 & 30 && 3 & 10 & 30\\
\midrule
SIFT & & 25.8 & 36.3 & 42.1 & & 35.1 & 47.4 & 53.9 & & 50.0 & 65.9 & 73.3 & & 37.0 & 49.9 & 56.4 \\
AL+LG & & 20.6 & 29.2 & 35.0 & & 40.8 & 56.1 & 64.0 & & 52.9 & 71.5 & 80.7 & & 38.1 & 52.2 & 59.9 \\
$\pi^3$ & & 48.2 & 77.1 & 89.9 & & 46.4 & 76.2 & 89.2 & & 47.1 & 76.5 & 89.3 & & 47.3 & 76.6 & 89.5 \\
$\pi^3$ + BA & & \cellcolor{tabfirst}55.3 & \cellcolor{tabsecond}78.8 & \cellcolor{tabsecond}90.1 & & \cellcolor{tabfirst}59.4 & \cellcolor{tabfirst}80.9 & \cellcolor{tabfirst}90.9 & & \cellcolor{tabfirst}60.5 & \cellcolor{tabfirst}81.3 & \cellcolor{tabfirst}91.2 & & \cellcolor{tabfirst}58.4 & \cellcolor{tabfirst}80.3 & \cellcolor{tabfirst}90.7 \\
GLUEMAP$^\dagger$ & & 47.0 & 76.6 & 89.5 & & 47.1 & 76.6 & 89.3 & & 48.2 & 77.1 & 89.6 & & 47.4 & 76.8 & 89.5 \\
GLUEMAP & & \cellcolor{tabsecond}54.8 & \cellcolor{tabfirst}79.3 & \cellcolor{tabfirst}90.3 & & \cellcolor{tabsecond}56.7 & \cellcolor{tabsecond}79.8 & \cellcolor{tabsecond}90.3 & & \cellcolor{tabsecond}58.7 & \cellcolor{tabsecond}80.7 & \cellcolor{tabsecond}90.7 & & \cellcolor{tabsecond}56.7 & \cellcolor{tabsecond}79.9 & \cellcolor{tabsecond}90.4 \\

      \bottomrule
    \end{tabular}
    }
    \label{tbl:co3d}
\end{table}

\textbf{SMERF}~\cite{duckworth2024smerf} contains four indoor captures covering multiple rooms.
Following the setup in ~\cite{pataki2025mp}, we use it to evaluate \textbf{low-overlap} scenarios.
Results are summarized in Table~\ref{tbl:smerf_mpsfm} and indicate that classical methods largely fail with low overlap.
$\pi^3$ also does not achieve high scores because the view-graph radius and symmetry is high for scenes in this dataset, leading to multiple rooms collapsing in the reconstructions.
By using Doppelganger++ filtering, our method can successfully distinguish different rooms, achieving satisfying results after motion averaging.
And with the help of virtual tracks, the proposed method improves on the tightest threshold from bundle adjustment while avoiding significant degradation of reconstruction completeness, which is captured by AUC@20.
Notably, when compared with MP-SFM~\cite{pataki2025mp}, which was specifically targeting at low-overlap scenes, we achieve better accuracy than the version with sparse tracks.
For the dense version of MP-SFM, it benefits from the dense correspondences and achieves better scores on the tightest threshold, while we achieve a similar level of completeness.

\begin{table}[t]
    \centering
    \caption{Results on SMERF~\cite{duckworth2024smerf} benchmark established in MP-SFM~\cite{pataki2025mp}. Results marked with * are with groundtruth calibration. While both classical and feedforward methods fail, our proposed method is able to reconstruct scenes with low overlap.}
    \vspace{-5px}
    \resizebox{\columnwidth}{!}{
    \setlength{\tabcolsep}{3pt} 
    \begin{tabular}{l l c c c l c c c l c c c l c c c
    } \toprule 
&& \multicolumn{3}{c}{minimal} && \multicolumn{3}{c}{low} && \multicolumn{3}{c}{medium} && \multicolumn{3}{c}{high} \\

        \cmidrule{3-5} \cmidrule{7-9} \cmidrule{11-13} \cmidrule{15-17}
        
AUC@&& 1 & 5 & 20 && 1 & 5 & 20 && 1 & 5 & 20 && 1 & 5 & 20 \\
\midrule
SIFT & & 3.5 & 4.4 & 5.4 & & 1.4 & 1.6 & 1.8 & & 1.5 & 2.3 & 3.1 & & 13.0 & 19.2 & 23.2 \\
AL+LG & & 4.3 & 6.9 & 9.8 & & 2.4 & 6.1 & 9.8 & & 8.6 & 19.1 & 28.0 & & 28.6 & 46.8 & 57.0 \\
$\pi^3$ & & 3.2 & 18.0 & 51.7 & & 1.5 & 14.3 & 49.8 & & 1.3 & 15.7 & 52.1 & & 0.9 & 15.5 & 51.4 \\
$\pi^3$ + BA & & 3.1 & 18.5 & 54.1 & & 1.5 & 14.3 & 53.4 & & 1.2 & 14.7 & 52.2 & & 0.7 & 12.5 & 43.4 \\
MASt3R-SfM & & 3.9 & 10.4 & 18.0 & & 4.3 & 11.7 & 23.0 & & 5.9 & 15.8 & 28.1 & & 10.4 & 22.9 & 39.9 \\
MP-SFM* (s) & & 9.2 & 41.0 & 69.8 & & 5.4 & 29.1 & 53.0 & & 14.0 & 47.6 & 72.9 & & 47.3 & 79.3 & 90.6 \\
MP-SFM* (d) & & \cellcolor{tabfirst}17.2 & 54.6 & 77.1 & & \cellcolor{tabfirst}26.6 & 63.2 & 84.1 & & \cellcolor{tabfirst}40.4 & 72.8 & 87.5 & & \cellcolor{tabfirst}57.1 & 84.6 & 94.1 \\
GLUEMAP$^\dagger$ & & 9.8 & \cellcolor{tabfirst}55.5 & \cellcolor{tabfirst}82.4 & & 12.9 & 70.2 & 92.1 & & 20.3 & 76.3 & 93.9 & & 30.9 & 82.9 & 95.7 \\
GLUEMAP & & 10.1 & \cellcolor{tabsecond}54.9 & \cellcolor{tabsecond}82.0 & & 14.6 & \cellcolor{tabsecond}71.4 & \cellcolor{tabsecond}92.4 & & 27.7 & \cellcolor{tabsecond}79.1 & \cellcolor{tabsecond}94.6 & & 47.4 & \cellcolor{tabsecond}88.1 & \cellcolor{tabsecond}97.0 \\
GLUEMAP* & & \cellcolor{tabsecond}10.5 & 54.8 & 81.5 & & \cellcolor{tabsecond}14.7 & \cellcolor{tabfirst}71.8 & \cellcolor{tabfirst}92.5 & & \cellcolor{tabsecond}28.1 & \cellcolor{tabfirst}79.5 & \cellcolor{tabfirst}94.8 & & \cellcolor{tabsecond}47.8 & \cellcolor{tabfirst}88.3 & \cellcolor{tabfirst}97.1 \\

      \bottomrule
    \end{tabular}
    }
    \label{tbl:smerf_mpsfm}
\end{table}

\setlength{\tabcolsep}{2pt} 

\begin{figure}[t]
    \centering
    \resizebox{\columnwidth}{!}{
    \begin{tabular}{c c c c 
    } 
    \centering
    \rotatebox{90}{~~~~~~~~~~~~~~~HGE} &
    \includegraphics[width=0.33\columnwidth]{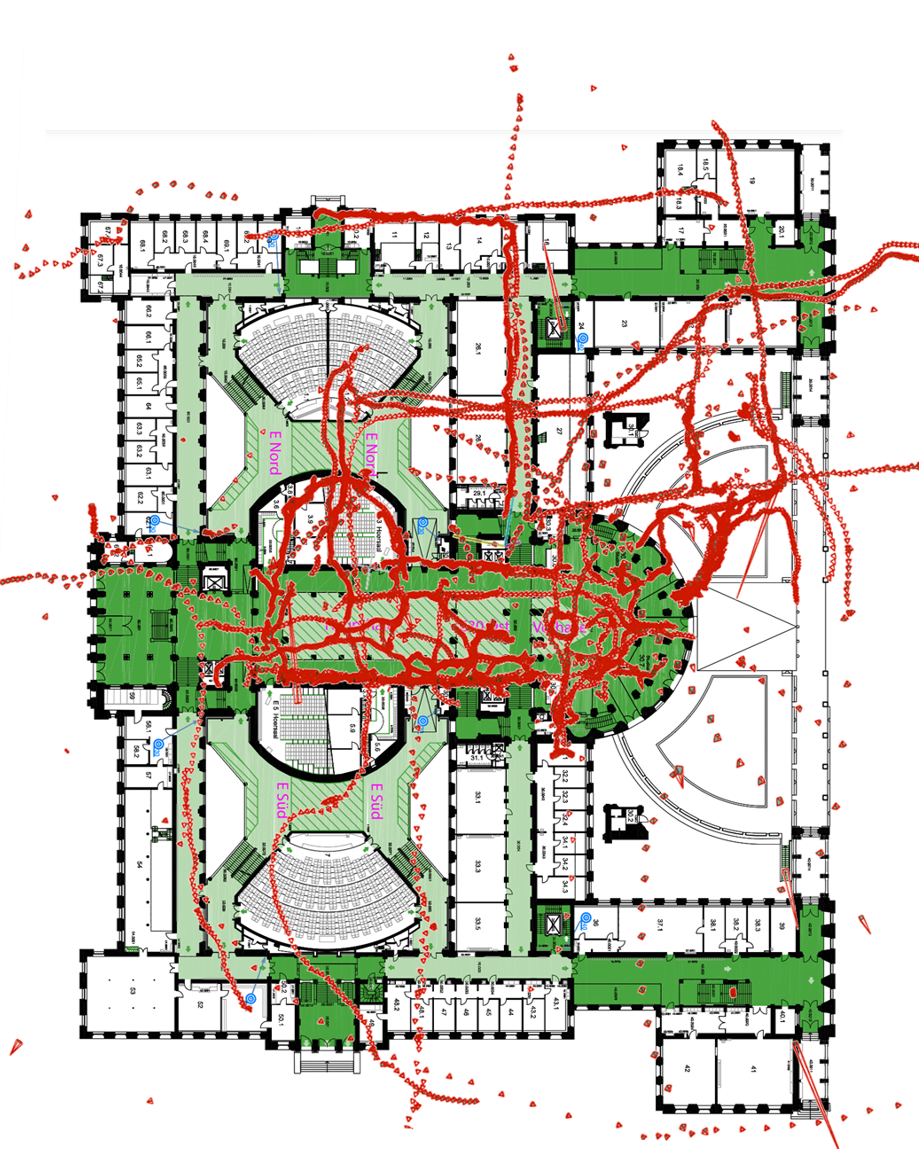} &
    \includegraphics[width=0.33\columnwidth]{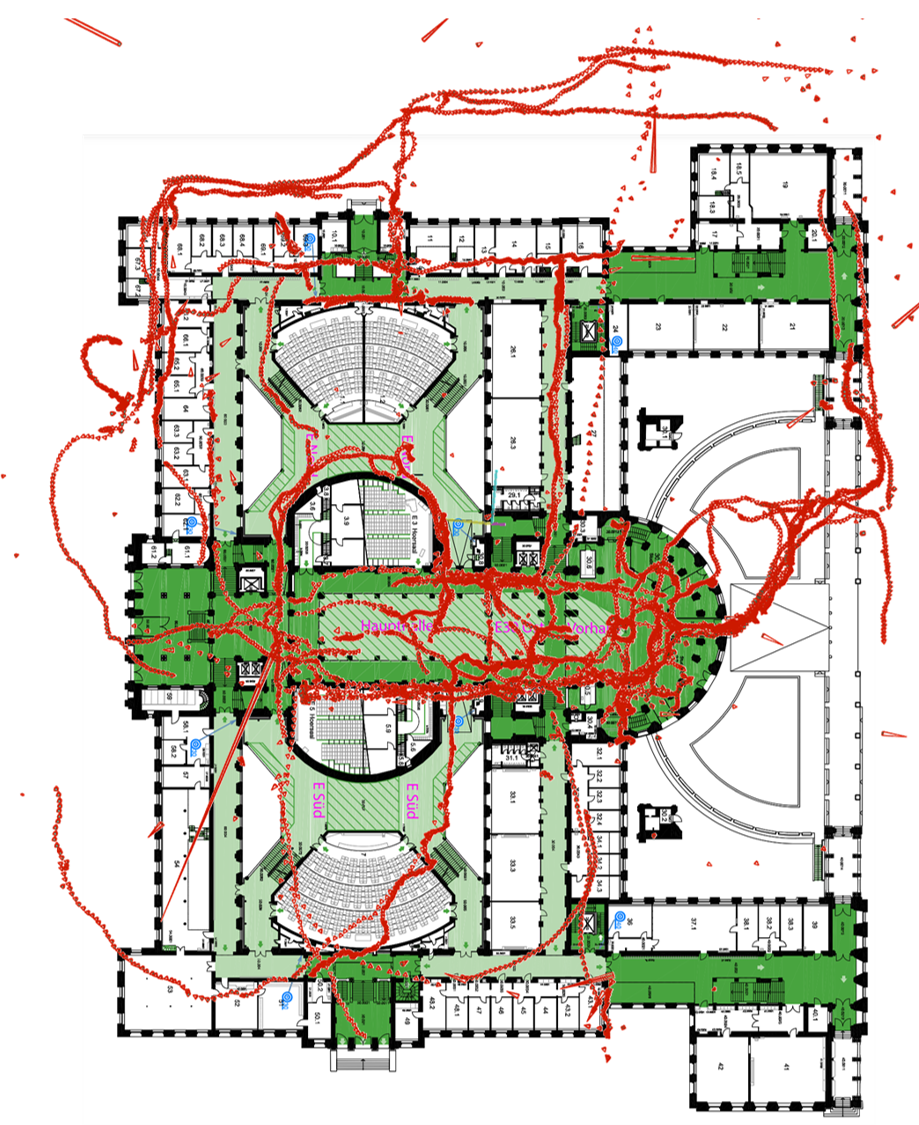} &
    \includegraphics[width=0.33\columnwidth]{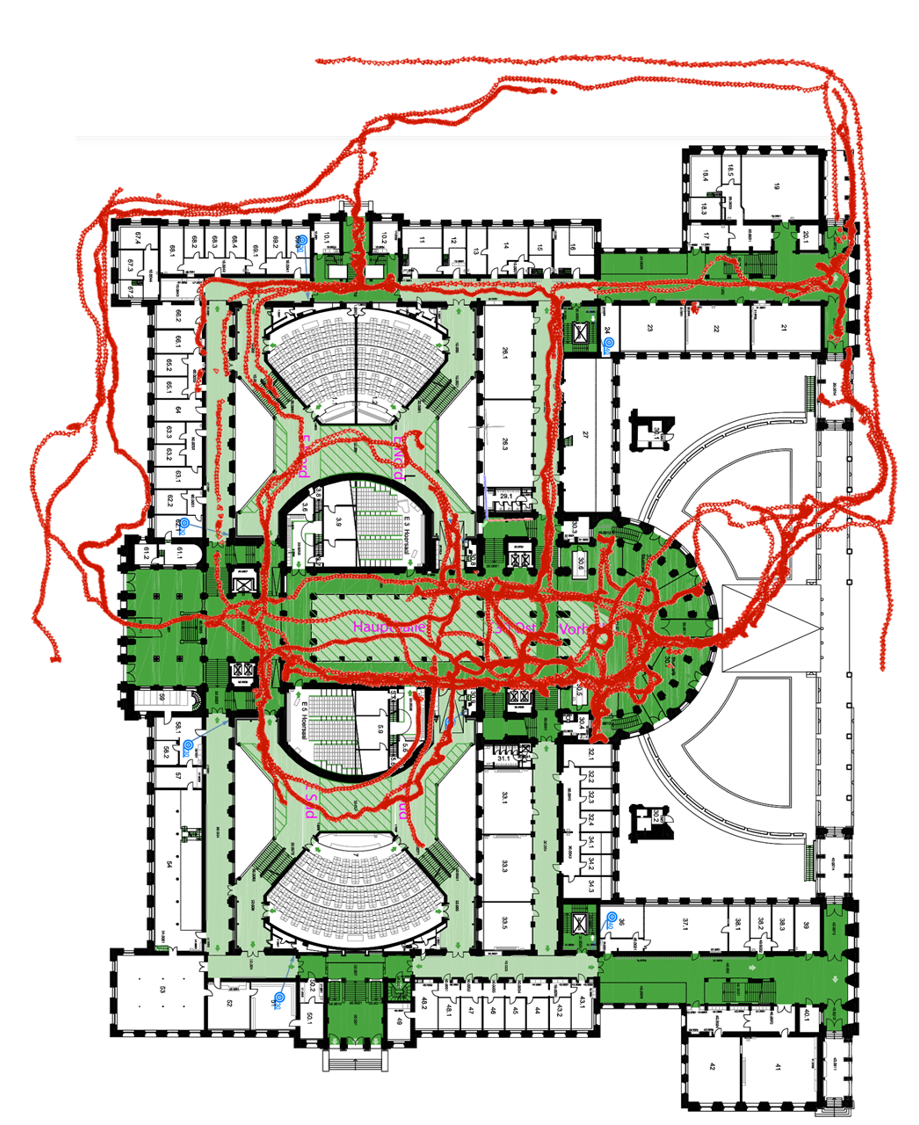} \\
    \rotatebox{90}{~~~~~~~~~~~~~~~LIN} &
    \includegraphics[width=0.33\columnwidth]{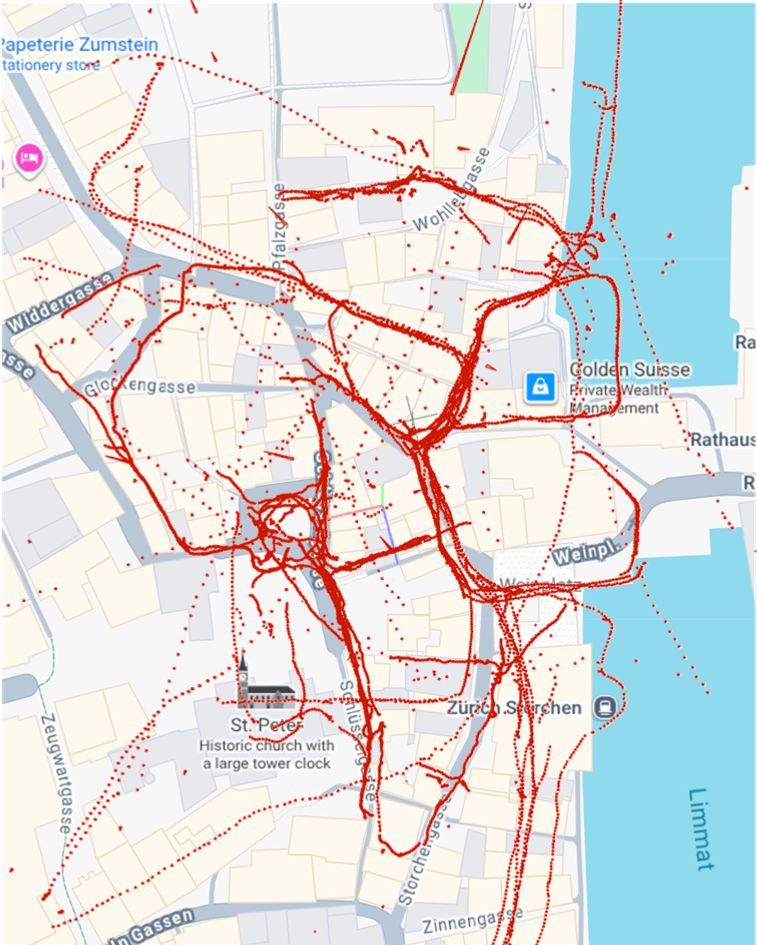} &
    \includegraphics[width=0.33\columnwidth]{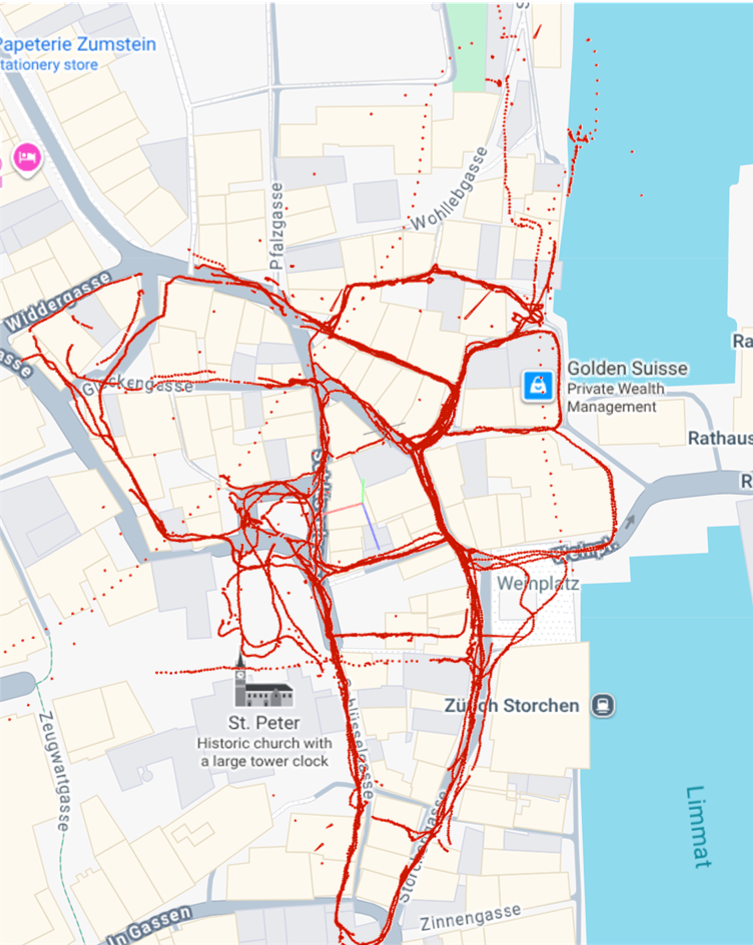} &
    \includegraphics[width=0.33\columnwidth]{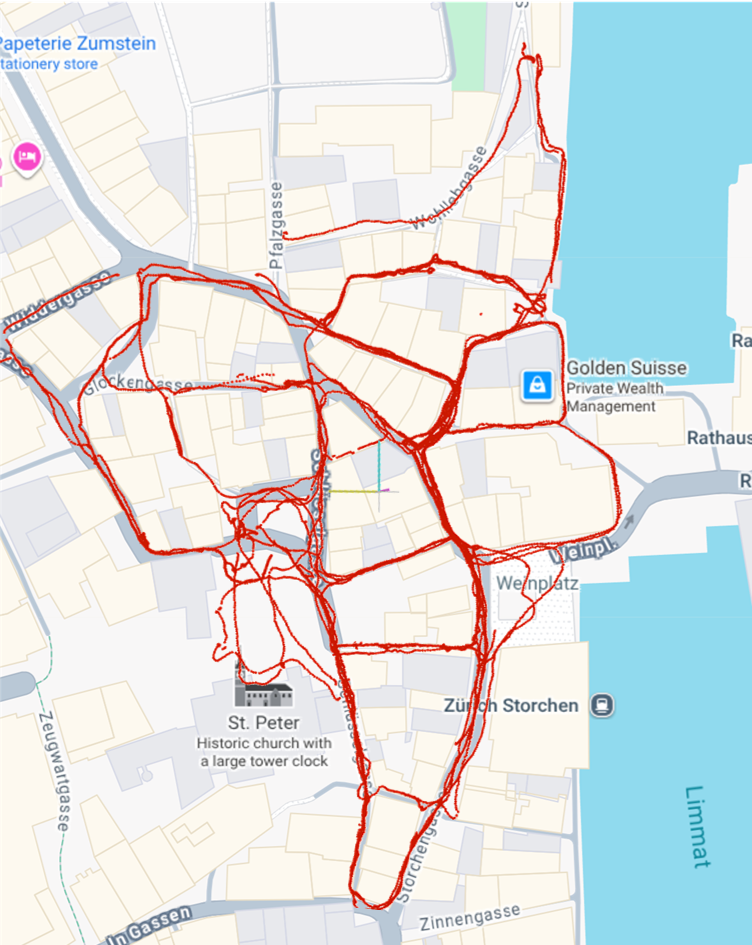} \\
    & (a) SIFT & (b) AL + LG & (c) GLUEMAP \\
\end{tabular}
}
    \caption{
    Qualitative reconstruction resuls of different methods on LaMAR~\cite{sarlin2022lamar}. Our reconstructions are close to the real-world annotation, indicating its high accuracy on this challenging dataset.
    }
\label{fig:lamar_vis}
\end{figure}

\textbf{LaMAR}~\cite{sarlin2022lamar} contains 3 large-scale indoor-outdoor scenes.
Each scene has many egocentric sequences with abundant causal movement and motion blur.
It presents the main challenges of \textbf{scalability} and \textbf{symmetries} while also exhibiting \textbf{low-overlap} and \textbf{low-texture} scenarios.
The view graph radii for CAB, HGE, and LIN are 49, 61, and 59, which underlines the difficulty of the problem, where feedforward methods already exhibit a drastic performance drop even for much smaller radii (\cf Figure~\ref{fig:ff_analysis}).


Results can be found in Table~\ref{tbl:lamar}.
In this experiment, we only use images captured by phones, as none of the evaluated methods natively support modeling rig constraints.
For this dataset, since scenes contain several thousand input views, feedforward methods fail due to out-of-memory issues.
Classical methods, especially GLOMAP with ALIKED and LightGlue, perform reasonable on LIN, which is an outdoor-only scene.
However, for the HGE and CAB scenes, classical methods struggle to estimate a good reconstructions.
In contrast, feedforward models provide good local reconstructions, which serve as a solid foundation for global motion averaging to obtain accurate global reconstructions by our method.
The accuracy of the proposed method on these datasets was further improved by BA, highlighting the robustness of the proposed method. 
From the qualitative results shown in Figure~\ref{fig:lamar_vis}, the reconstructions from our pipeline align closely with the real-world annotation, indicating its high accuracy.

\begin{table}[t]
    \centering
    \caption{Results on LaMAR~\cite{sarlin2022lamar}. Our method outperforms classical methods by a large margin while others entirely fail due to out-of-memory (OOM) issues.}
    \vspace{-5px}
    \setlength{\tabcolsep}{3pt} 
    \resizebox{\columnwidth}{!}{
    \begin{tabular}{l l c c c l c c c l c c c l c c c
    } \toprule 
&& \multicolumn{3}{c}{CAB (6587)} && \multicolumn{3}{c}{HGE (7553)} && \multicolumn{3}{c}{LIN (9319)} && \multicolumn{3}{c}{Average} \\
        \cmidrule{3-5} \cmidrule{7-9} \cmidrule{11-13} \cmidrule{15-17}
        
AUC@&& 3 & 10 & 30 && 3 & 10 & 30 && 3 & 10 & 30 && 3 & 10 & 30 \\
\midrule
SIFT && 0.6 & 1.4 & 2.8 && 2.6 & 12.4 & 33.3 && 4.6 & 23.7 & 48.8 && 2.6 & 12.4 & 28.3 \\
AL+LG && 1.1 & 2.9 & 6.2 && 8.0 & 31.7 & 58.9 && 23.7 & 55.7 & 72.8 && 10.9 & 30.1 & 46.0 \\
MASt3R-SfM && OOM & OOM & OOM && OOM & OOM & OOM && OOM & OOM & OOM && OOM & OOM & OOM \\ 
$\pi^3$ && OOM & OOM & OOM && OOM & OOM & OOM && OOM & OOM & OOM && OOM & OOM & OOM \\ 
$\pi^3$ + BA && OOM & OOM & OOM && OOM & OOM & OOM && OOM & OOM & OOM && OOM & OOM & OOM \\ 
GLUEMAP$^\dagger$ & & \cellcolor{tabsecond}2.6 & \cellcolor{tabsecond}21.0 & \cellcolor{tabsecond}61.2 & & \cellcolor{tabsecond}22.1 & \cellcolor{tabsecond}70.1 & \cellcolor{tabsecond}89.3 & & \cellcolor{tabsecond}30.2 & \cellcolor{tabsecond}69.9 & \cellcolor{tabsecond}83.9 & & \cellcolor{tabsecond}18.3 & \cellcolor{tabsecond}53.7 & \cellcolor{tabsecond}78.1 \\
GLUEMAP & & \cellcolor{tabfirst}4.5 & \cellcolor{tabfirst}27.1 & \cellcolor{tabfirst}64.8 & & \cellcolor{tabfirst}37.3 & \cellcolor{tabfirst}78.1 & \cellcolor{tabfirst}92.3 & & \cellcolor{tabfirst}37.3 & \cellcolor{tabfirst}72.1 & \cellcolor{tabfirst}84.6 & & \cellcolor{tabfirst}26.4 & \cellcolor{tabfirst}59.1 & \cellcolor{tabfirst}80.6 \\
      \bottomrule
    \end{tabular}
    }
    \label{tbl:lamar}
\end{table}


\subsection{Limitations \& Future Work}

Our method's performance critically depends on the quality of local reconstructions by feedforward methods.
For example, we can currently not handle fisheye images since the feedforward models are only trained on images taken by pinhole camera models, while later stages in our pipeline can theoretically handle them.
As newer feedforward models improve robustness and generality, our method will benefit as well.
Furthermore, the formulation currently does handle purely rotational motion due to our augmented bundle adjustment formulation.
Future work on incorporating (soft) depth priors from the local reconstructions can improve robustness in these situations.
Last but not least, our method currently requires the combination of different feedforward methods. An interesting direction for future work will be in developing a network that can solve the problem with a shared feedforward architecture.

\section{Conclusion}

In this work, we systematically analyze the strengths and weaknesses of both classical and feedforward approaches to 3D reconstruction.
We then present a novel end-to-end reconstruction system that integrates the advantages of both categories:
Starting with establishing a tentative view graph and leverages feedforward methods to perform local reconstructions.
Then, global motion averaging merges them to initialize an augmented bundle adjustment stage to improve the final accuracy.
We extensively evaluate our approach on diverse datasets encompassing a wide range of real-world challenges. By combining the local robustness of feedforward methods with the scalability, accuracy, and global consistency of classical techniques, our system achieves state-of-the-art performance on a wide range of scenarios.
We note that further evaluation on unordered image collections with severe appearance changes and transient objects remains an important direction for future work.

\paragraph{Acknowledgement}
This work was supported under project ID a144 as part of the Swiss AI Initiative, through a grant from the ETH Domain and computational resources provided by the Swiss National Supercomputing Centre (CSCS) under the Alps infrastructure.

{
    \small
    \bibliographystyle{ieeenat_fullname}
    \bibliography{main}
}

\clearpage
\setcounter{page}{1}
\maketitlesupplementary


\section{Ablations}
To understand the contribution of each component, we conduct ablation studies along three axes: the track types used in augmented bundle adjustment, the choice of feedforward backbone, and the covisibility filtering strategy.
\subsection{Augmented Bundle Adjustment}
Augmented bundle adjustment (A-BA) is ablated in Table~\ref{tbl:augmented_ba}, where all variants start from the same results after motion averaging.
Three types of tracks are considered: SIFT tracks from classical feature matching, deep tracks (DT) obtained from VGGSfM~\cite{wang2024vggsfm} inference, and virtual tracks (VT) synthesized by reprojecting sampled rays across neighboring views.
On ETH3D~\cite{schops2017multi}, SIFT tracks contribute most to accuracy at tight thresholds, confirming their precision for well-textured scenes.
On SMERF~\cite{duckworth2024smerf}, virtual tracks are essential for preventing drift across rooms with minimal overlap, while deep tracks provide complementary robustness under strong appearance changes, particularly for indoor scenes.

\begin{table}[h]
    \centering
    \caption{Ablation results for Augmented Bundle Adjustment.}
    \resizebox{\columnwidth}{!}{
    \setlength{\tabcolsep}{4pt} 
    \begin{tabular} {ccc l c c c l c c c l c c c 
    } \toprule 
 \multicolumn{3}{c}{AUC@}&& \multicolumn{3}{c}{ETH3D} && \multicolumn{3}{c}{SMERF \footnotesize{(minimal)}} && \multicolumn{3}{c}{SMERF \footnotesize{(low)}}  \\
    \cmidrule{1-3} \cmidrule{5-7} \cmidrule{9-11} \cmidrule{13-15}
    \multicolumn{3}{c}{Components} && 1 & 3 & 5 && 1 & 5 & 20 && 1 & 5 & 20 \\ 
    \midrule
 
\cellcolor{blue!10}DT & \cellcolor{green!10}SIFT & \cellcolor{orange!10}VT && 52.6 & 76.6 & 83.3 && \cellcolor{tabfirst}10.0 & \cellcolor{tabsecond}54.5 & \cellcolor{tabfirst}82.1 && \cellcolor{tabfirst}14.9 & \cellcolor{tabfirst}71.5 & \cellcolor{tabfirst}92.4 \\
\cellcolor{blue!10}DT &  & \cellcolor{orange!10}VT && 45.4 & 72.3 & 80.2 && \cellcolor{tabsecond}9.7 & \cellcolor{tabfirst}54.6 & \cellcolor{tabsecond}82.0 && \cellcolor{tabsecond}14.3 & \cellcolor{tabsecond}70.6 & \cellcolor{tabsecond}92.1 \\
 & \cellcolor{green!10}SIFT & \cellcolor{orange!10}VT && 46.6 & 72.8 & 80.5 && 9.2 & \cellcolor{tabfirst}54.6 & \cellcolor{tabsecond}82.0 && 12.5 & 70.0 & 92.0 \\
\cellcolor{blue!10}DT & \cellcolor{green!10}SIFT &  && \cellcolor{tabsecond}54.8 & \cellcolor{tabsecond}77.7 & \cellcolor{tabsecond}84.0 && 7.2 & 32.6 & 68.3 && 9.9 & 47.2 & 79.5 \\
\cellcolor{blue!10}DT &  &  && 47.0 & 72.7 & 80.4 && 5.5 & 30.7 & 66.0 && 7.7 & 41.0 & 75.7 \\
 & \cellcolor{green!10}SIFT &  && \cellcolor{tabfirst}55.8 & \cellcolor{tabfirst}78.4 & \cellcolor{tabfirst}84.7 && 8.2 & 34.2 & 68.7 && 4.6 & 31.7 & 71.4 \\

      \bottomrule
    \end{tabular}
    }
    \label{tbl:augmented_ba}
\end{table}

\subsection{Different Backbones}
End-to-end results with different feedforward backbones are shown in Table~\ref{tbl:ablation_backbone}.
Among the tested backbones, $\pi^3$ achieves the best overall performance, owing to its higher accuracy in local multi-view estimation.
Importantly, our proposed pipeline generalizes across all backbones: motion averaging (init) and A-BA each provide consistent improvements over their respective baselines, regardless of the underlying feedforward model.
This demonstrates that the gains from our global optimization are complementary to the local estimation quality.

\begin{table}[h]
    \centering
    \caption{Ablation on different choices of backbones.}
    \resizebox{\columnwidth}{!}{
    \setlength{\tabcolsep}{4pt} 
    \begin{tabular}{l l c c c l c c c l c c c 
    } \toprule 
 && \multicolumn{3}{c}{ETH3D} && \multicolumn{3}{c}{SMERF \footnotesize{(minimal)}} && \multicolumn{3}{c}{SMERF \footnotesize{(low)}}  \\
        \cmidrule{3-5} \cmidrule{7-9}  \cmidrule{11-13}
 \multicolumn{2}{c}{AUC@}  & 1 & 3 & 5 && 1 & 5 & 20 && 1 & 5 & 20 \\ 
 \midrule
$\pi^3$ & & 13.2 & 36.1 & 48.9 & & 3.2 & 18.0 & 51.7 & & 1.5 & 14.3 & 49.8 \\
VGGT & & 8.6 & 24.0 & 35.0 & & 2.8 & 6.0 & 22.1 & & 1.4 & 6.0 & 28.2 \\
MA & & 5.1 & 11.1 & 18.3 & & 3.0 & 14.5 & 42.7 & & 1.7 & 17.7 & 56.1 \\
\midrule
$\pi^3$ (init) & & 20.3 & 49.0 & 61.9 & & \cellcolor{tabsecond}9.8 & \cellcolor{tabfirst}55.5 & \cellcolor{tabfirst}82.3 & & \cellcolor{tabsecond}12.9 & \cellcolor{tabsecond}70.2 & \cellcolor{tabsecond}92.1 \\
VGGT (init) & & 15.8 & 38.3 & 50.5 & & 6.7 & 45.4 & 76.7 & & 11.3 & 59.9 & 87.3 \\
MA (init) & & 5.3 & 12.3 & 19.7 & & 4.2 & 33.7 & 70.5 & & 3.2 & 39.7 & 78.7 \\
\midrule
$\pi^3$ (A-BA) & & \cellcolor{tabsecond}53.1 & \cellcolor{tabfirst}77.0 & \cellcolor{tabfirst}83.7 & & \cellcolor{tabfirst}9.9 & \cellcolor{tabsecond}54.5 & \cellcolor{tabsecond}81.6 & & \cellcolor{tabfirst}14.9 & \cellcolor{tabfirst}71.9 & \cellcolor{tabfirst}92.5 \\
VGGT (A-BA) & & \cellcolor{tabfirst}53.7 & \cellcolor{tabsecond}76.9 & \cellcolor{tabsecond}83.3 & & 7.2 & 45.6 & 76.4 & & 12.0 & 60.2 & 87.2 \\
MA (A-BA) & & 17.5 & 45.6 & 59.3 & & 4.4 & 31.6 & 69.6 & & 3.4 & 39.3 & 78.3 \\
      \bottomrule
    \end{tabular}
    }
    \label{tbl:ablation_backbone}
\end{table}

\subsection{Covisibility Filtering}
An ablation of our two filtering steps can be found in Table~\ref{tbl:verification}, where \textit{DG} stands for Doppelgangers++ and \textit{VO} for visual overlap ratio. The results of all reported variants are after motion averaging.
On ETH3D, filtering with Doppelgangers++ is sufficient for achieving high accuracy, while for more complex scenes, such as SMERF, it is essential to use our proposed visual overlap ratio filtering and weighting in the optimization.

\begin{table}[h]
    \centering
    \caption{Ablation results for covisibility filtering.}
    \resizebox{\columnwidth}{!}{
    \setlength{\tabcolsep}{4pt} 
    \begin{tabular}{l l c c c l c c c  l c c c 
    } \toprule 
 && \multicolumn{3}{c}{ETH3D} && \multicolumn{3}{c}{SMERF \footnotesize{(minimal)}} && \multicolumn{3}{c}{SMERF \footnotesize{(low)}}  \\
        \cmidrule{3-5} \cmidrule{7-9}  \cmidrule{11-13}
 \multicolumn{2}{c}{AUC@}  & 1 & 3 & 5 && 1 & 5 & 20 && 1 & 5 & 20 \\ 
 \midrule
DG+VO & & \cellcolor{tabfirst}20.3 & \cellcolor{tabsecond}49.0 & \cellcolor{tabsecond}61.9 & & \cellcolor{tabfirst}9.8 & \cellcolor{tabfirst}55.5 & \cellcolor{tabfirst}82.4 & & \cellcolor{tabfirst}12.9 & \cellcolor{tabfirst}70.2 & \cellcolor{tabfirst}92.1 \\
DG & & \cellcolor{tabsecond}20.2 & \cellcolor{tabfirst}49.1 & \cellcolor{tabfirst}62.1 & & \cellcolor{tabsecond}9.4 & \cellcolor{tabsecond}52.7 & \cellcolor{tabsecond}78.1 & & \cellcolor{tabsecond}12.2 & \cellcolor{tabsecond}69.8 & \cellcolor{tabsecond}92.0 \\
VO & & 13.5 & 37.0 & 49.9 & & 3.2 & 17.2 & 46.8 & & 1.8 & 13.6 & 32.3 \\
      \bottomrule
    \end{tabular}
    }
    \label{tbl:verification}
\end{table}

\section{Alternative System Designs}
\subsection{Different Radius for Local Estimation}
The proposed method fixes the radius to 1 for local estimation, which maximizes the overlap between neighboring views within each star graph.
Increasing the radius is not straightforward: it would require a graph expansion step, and feedforward tracking only works when frames pairs have visual overlap.
Moreover, a larger radius increases computational cost and degrades local inference quality, as demonstrated in the main paper.
We therefore rely on classical optimization to propagate consistency over large distances in the graph.
For sequential inputs with high sampling rates or images captured from similar viewpoints, radius-1 stars may provide redundant coverage.
We leave the exploration of adaptive or larger graph radii as future work.

\subsection{Alternative Similarity Averaging Formulation}
The similarity averaging problem in Eq.~\ref{eq:sa} can also be reformulated as
\begin{equation}
    \min_{c, \tilde{s}} \sum_{l, (i, j)\in S_l} o_{ij}\cdot d\left(\tilde{s}_{l}\cdot R_{ij}^\top t_{ij} - (c_i-c_j)\right), \; \tilde{s}_0 = 1
\label{eq:sa_variant}
\end{equation}
This formulation results in a convex optimization problem when $d$ is convex.
However, the error can be large and scale variant, while Eq.~\ref{eq:sa} is more constrained when the scale of each star is normalized.
This is similar to the comparative advantage of BATA~\cite{zhuang2018baseline} over LUD~\cite{ozyesil2015robust} in translation averaging.

\section{Runtime}
Component-level runtime statistics are summarized in Table~\ref{tbl:runtime_batch} and Table~\ref{tbl:runtime_global}.
All experiments are conducted on an Neoverse-V2 CPU with 856\,GB RAM and an NVIDIA GH200 GPU with 96\,GB memory.
A \textit{batch} refers to an image pair or a star.
Because the number of retrieved pairs and the maximum number of neighbors per image are fixed, the runtime of Doppelgangers++~\cite{xiangli2025doppelgangers++} and local inference (star reconstruction and tracking) scales linearly with the number of images.
The global motion averaging and A-BA steps are comparatively inexpensive, especially when the number of images is small.
If lower accuracy is acceptable for a downstream application, tracking and A-BA can be skipped entirely, significantly reducing the overall computation.



\begin{table}[h]
    \centering
    \caption{Runtime (in seconds) of batch inference across datasets.}
    \resizebox{\columnwidth}{!}{
    \setlength{\tabcolsep}{3pt} 
    \begin{tabular}{l l c c l c c l c c 
    } \toprule 
&& \multicolumn{2}{c}{Two View~\cite{xiangli2025doppelgangers++}} && \multicolumn{2}{c}{Star~\cite{wang2025pi}} && \multicolumn{2}{c}{Tracking~\cite{wang2024vggsfm}} \\
\cmidrule{3-4} \cmidrule{6-7}  \cmidrule{9-10} 
Percentile @ && 50 & 90 && 50 & 90 && 50 & 90 \\
\midrule
Per batch && 1.20 & 1.23 && 0.31 & 0.77 && 1.27 & 1.81 \\
      \bottomrule
    \end{tabular}
    }
    \label{tbl:runtime_batch}
\end{table}

\begin{table}[h]
    \centering
    \caption{Runtime (in seconds) for the global refinement}
    \resizebox{\columnwidth}{!}{
    \begin{tabular}{l l c c l c c 
    } \toprule 
&& \multicolumn{2}{c}{Motion Averaging} && \multicolumn{2}{c}{Augmented Bundle Adjustment} \\
\cmidrule{3-4} \cmidrule{6-7}
Percentile @ && 50 & 90 && 50 & 90 \\
\midrule
ETH3D && 0.28 & 1.02 && 95.58 & 294.92 \\
CO3D && 0.09 & 0.42 && 8.58 & 45.92 \\
IMC2021 && 0.07 & 0.35 && 4.91 & 28.64 \\
SMERF && 1.18 & 3.13 && 255.37 & 1142.33 \\
LaMAR && 177.5 & 280.00 && 19505.91 & 22578.51 \\
      \bottomrule
    \end{tabular}
    }
    \label{tbl:runtime_global}
\end{table}

\section{Sampling Method for Analysis}

We consider all sequences in the LaMAR~\cite{sarlin2022lamar} dataset.
For each sequence, we select a random center frame every 200 images. Around each center frame, we extract subsequences at multiple temporal densities:
\begin{itemize}
    \item Consecutive sampling (high density): subsequences of length 4, 8, 16, 32, 64, and 128 frames.
    \item Sampling every 2 frames (medium density): subsequences of 4, 8, 16, 32, and 64 frames.
    \item Sampling every 4 frames (low density): subsequences of 4, 8, 16, and 32 frames.
\end{itemize}
For example, from a 64-frame window, sampling at intervals of 2 or 4 yields subsequences of length 32 or 16, respectively.
These subsequences observe the same scene but at different spatial-temporal densities.

The provided ground truth mesh is used to render depth maps, where sequences with inconsistent depths are discarded ($\approx 11.7\%$).
The remaining sequences form the complete sample set.

\section{Track Mixing Strategy}
The three source of tracks, namely SIFT, feedforward tracks, and virtual tracks, are combined through a priority-based mixing strategy before being passed to the final bundle adjustment.
The goal is to ensure that every image pair receives sufficient constraints while prioritizing SIFT tracks which have the highest accuracy and avoiding redundant tracks on pairs that are already well covered.

Concretely, we first include \emph{all} SIFT tracks unconditionally.
Next, we iterate over the deep tracks produced by the feedforward network.
For each deep track, we check whether any image pair it spans is undercovered, which we define as having fewer than 512 existing matches.
If at least one such under-covered pair exists, the deep track is added to the track set; otherwise it is discarded.
Finally, the same procedure is applied to virtual tracks.

\section{More Visualizations}

To demonstrate the concrete challenges faced by Structure-from-Motion, we provide further visual examples.

\begin{figure*}[t]
    \centering
    \resizebox{0.8\textwidth}{!}{
    \begin{tabular}{c l c c c 
    } 
    \centering
    \rotatebox{90}{~~exhibition\_hall} & &
    \includegraphics[height=0.4\columnwidth]{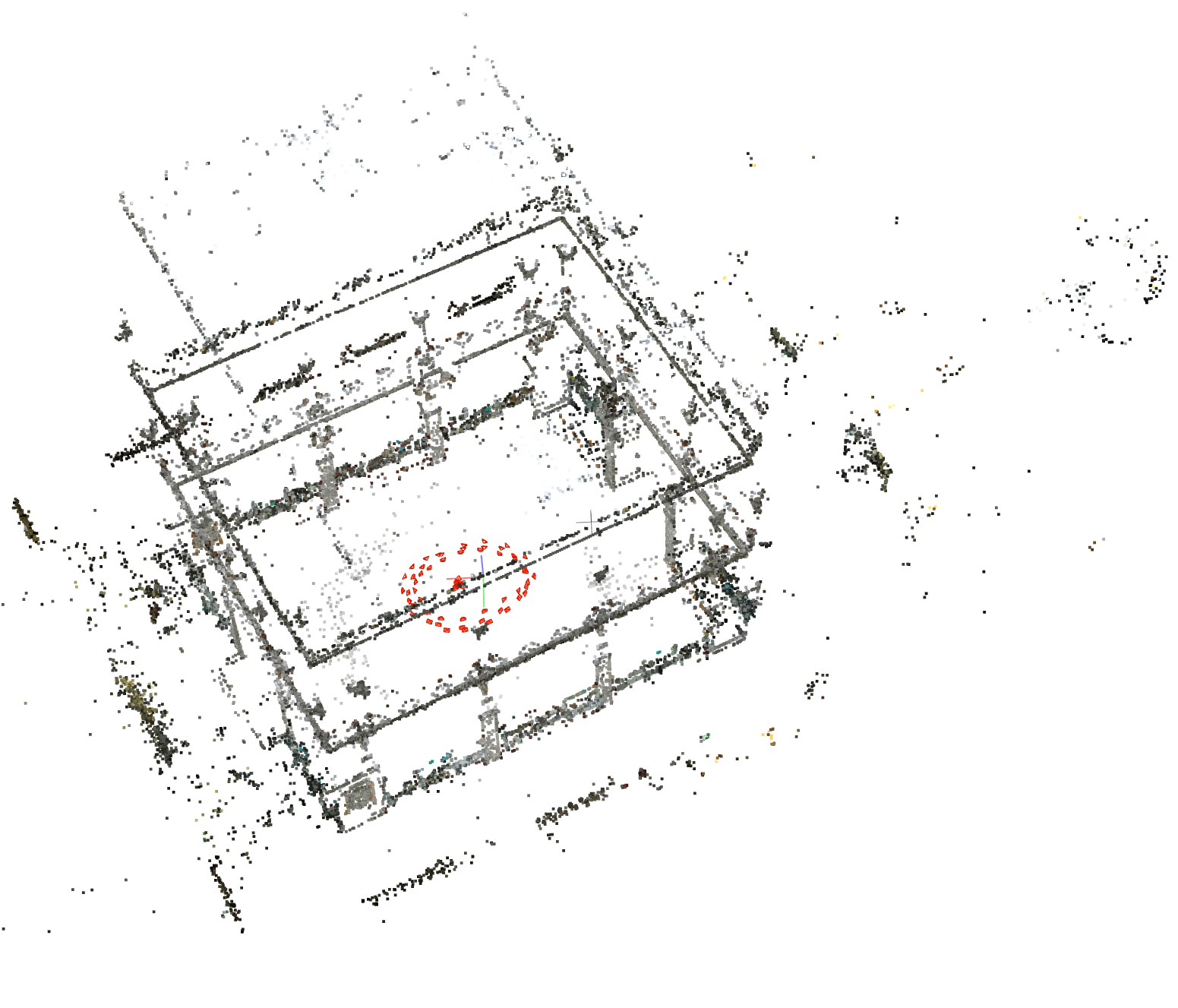} &
    \includegraphics[height=0.4\columnwidth]{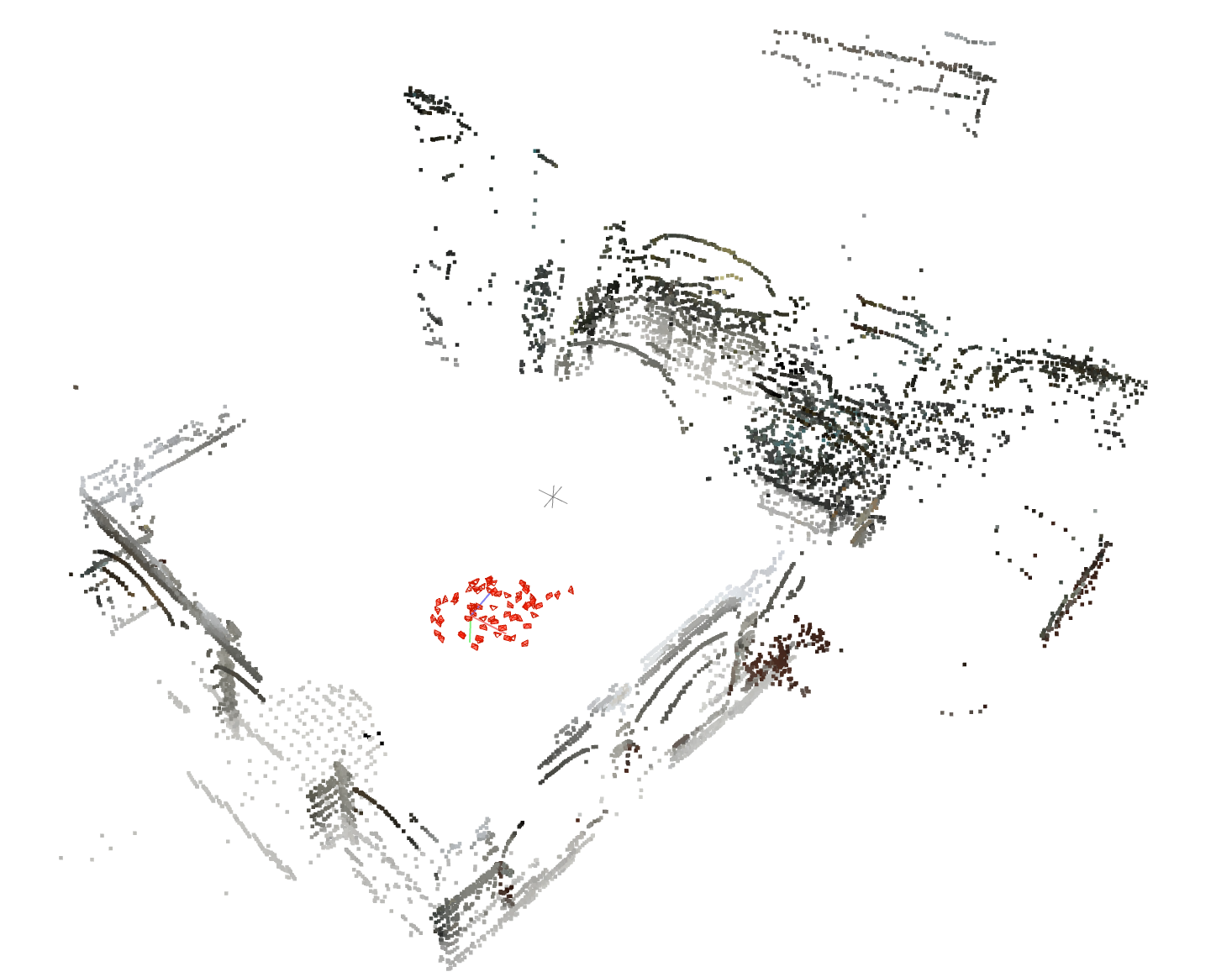} &
    \includegraphics[height=0.4\columnwidth]{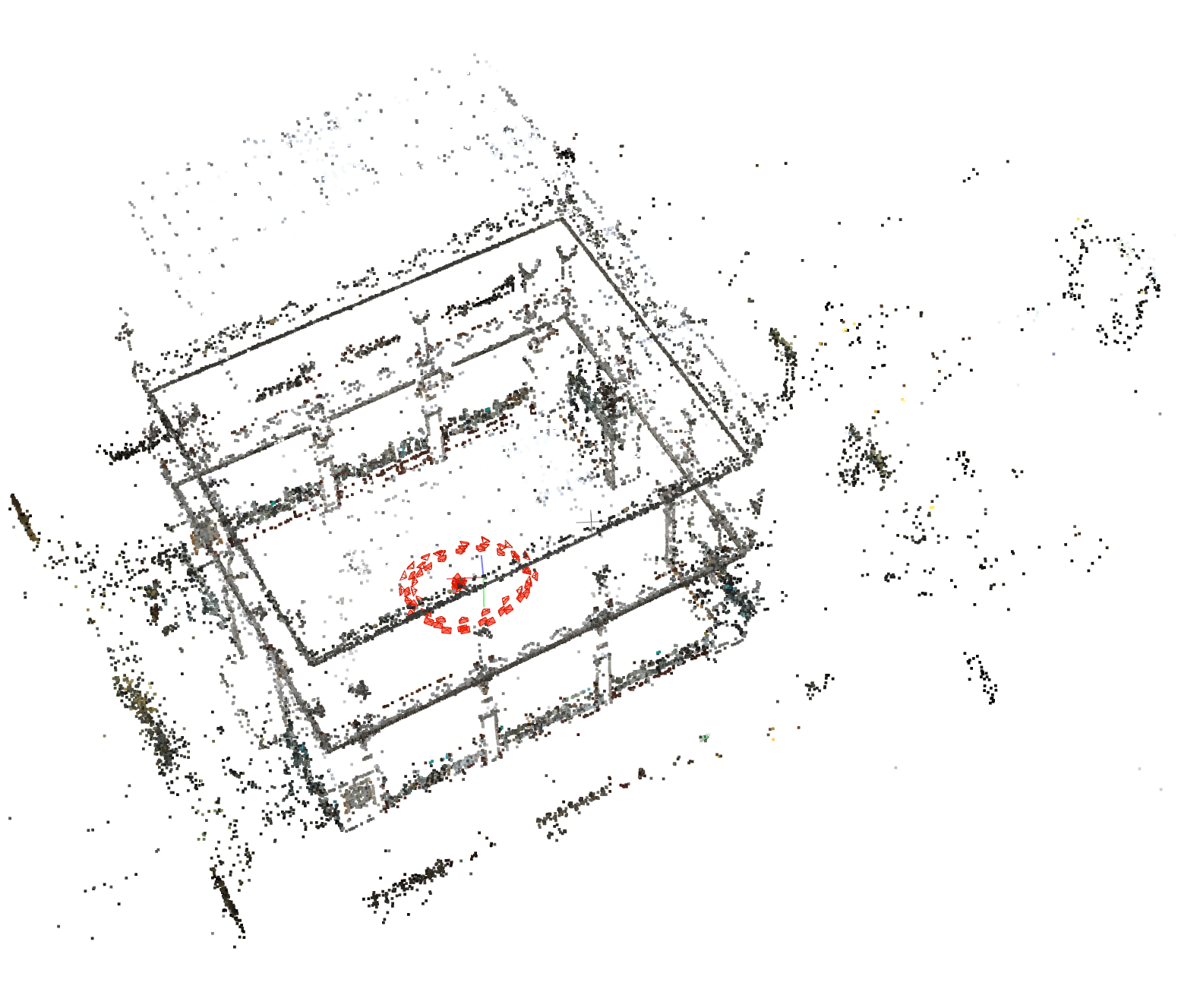} \\
    \rotatebox{90}{~~~~~~~~~~bridge} & &
    \includegraphics[height=0.4\columnwidth]{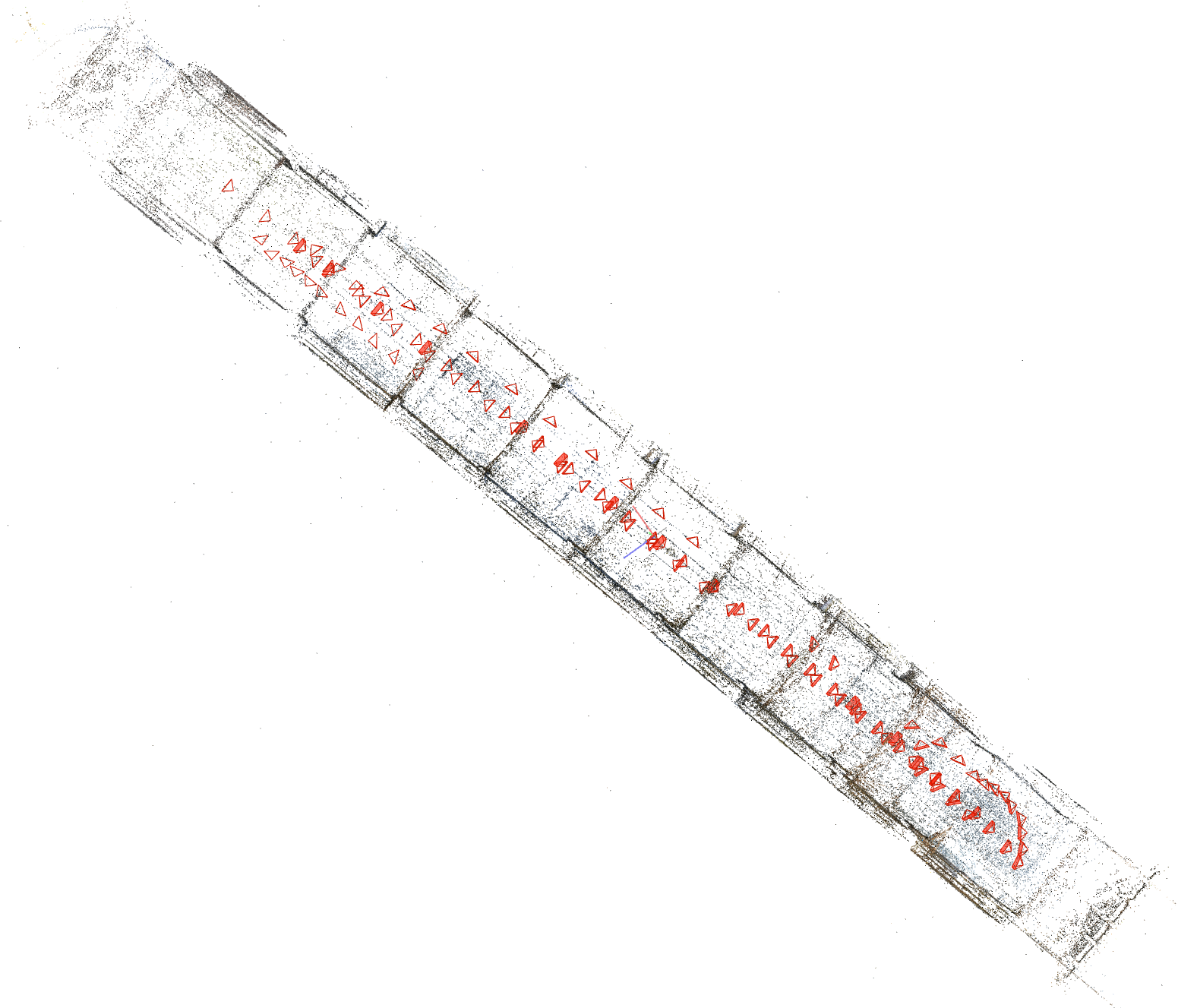} &
    \includegraphics[height=0.4\columnwidth]{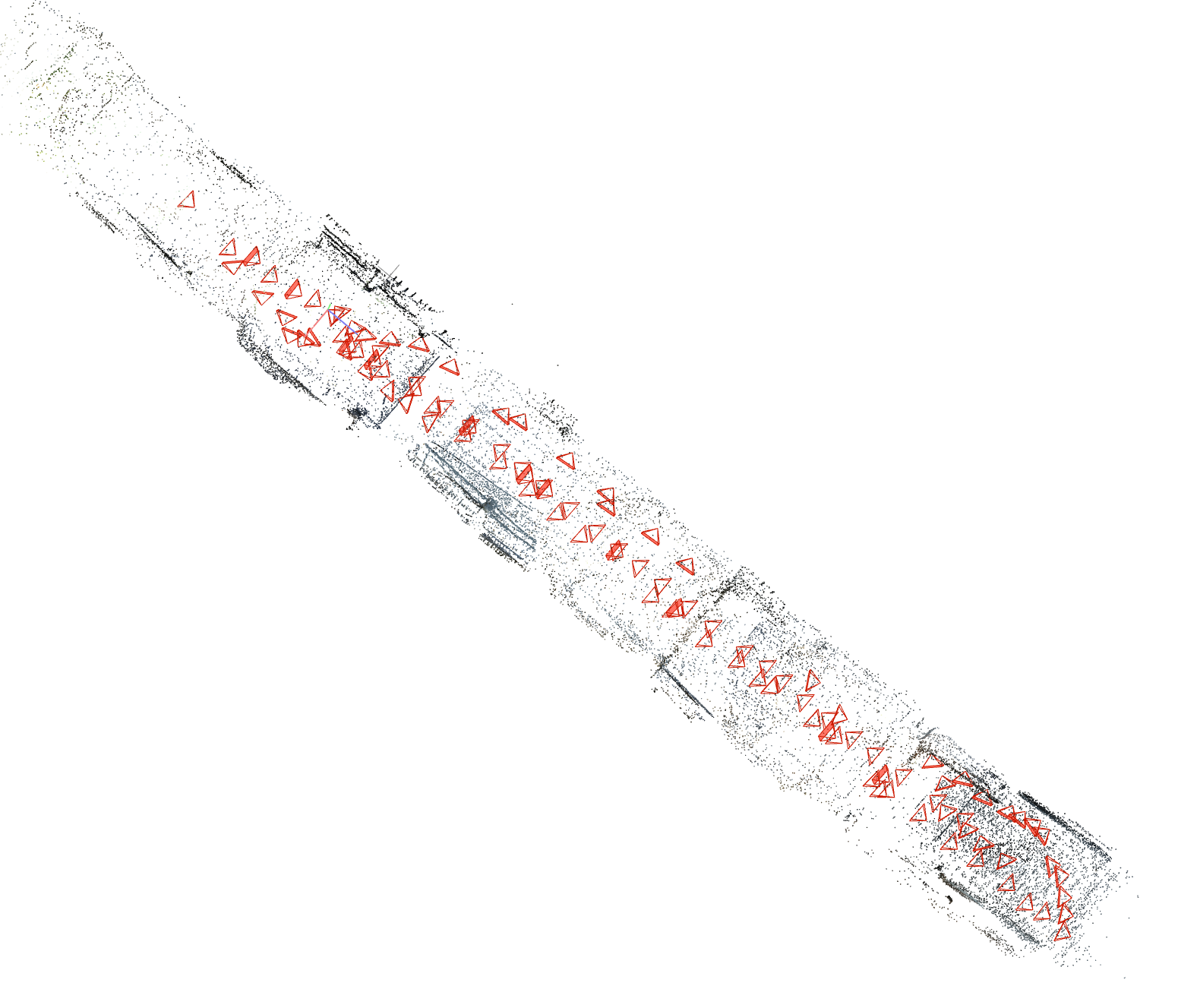} &
    \includegraphics[height=0.4\columnwidth]{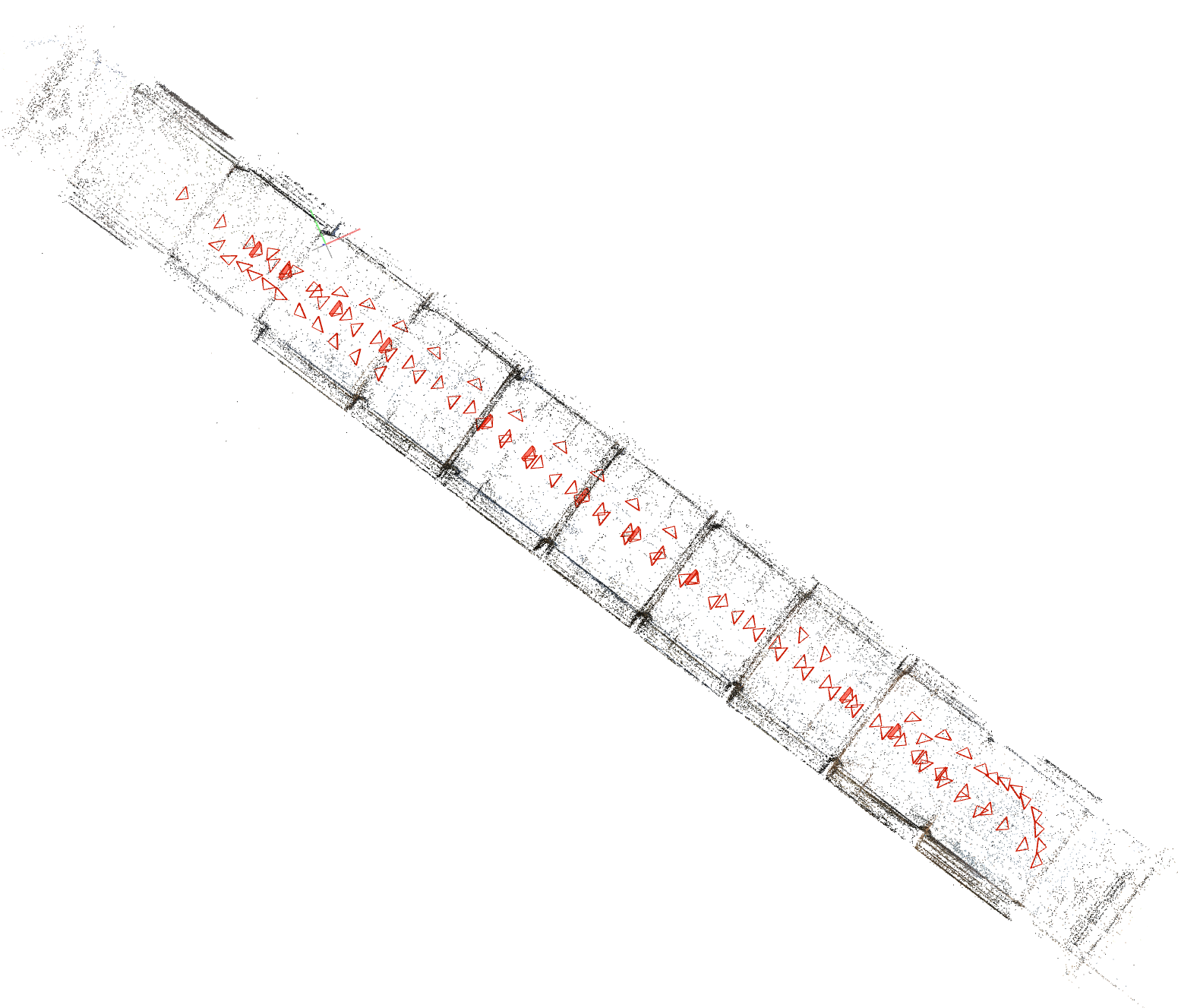} \\
    && Ours & $\pi^3$ & Ground Truth
\end{tabular}
}
    \caption{
    With the help of Doppelganger++~\cite{xiangli2025doppelgangers++}, our proposed method works well on scenes with high symmetry in ETH3D~\cite{schops2017multi}. The result of feedforward methods like $\pi^3$~\cite{wang2025pi} collapses.
    }
\label{fig:doppelgangers}
\end{figure*}

In symmetric scenes, feedforward methods often have difficulty distinguishing visually similar structures, resulting in collapsed reconstructions where distinct parts of the scene are incorrectly merged.
One such example with four-way symmetry can be found in Figure~\ref{fig:doppelgangers}.
By leveraging Doppelgangers++~\cite{xiangli2025doppelgangers++} to filter non-covisible pairs, our proposed method can reliably distinguish between the different facades and produce a correct reconstruction.

Examples from the SMERF~\cite{duckworth2024smerf} benchmark as proposed by MP-SFM~\cite{pataki2025mp} can be found in Figure~\ref{fig:low_overlap1} and Figure~\ref{fig:low_overlap2}.
This benchmark is established by selecting a sparse subset of images from dense indoor captures, resulting in minimal to no multi-view overlap between consecutive frames, often compounded by low texture.
The multi-room layout further introduces symmetry challenges, as structurally similar rooms can easily be confused.
Fueled by feedforward methods for local reconstructions, our proposed method successfully reconstructs scenes even under minimal overlap, while maintaining high accuracy when overlap is abundant.
In contrast, purely feedforward models struggle to distinguish visually similar rooms (\eg, Berlin and London) and exhibit drastic scale drift across rooms.
Furthermore, under low overlap, $\pi^3$~\cite{wang2025pi} lacks sufficient multi-view correspondences for bundle adjustment to be effective; in these cases, BA may fail to improve accuracy or even degrade it by overfitting to noisy observations.
Consequently, $\pi^3$ achieves low scores even on scenes with high image density.

\begin{figure*}[t]
    \centering
    \resizebox{\textwidth}{!}{
    \begin{tabular}{c c c c c 
    } 
    \centering
     & Minimal & Low & Medium & High \\
    \rotatebox{90}{~~~~~~~Alameda (Ours)} &
    \includegraphics[height=0.36\columnwidth]{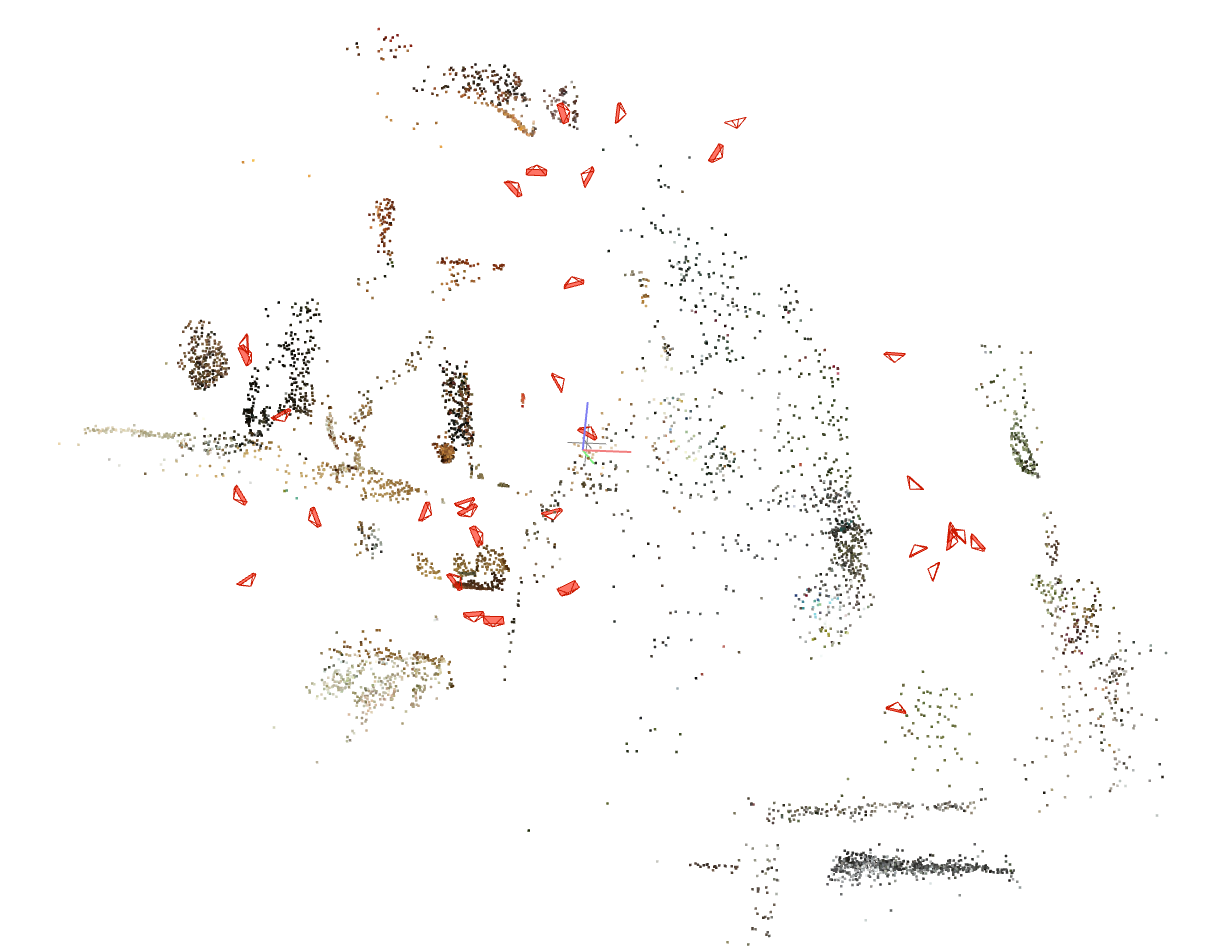} & 
    \includegraphics[height=0.36\columnwidth]{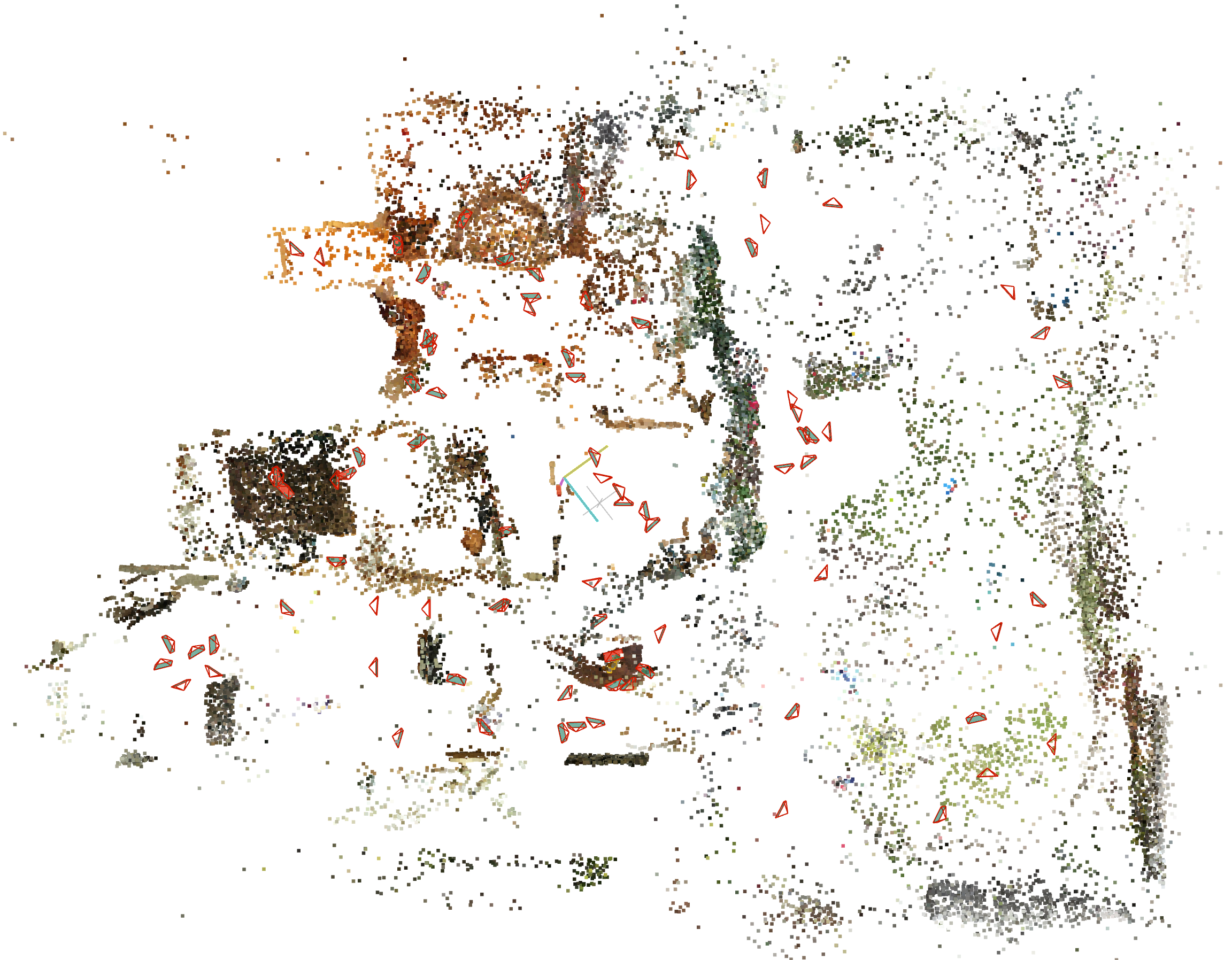} & 
    \includegraphics[height=0.36\columnwidth]{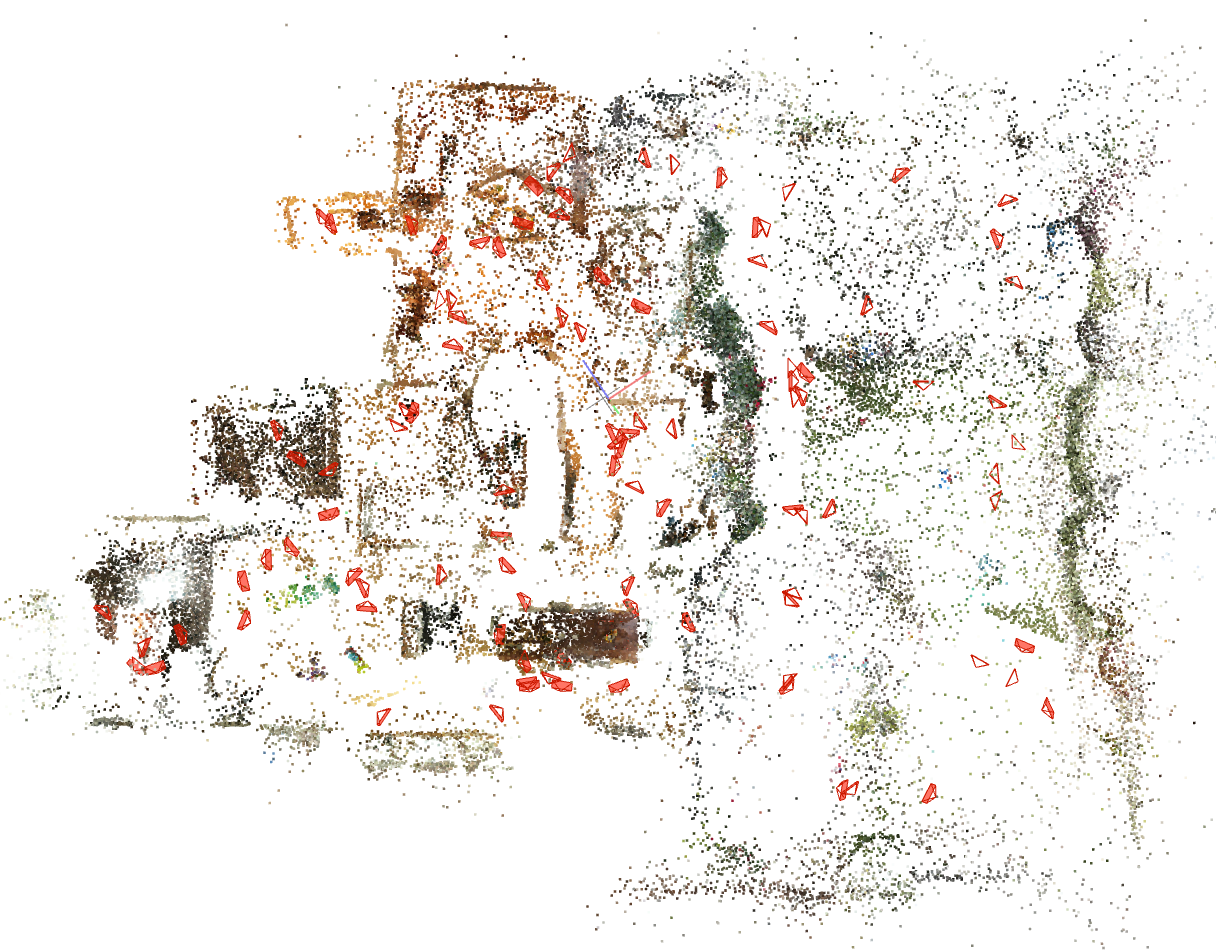} &
    \includegraphics[height=0.36\columnwidth]{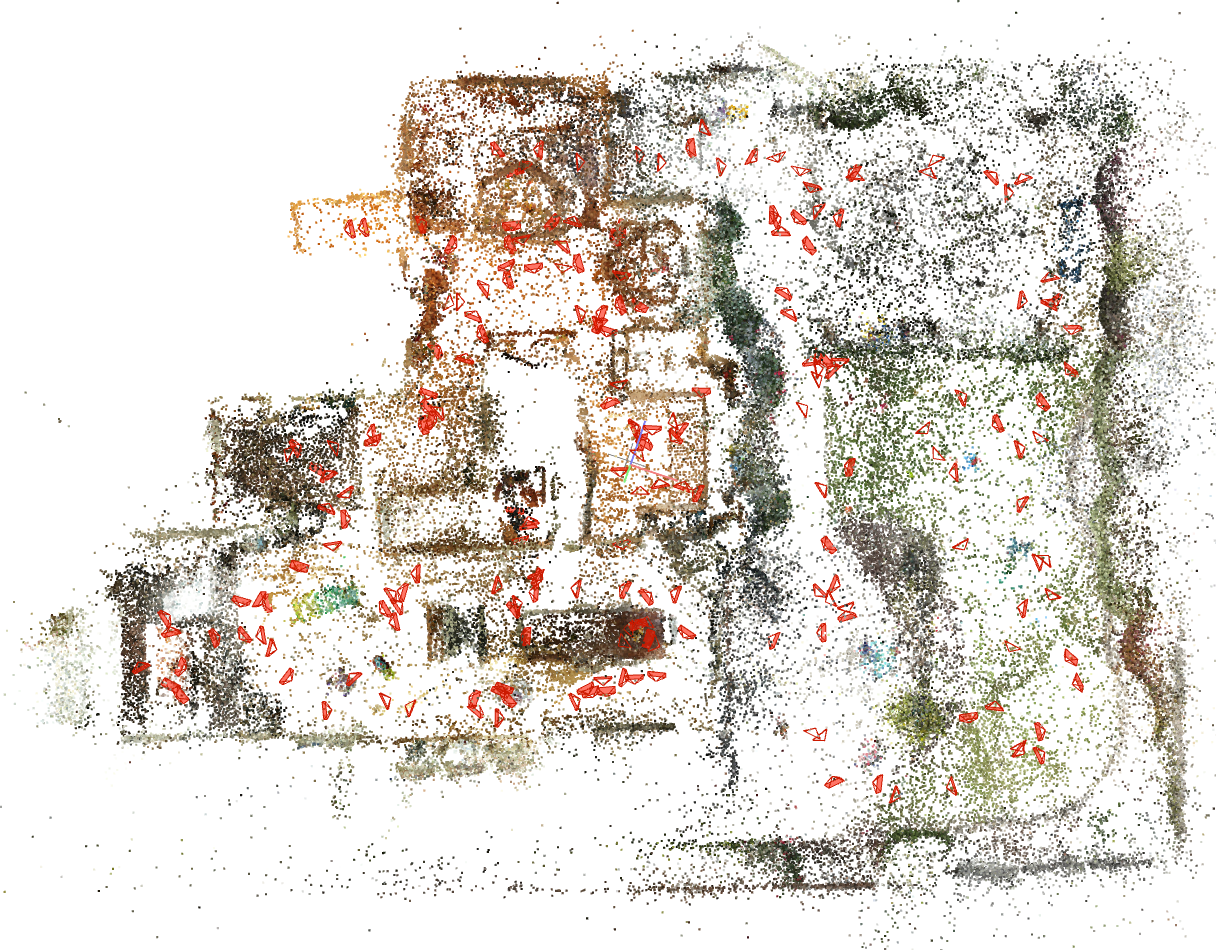} \\
    & 14.0/75.5/93.9 & 26.3/83.2/95.8 & 46.1/88.0/97.0 & 58.8/91.4/97.9 \\ 
    \rotatebox{90}{~~~~~~~~~Alameda ($\pi^3$)} &
    \includegraphics[height=0.36\columnwidth]{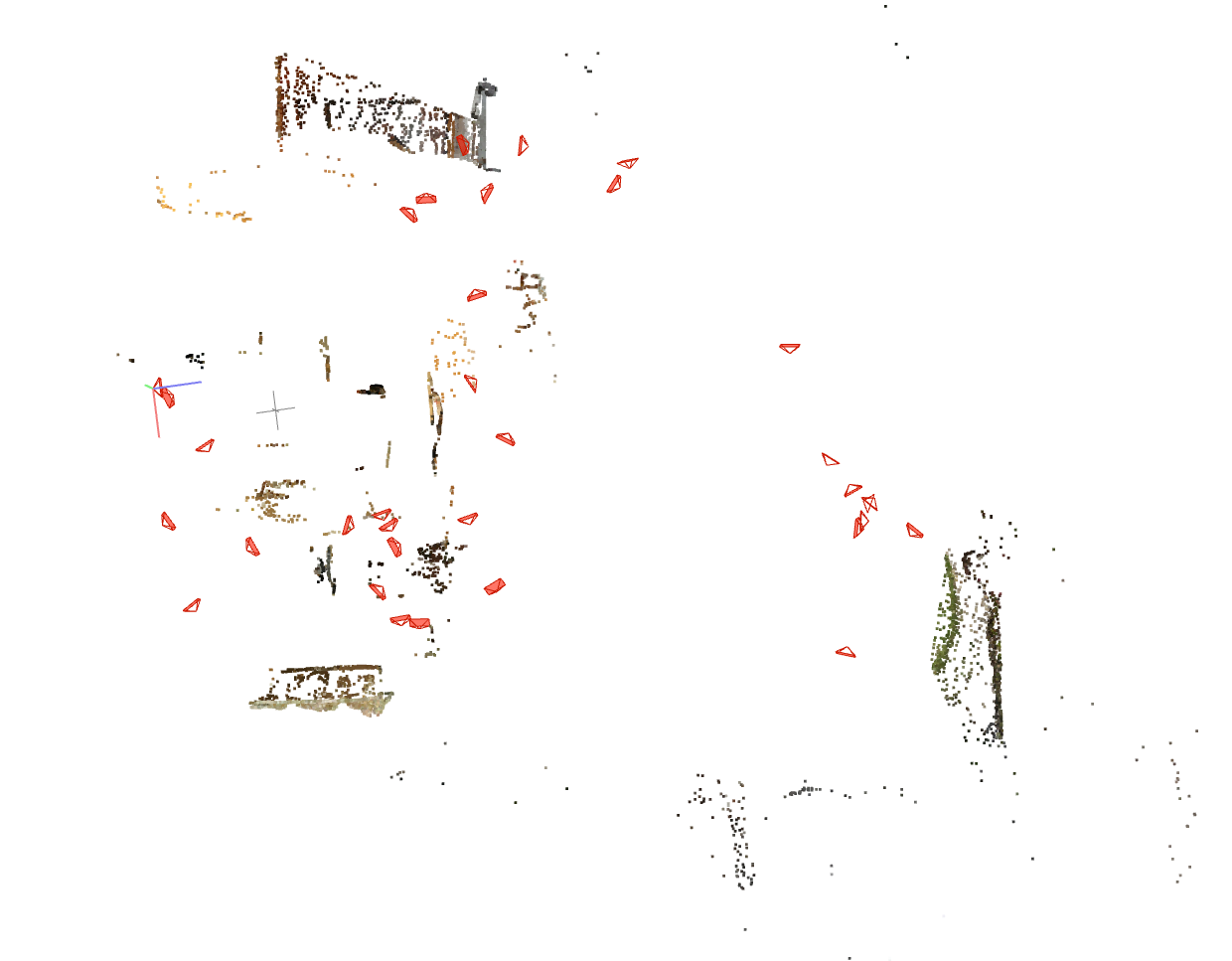} & 
    \includegraphics[height=0.36\columnwidth]{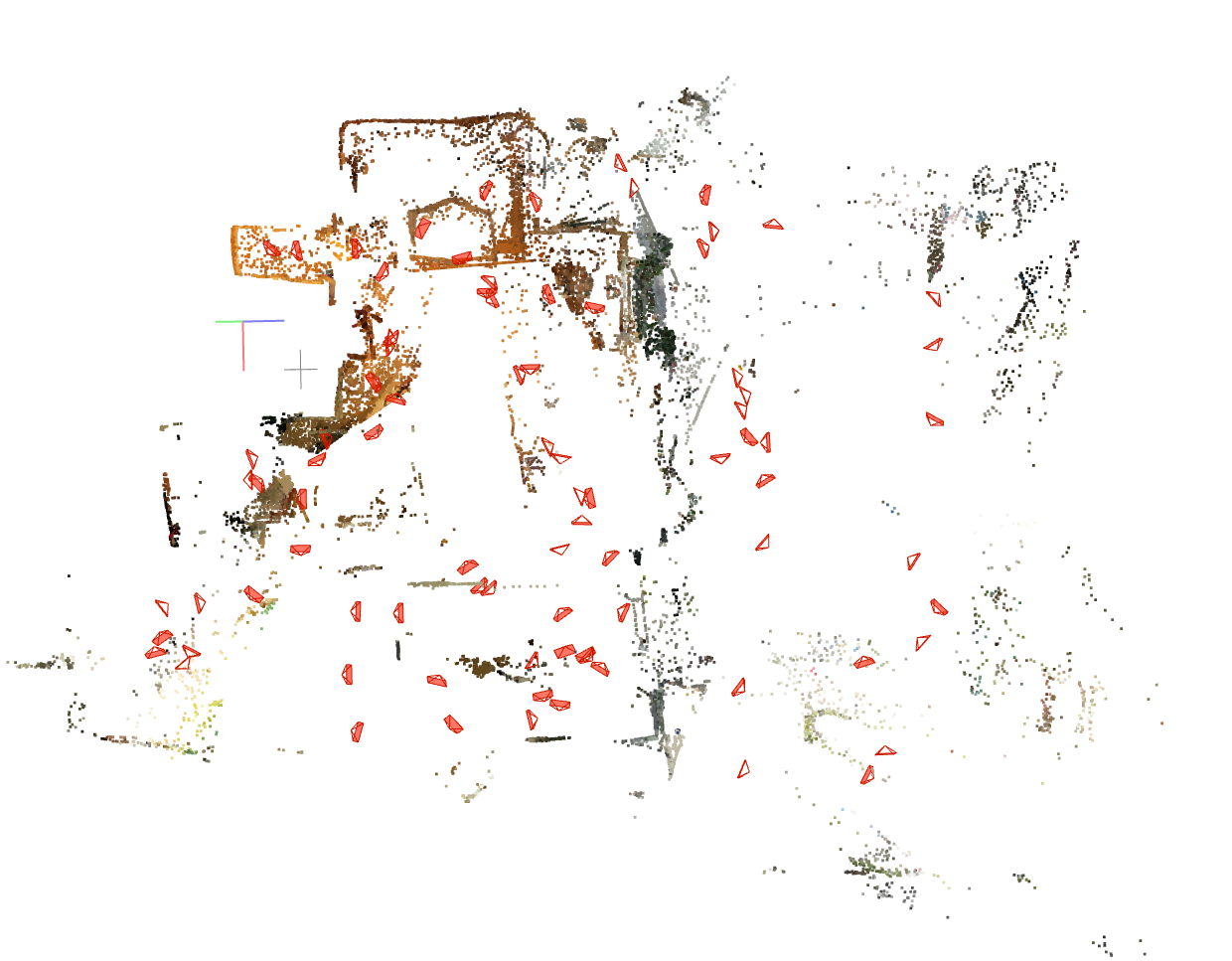} & 
    \includegraphics[height=0.36\columnwidth]{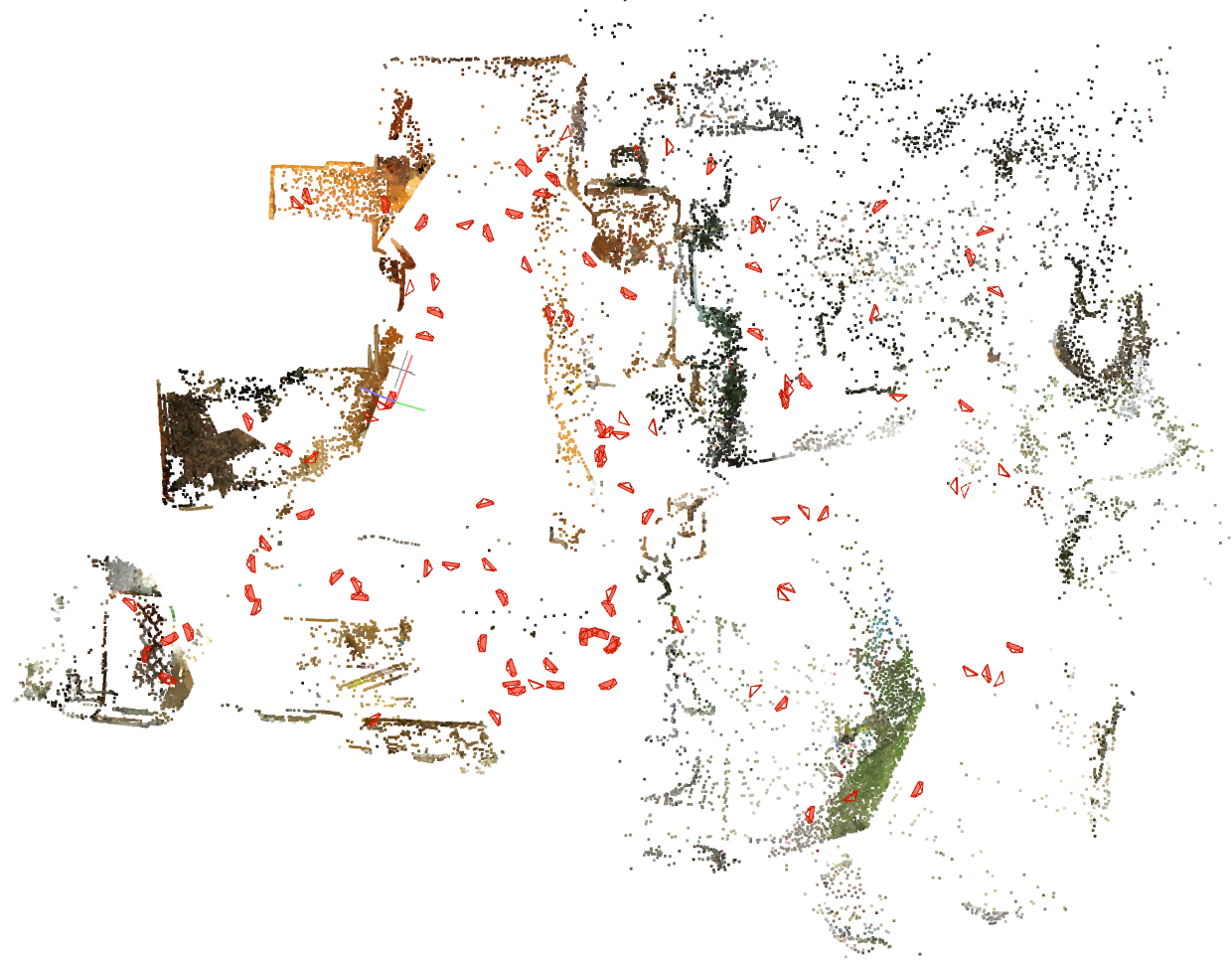} &
    \includegraphics[height=0.36\columnwidth]{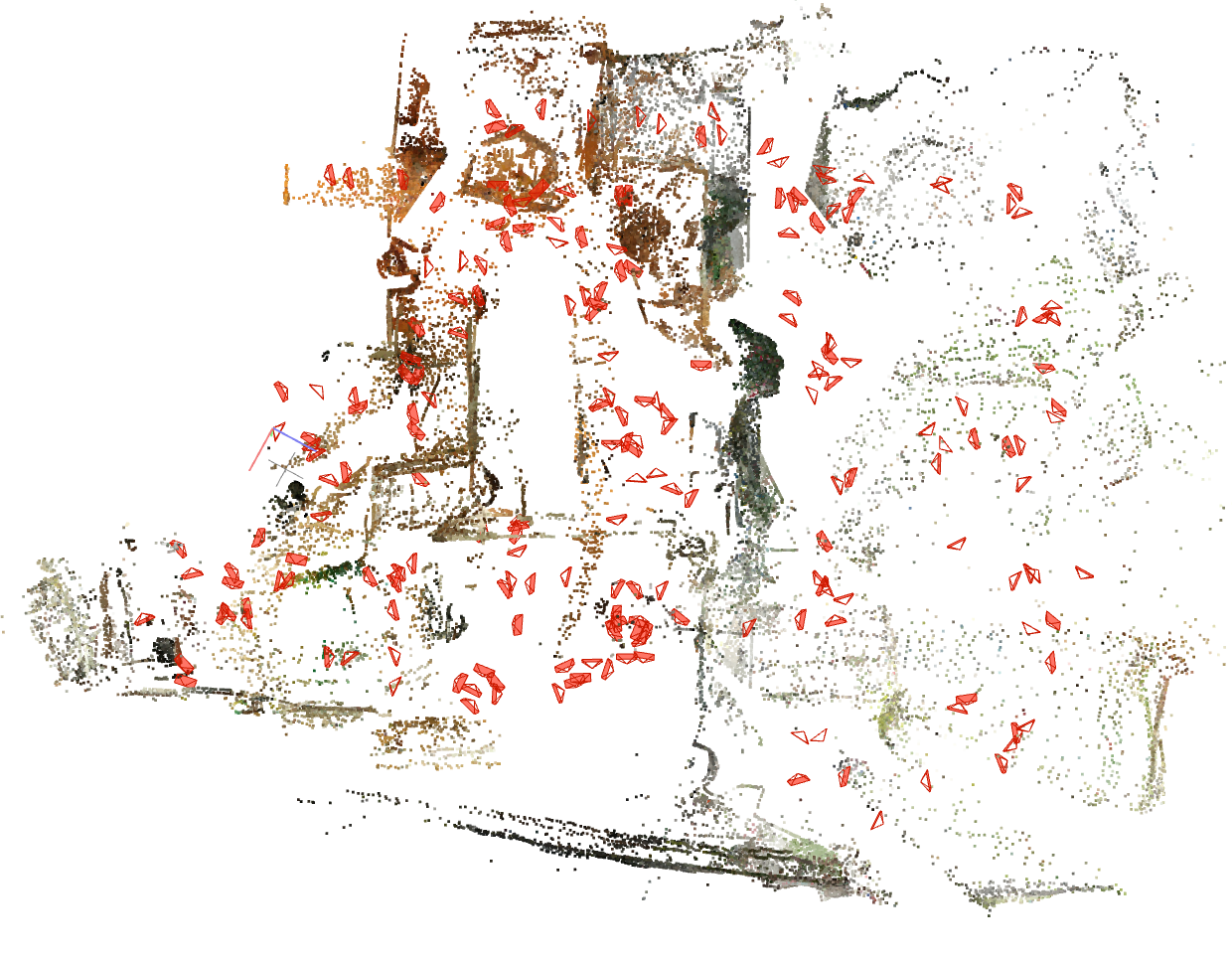} \\
     & 3.6/31.9/77.4 & 1.3/19.9/67.5 & 1.2/31.2/77.5 & 0.7/22.8/70.1 \\

    \rotatebox{90}{~~~~~~~Berlin (Ours)} &
    \includegraphics[height=0.36\columnwidth]{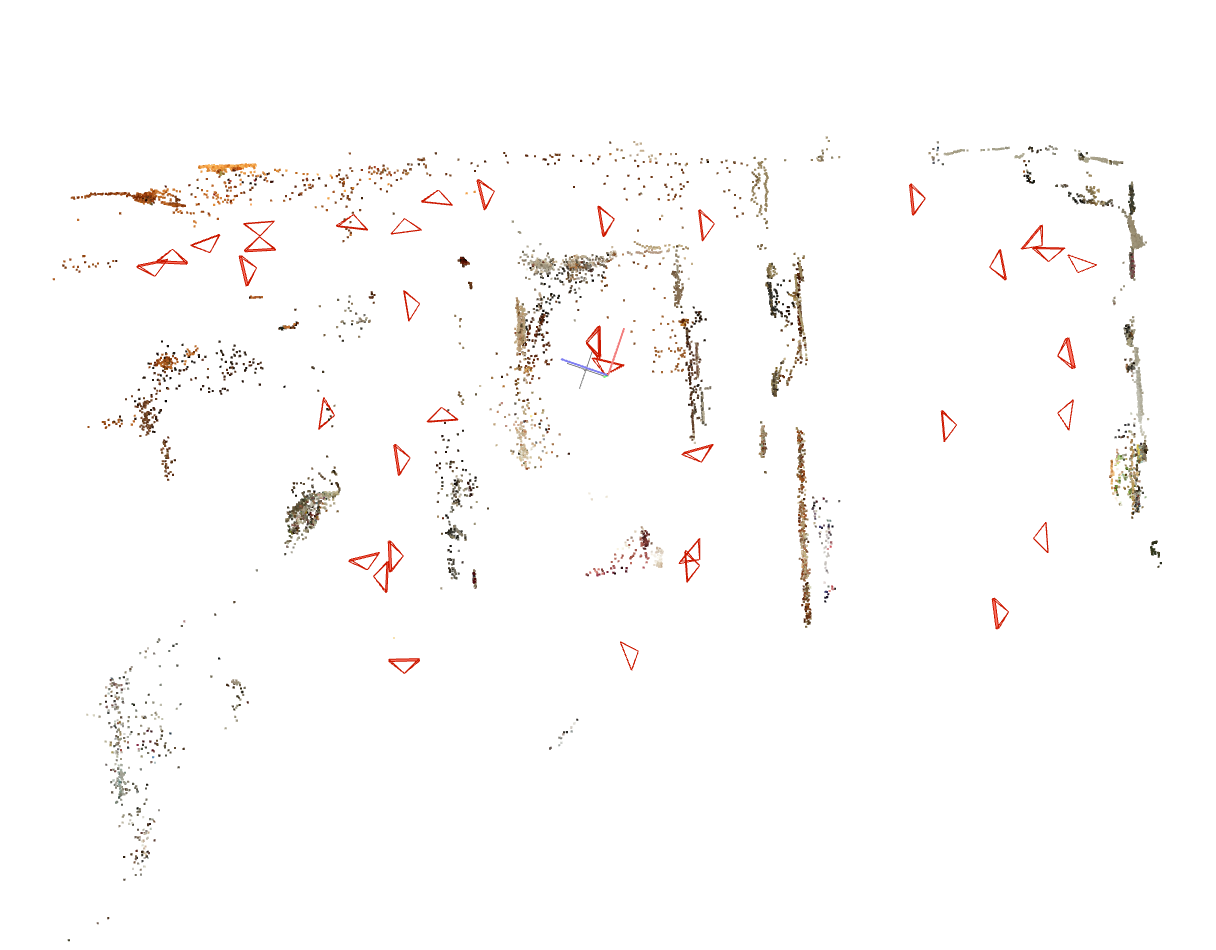} & 
    \includegraphics[height=0.36\columnwidth]{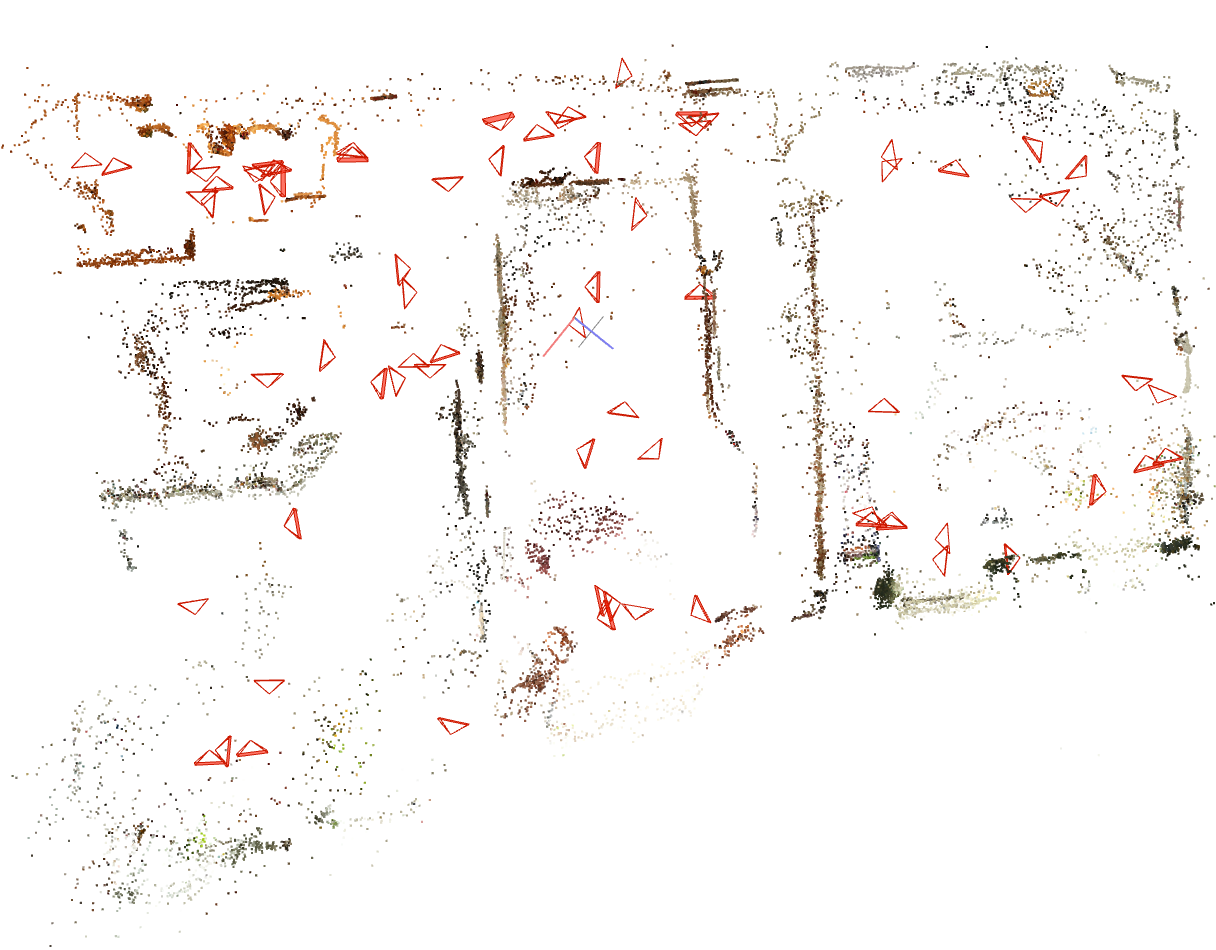} & 
    \includegraphics[height=0.36\columnwidth]{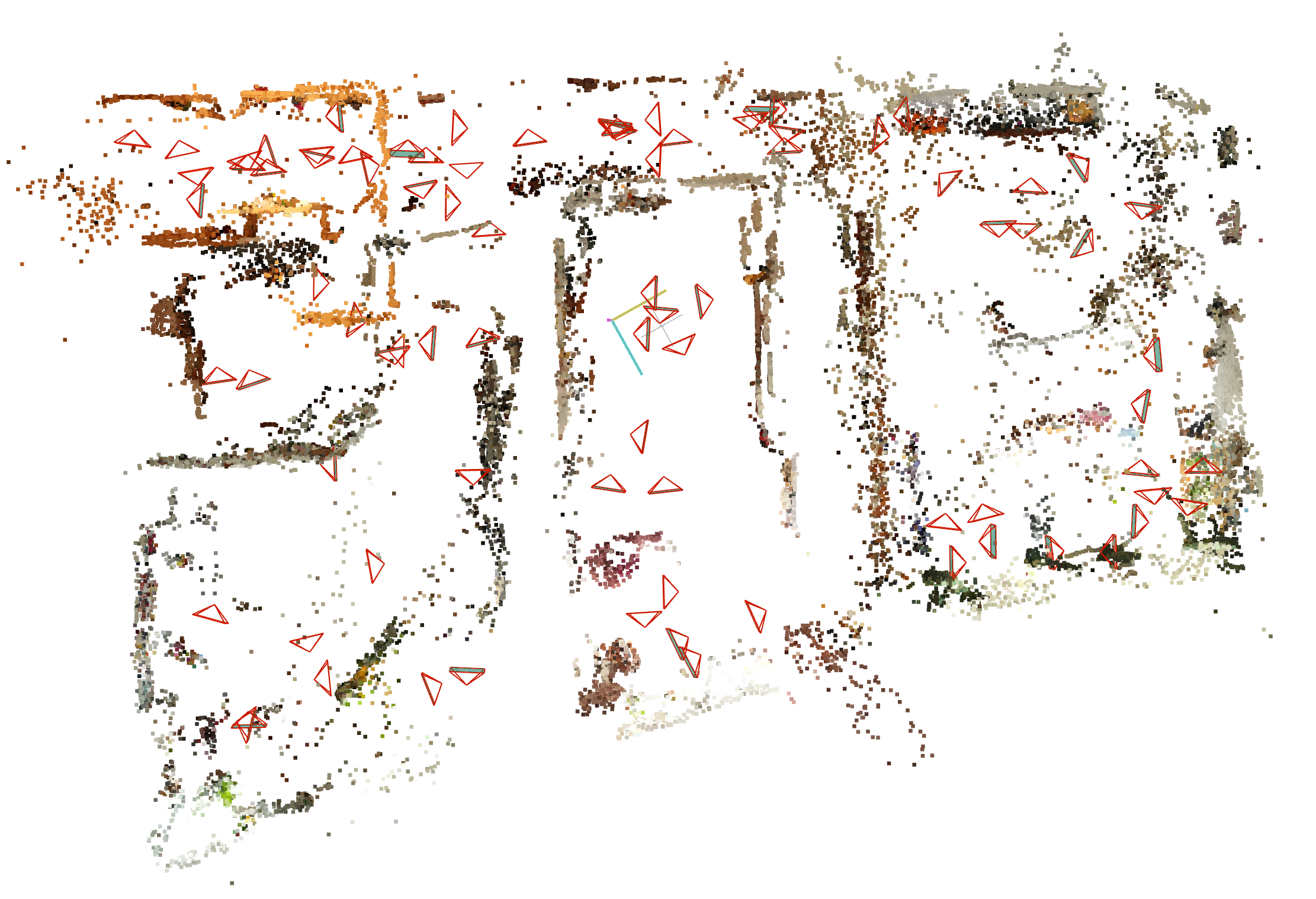} & 
    \includegraphics[height=0.36\columnwidth]{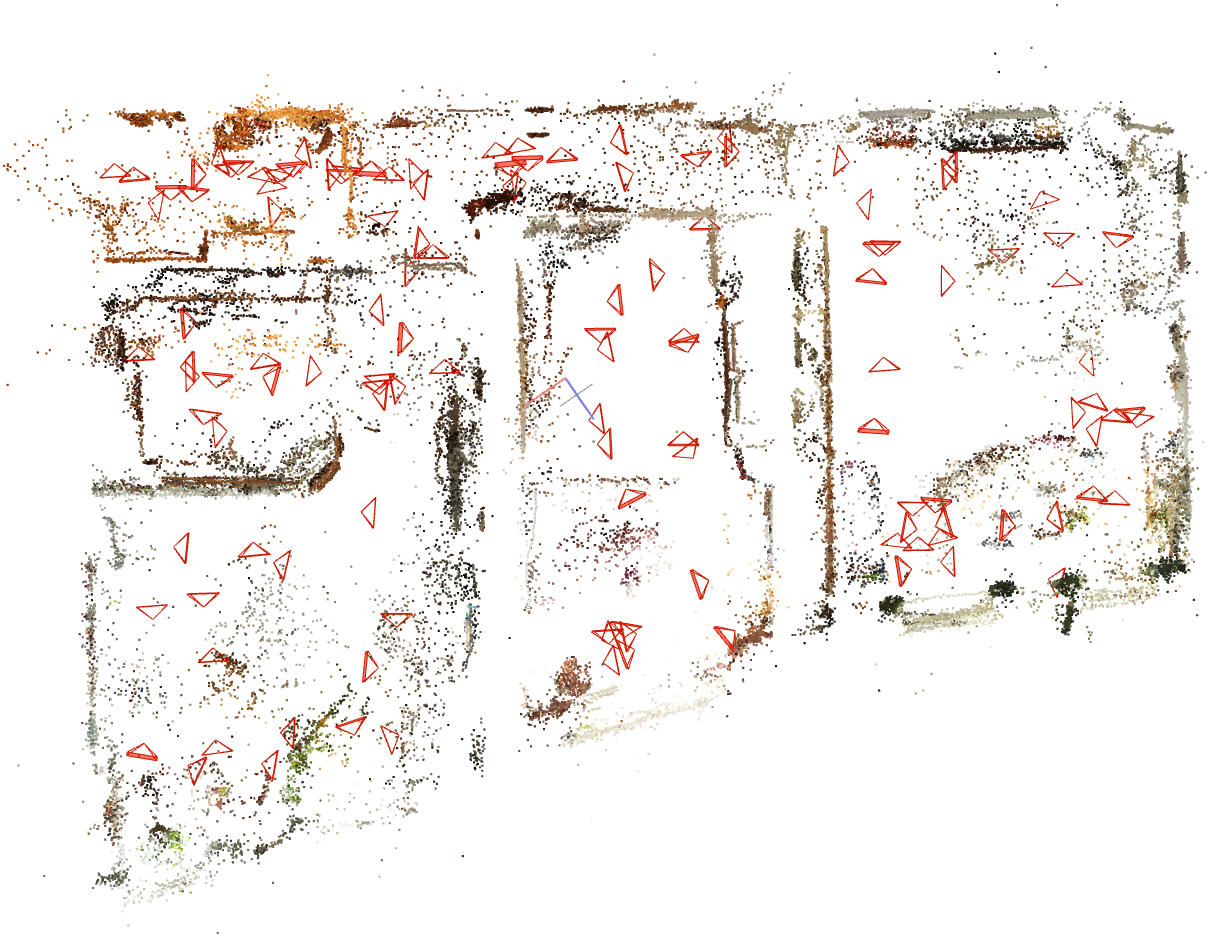} \\
    & 7.5/49.7/85.3 & 6.6/62.8/90.4 & 14.4/67.6/91.7 & 28.3/83.2/95.8 \\

    \rotatebox{90}{~~~~~~~~~Berlin ($\pi^3$)} &
    \includegraphics[height=0.36\columnwidth]{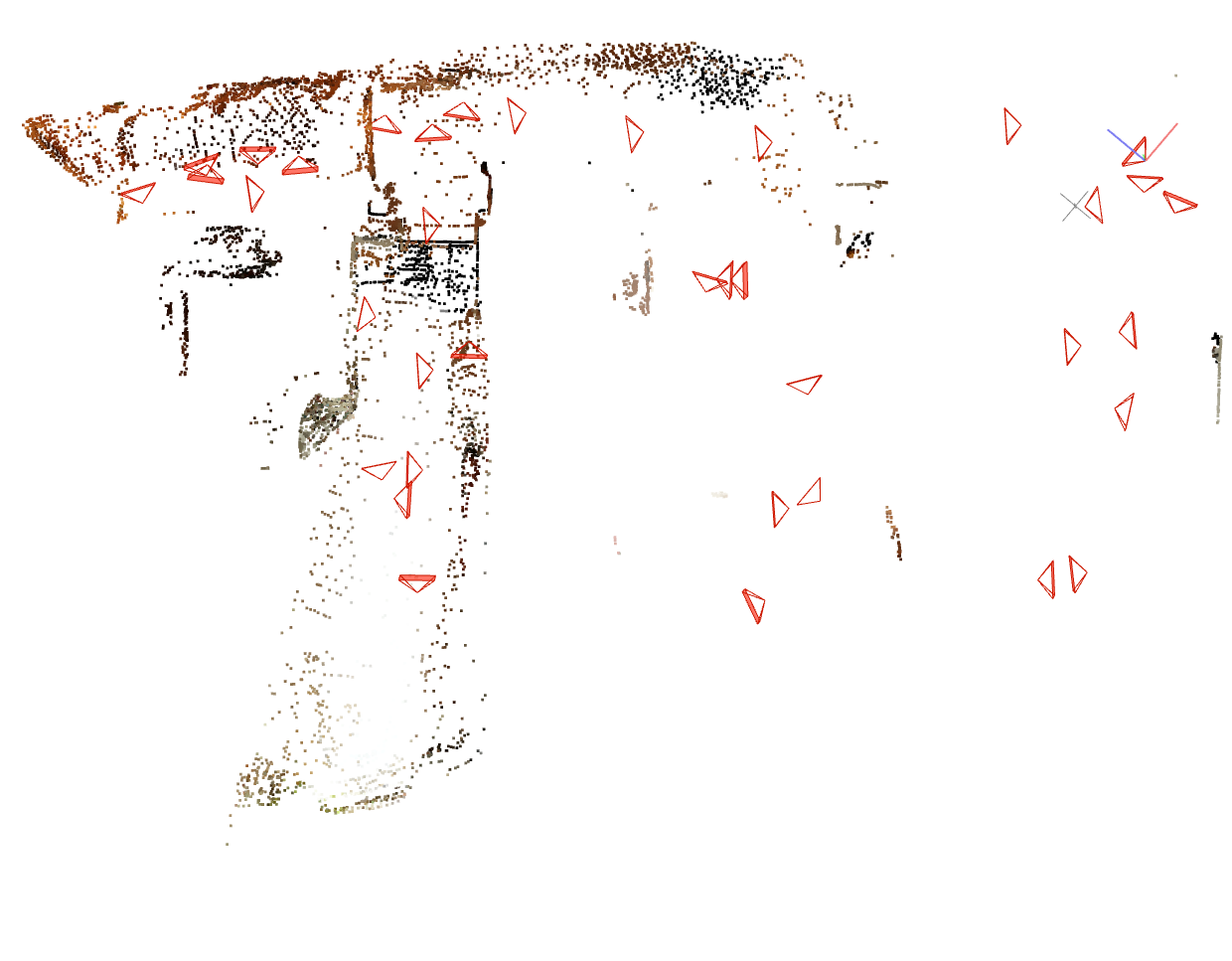} & 
    \includegraphics[height=0.36\columnwidth]{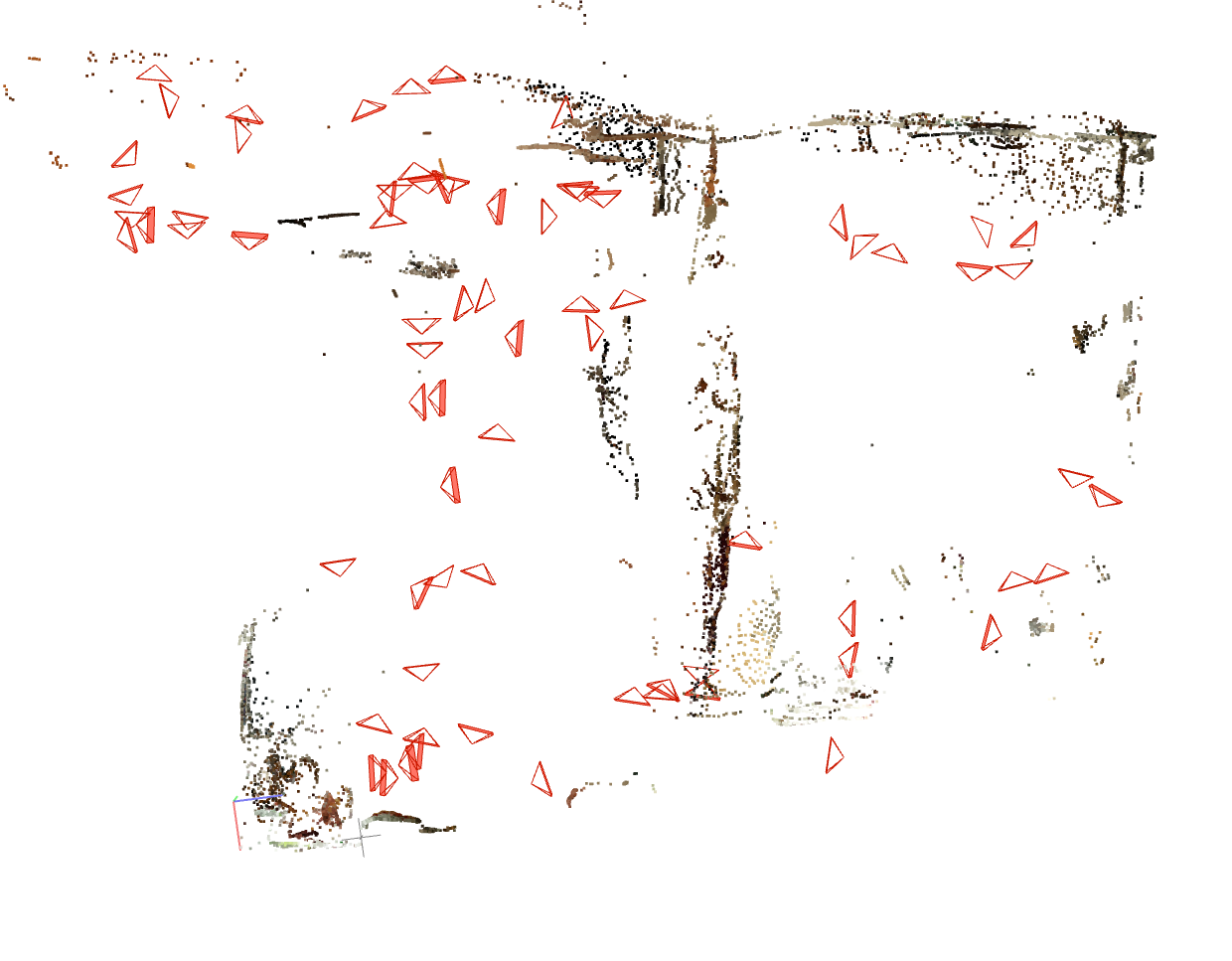} & 
    \includegraphics[height=0.36\columnwidth]{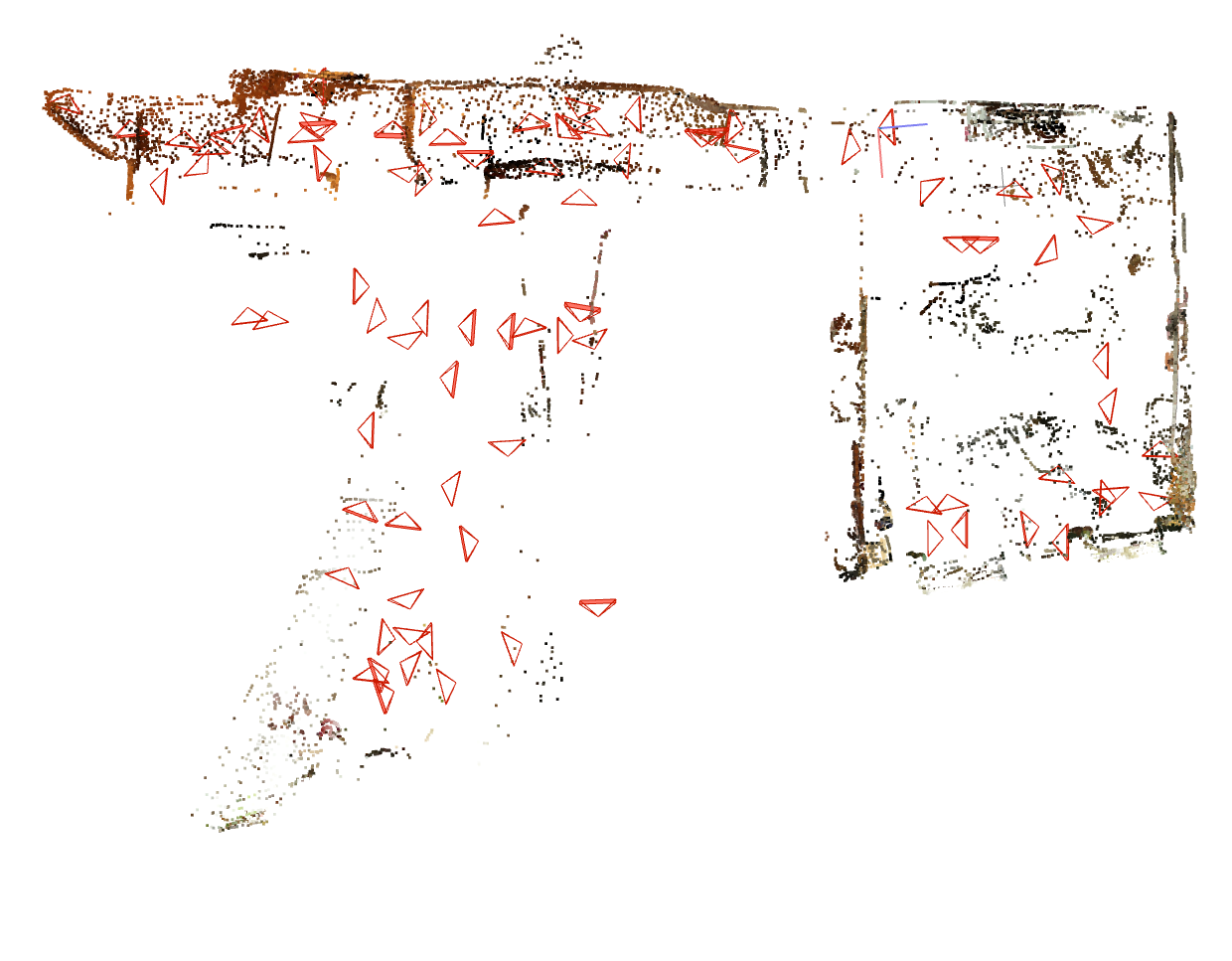} & 
    \includegraphics[height=0.36\columnwidth]{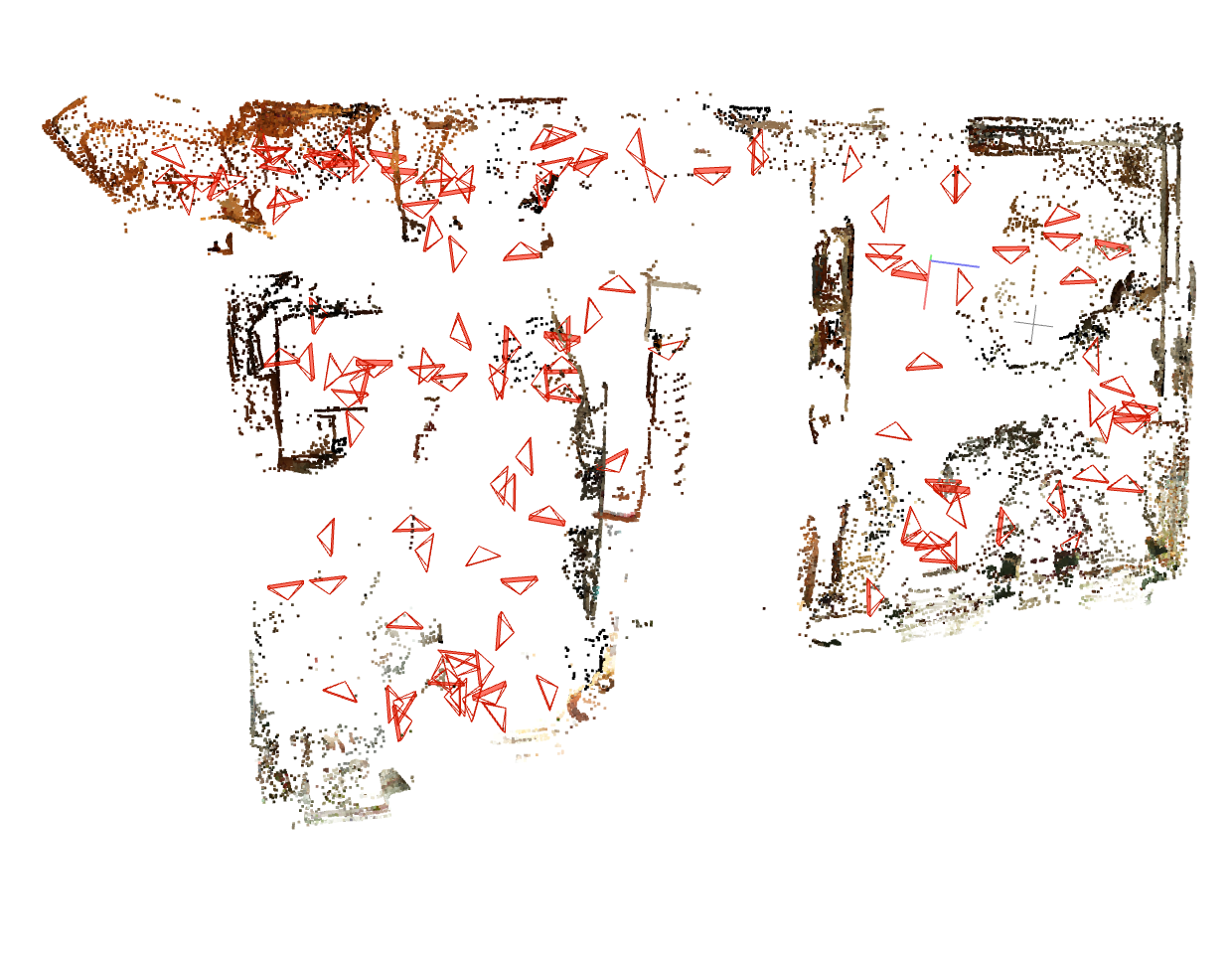} \\
     & 2.9/21.9/69.3 & 1.4/4.3/30.1 & 1.4/14.2/49.7 & 0.9/11.4/48.9\\

\end{tabular}
}
    \caption{
    Our proposed method works well from minimal to high overlap on SMERF~\cite{duckworth2024smerf} using the benchmark setup by MP-SFM~\cite{pataki2025mp}. The AUC scores at 1/5/20 degrees are shown below the image. Our proposed method works well even under minimal overlap, while $\pi^3$~\cite{wang2025pi} has difficulty in generating accurate reconstructions.
    }
\label{fig:low_overlap1}
\end{figure*}

\begin{figure*}[t]
    \centering
    \resizebox{\textwidth}{!}{
    \begin{tabular}{c c c c c 
    } 
    \centering
    \rotatebox{90}{~~~~~~~~London (Ours)} &
    \includegraphics[height=0.36\columnwidth]{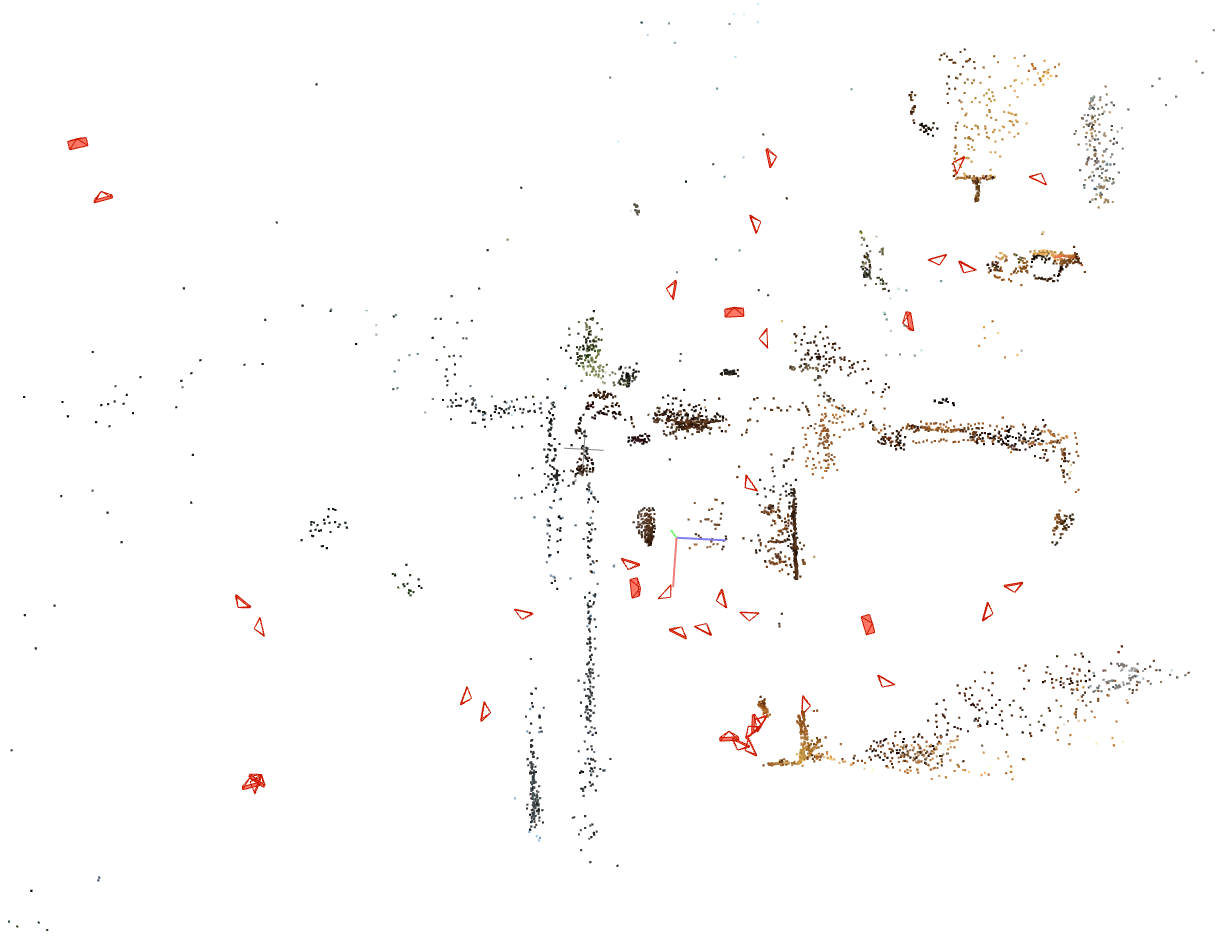} &
    \includegraphics[height=0.36\columnwidth]{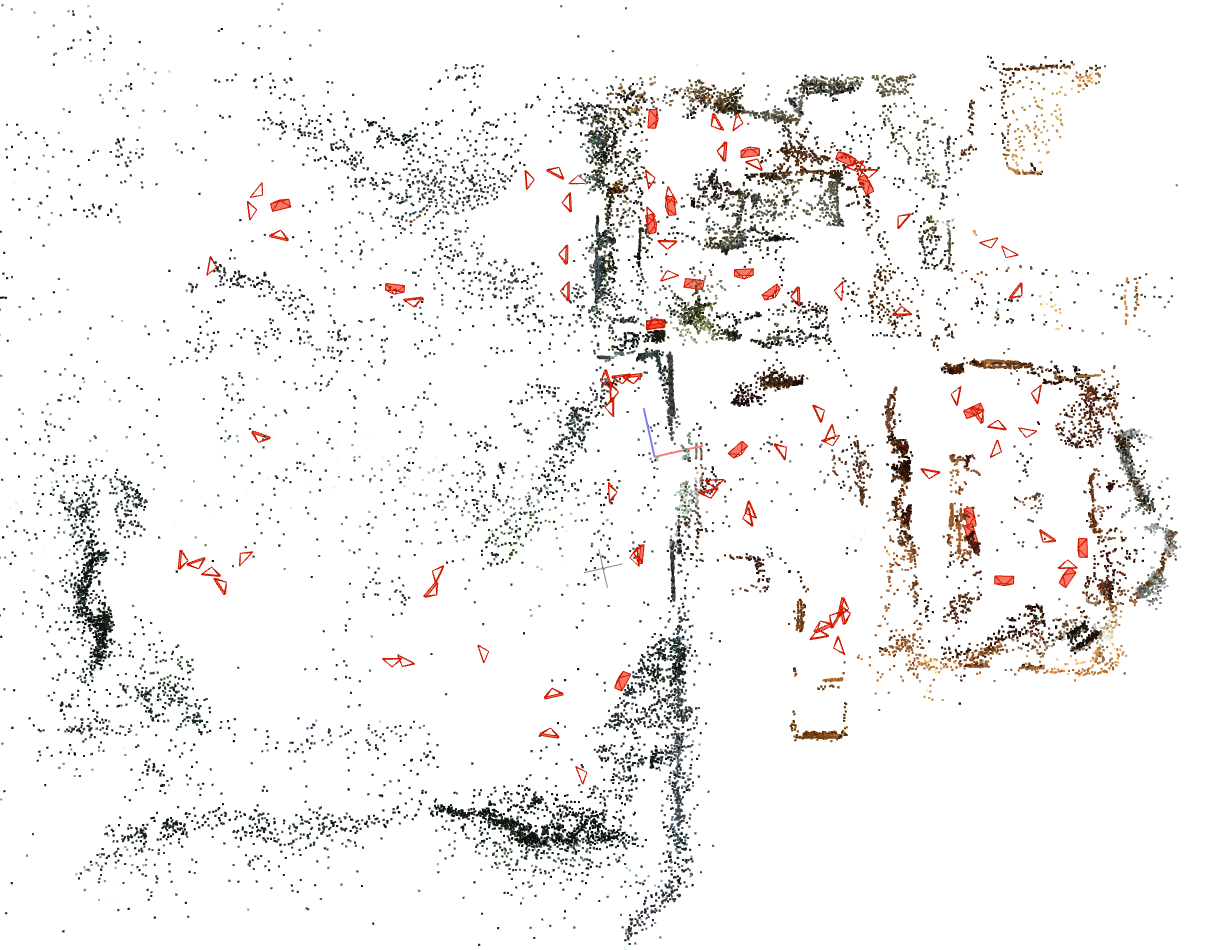} &
    \includegraphics[height=0.36\columnwidth]{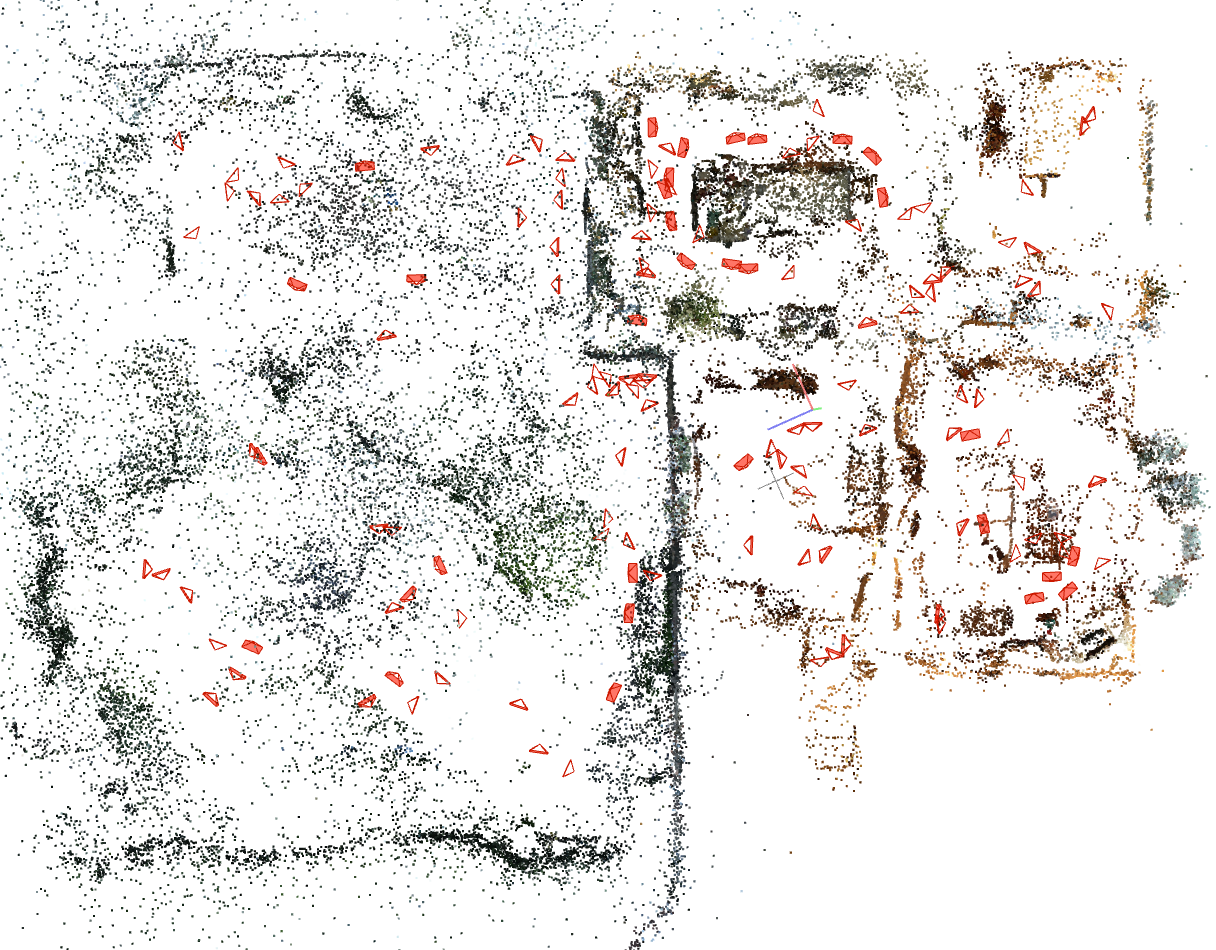} &
    \includegraphics[height=0.36\columnwidth]{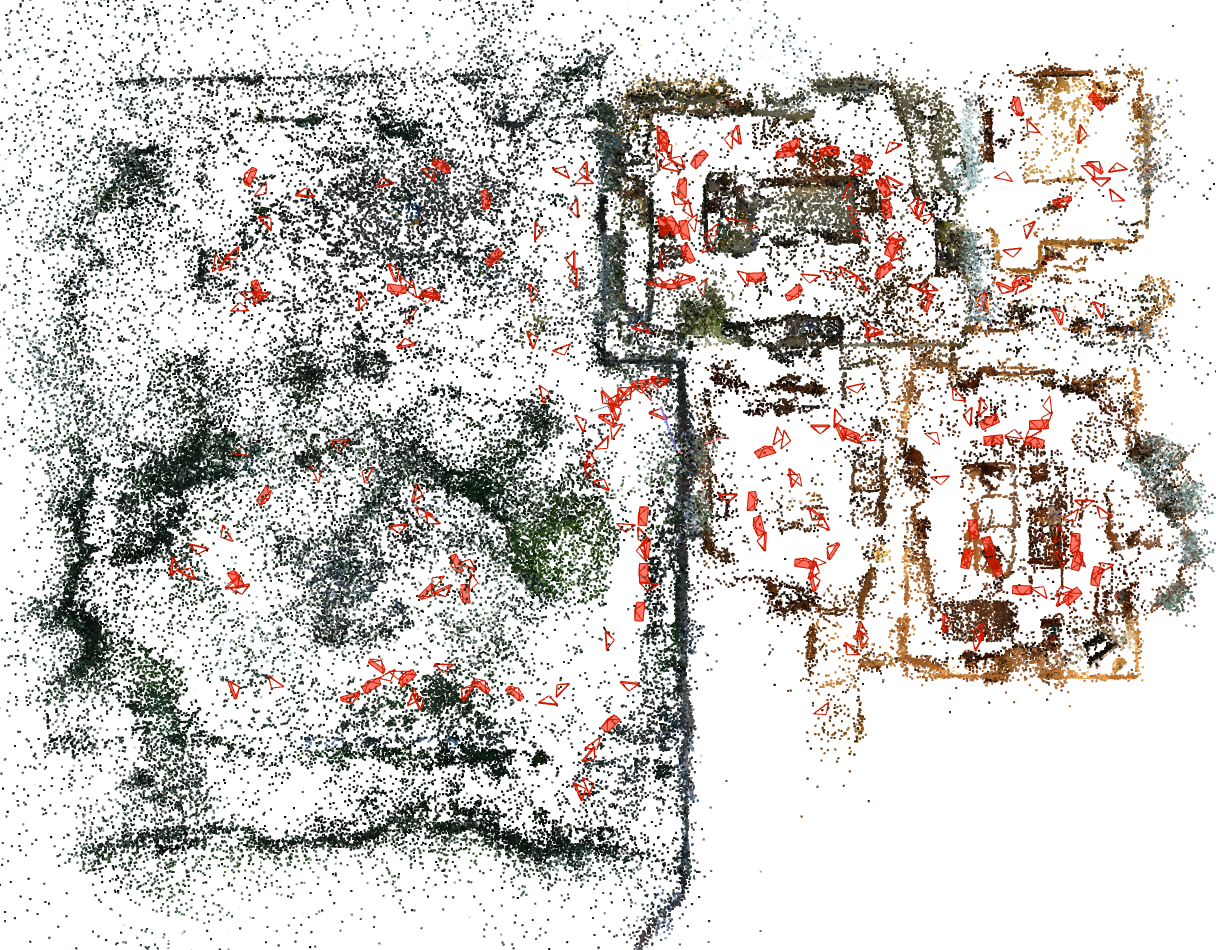} \\
    & 5.3/43.7/75.3 & 14.7/72.9/90.0 & 33.5/84.8/96.2 & 53.3/90.0/97.5 \\

    \rotatebox{90}{~~~~~~~~London ($\pi^3$)} &
    \includegraphics[height=0.36\columnwidth]{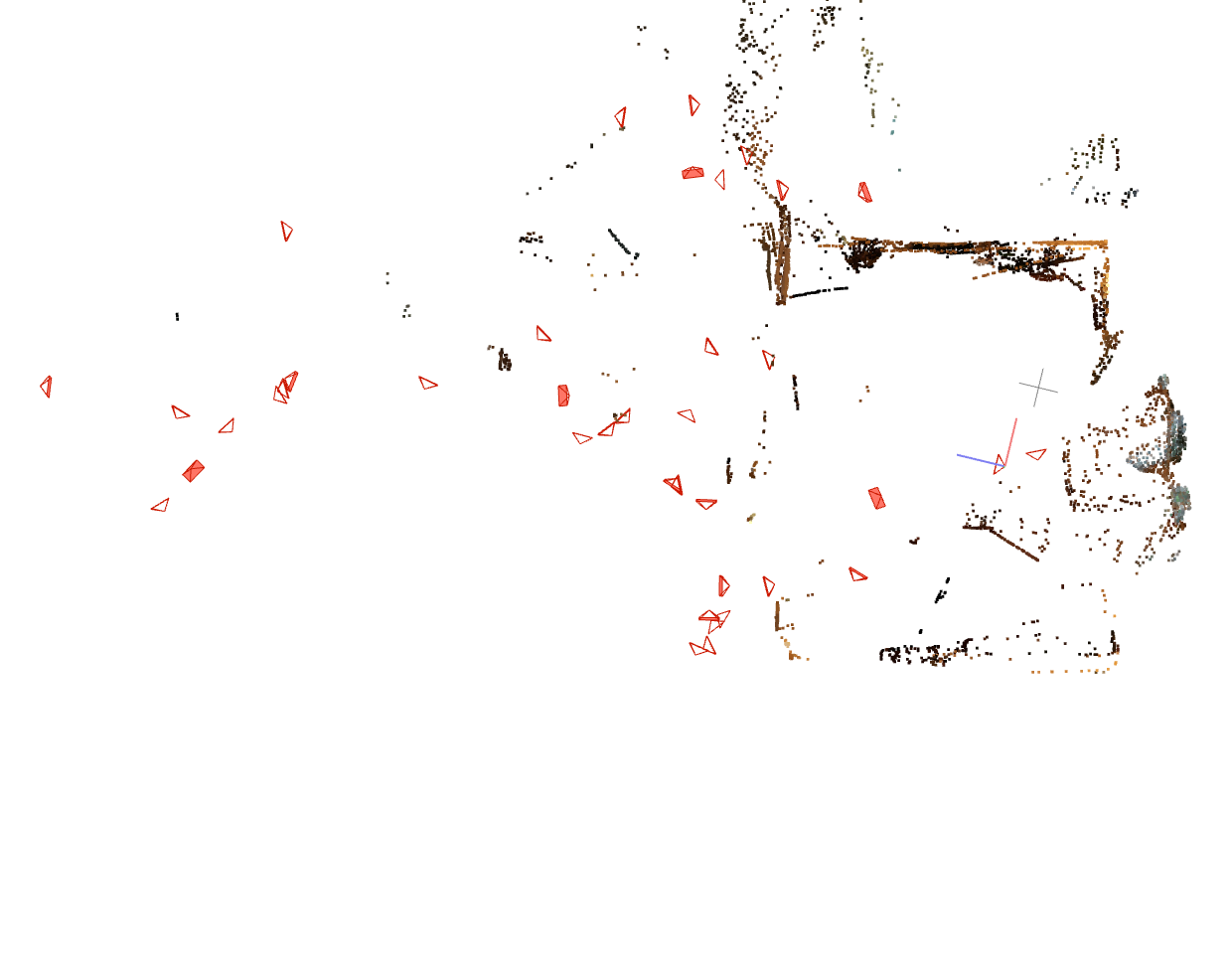} &
    \includegraphics[height=0.36\columnwidth]{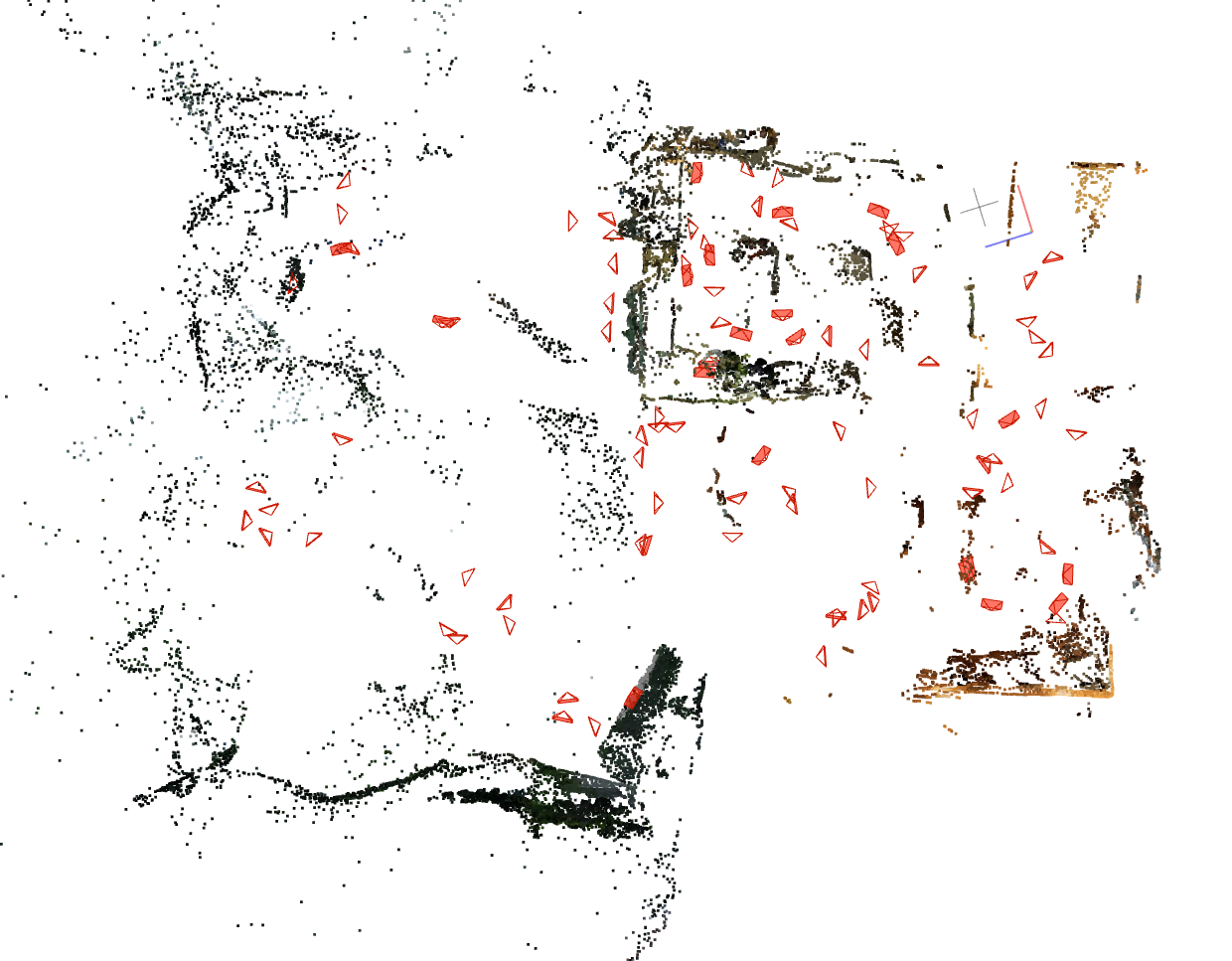} &
    \includegraphics[height=0.36\columnwidth]{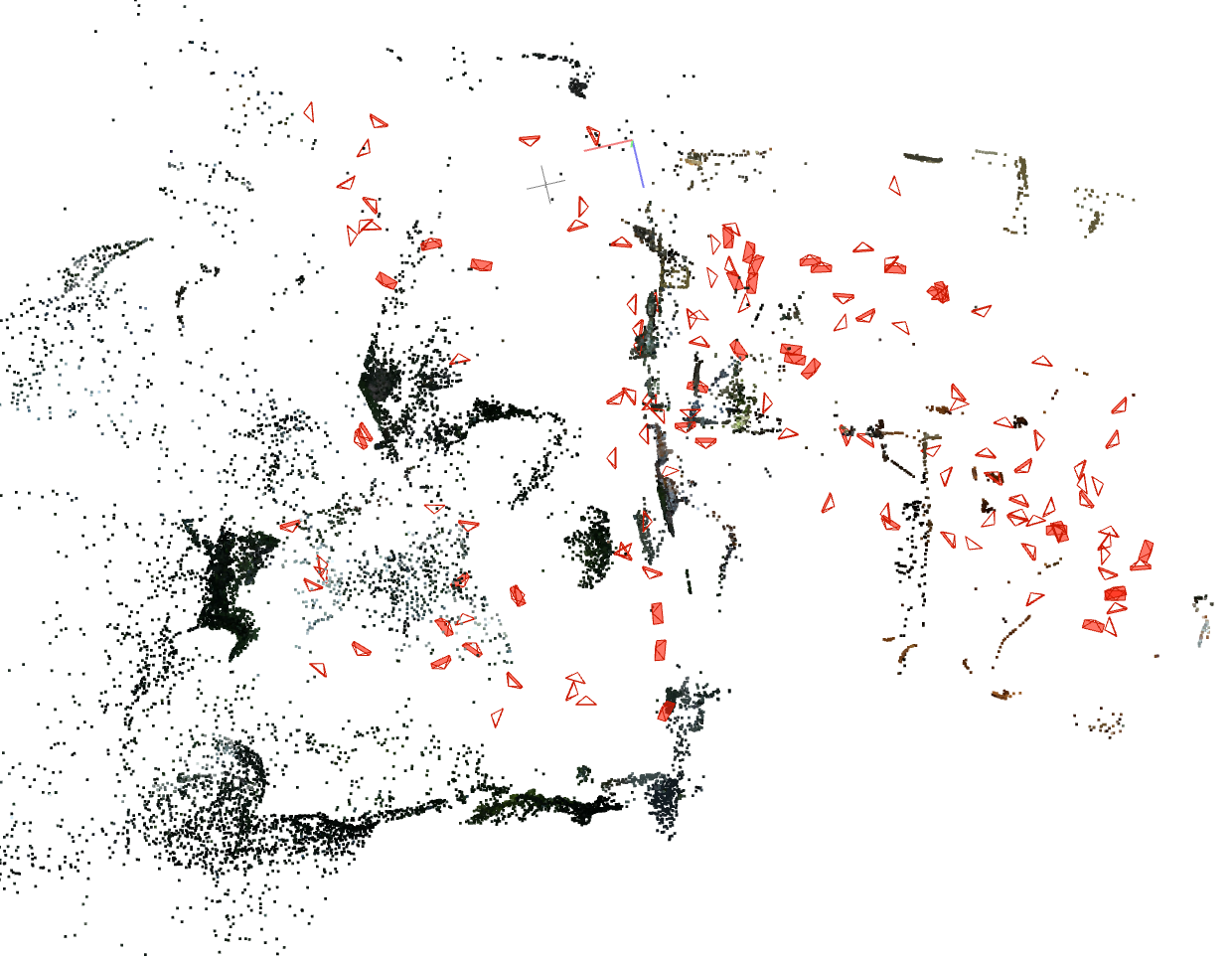} &
    \includegraphics[height=0.36\columnwidth]{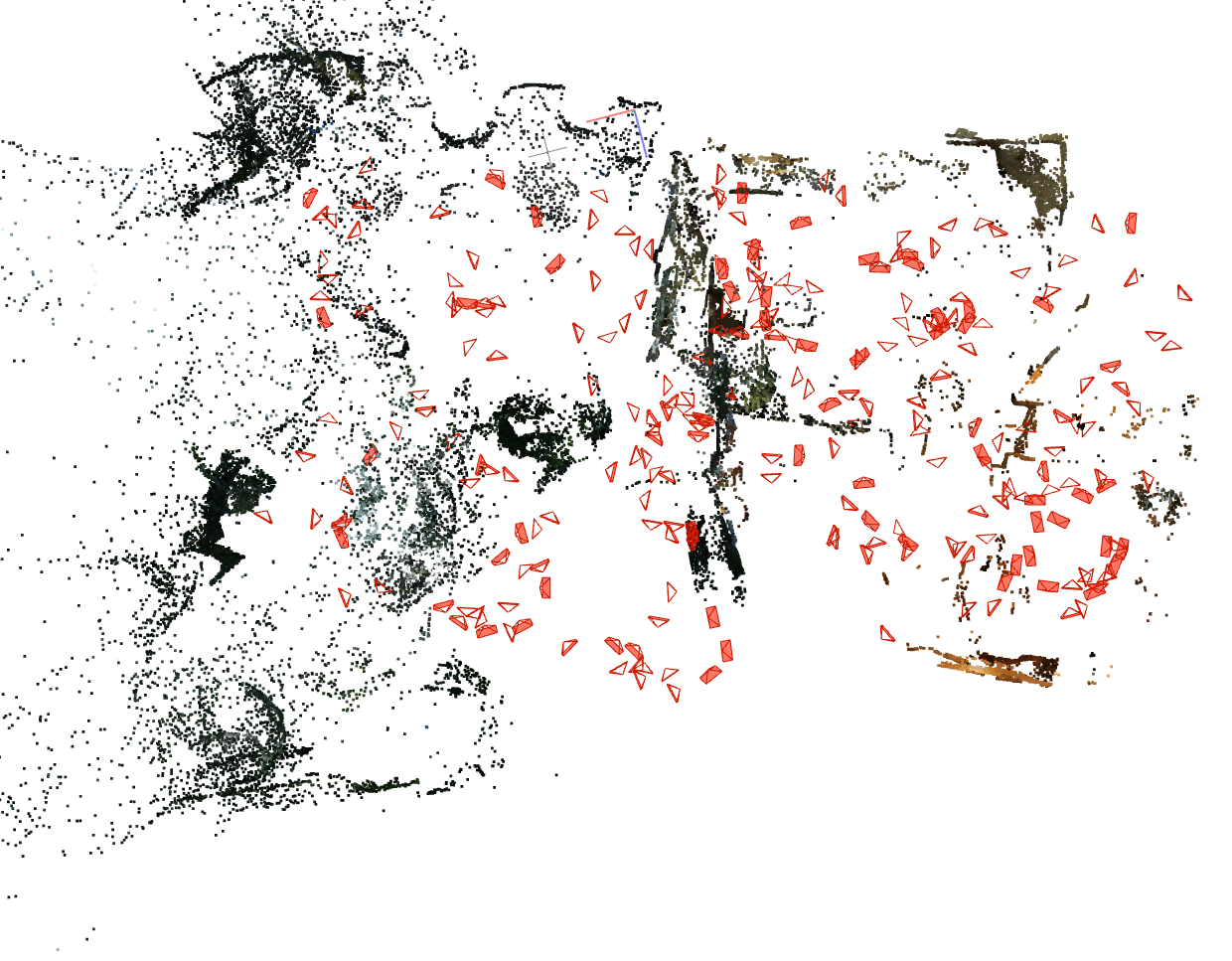} \\
     & 2.7/8.6/25.3 & 1.1/13.3/61.9 & 0.7/2.9/27.2 & 0.4/4.1/38.1\\

    \rotatebox{90}{~~~~~~~~~~~NYC (Ours)} &
    \includegraphics[height=0.36\columnwidth]{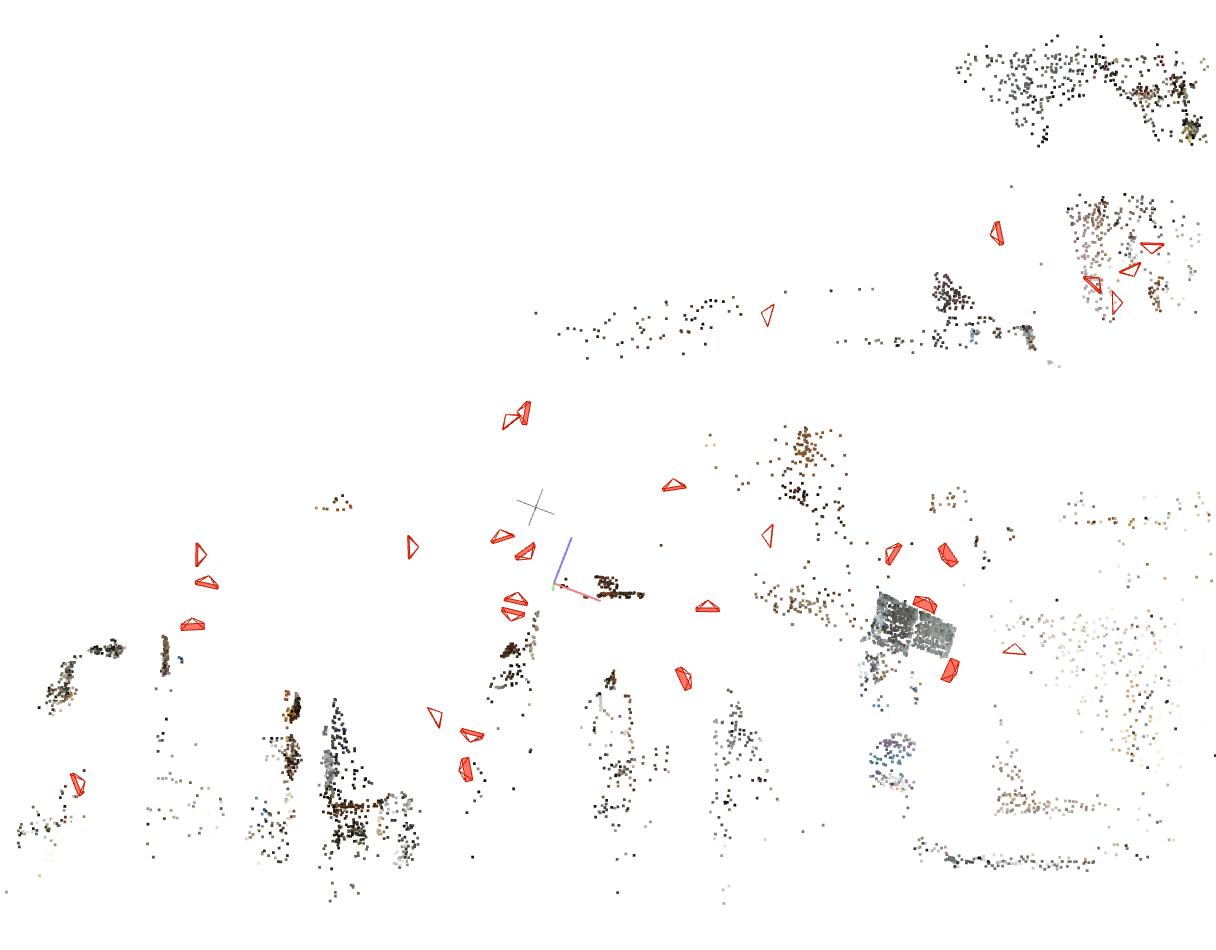} &
    \includegraphics[height=0.36\columnwidth]{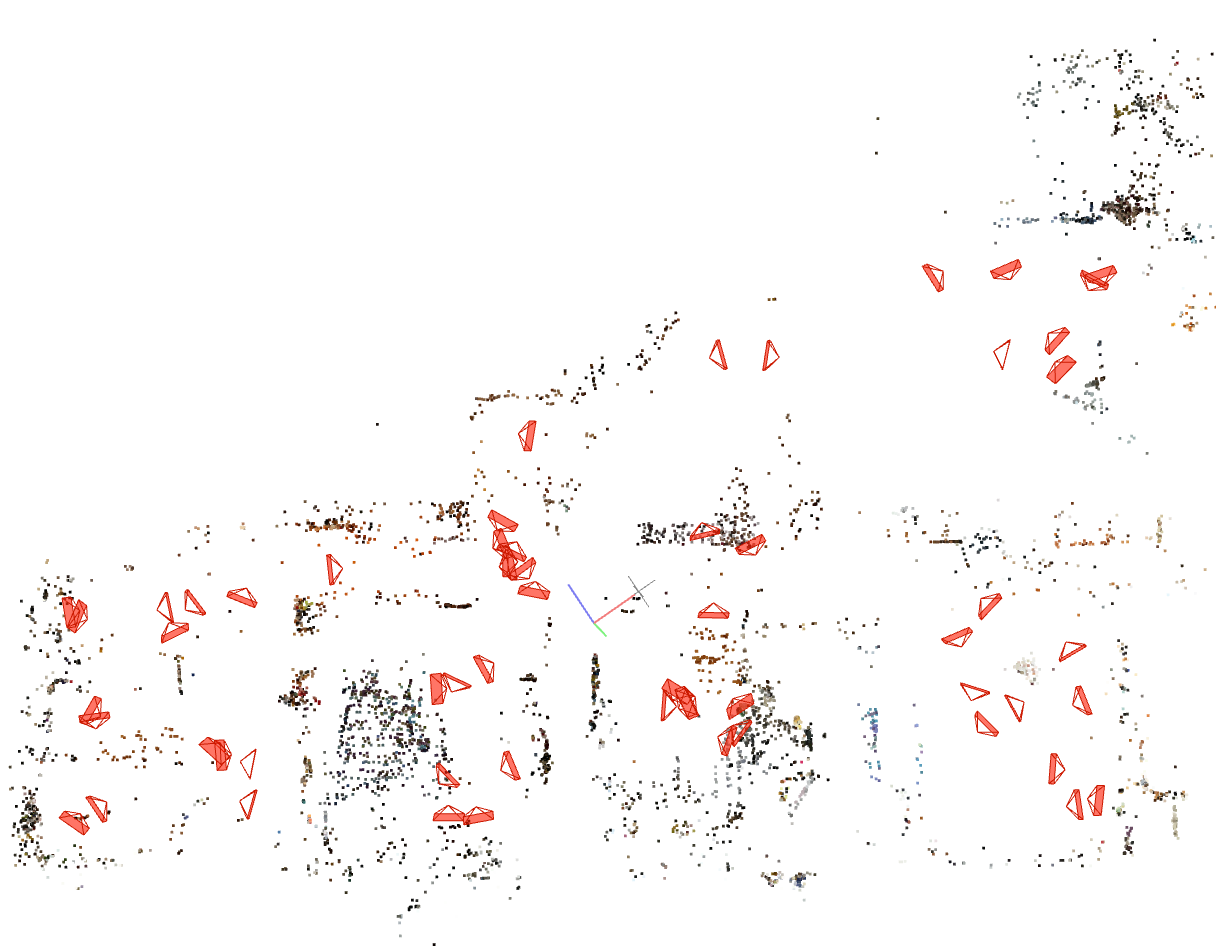} &
    \includegraphics[height=0.36\columnwidth]{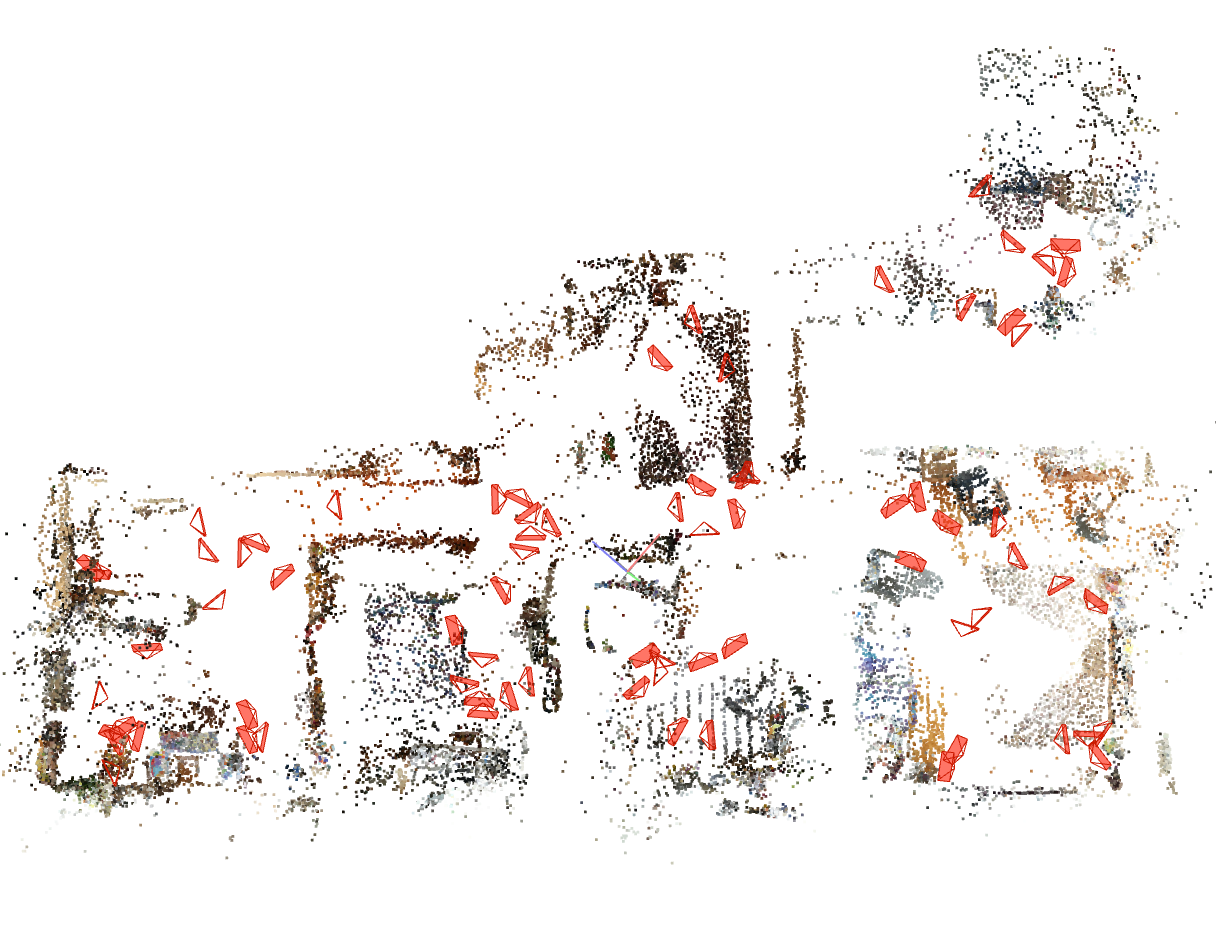} &
    \includegraphics[height=0.36\columnwidth]{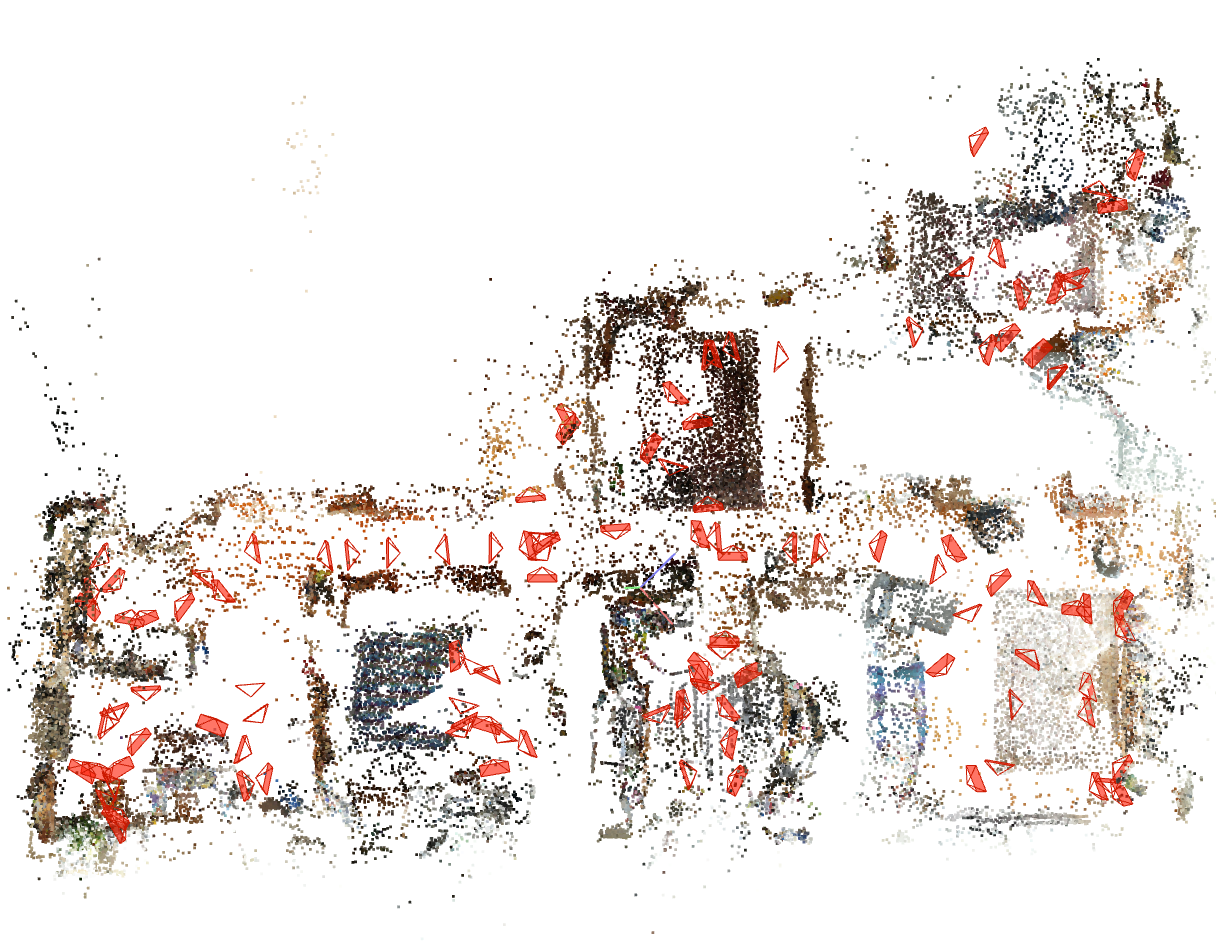} \\
    & 15.2/76.1/93.9 & 16.9/75.4/93.7 & 25.8/81.7/95.3 & 46.6/88.3/97.1 \\
    
    \rotatebox{90}{~~~~~~~~~~~NYC ($\pi^3$)} &
    \includegraphics[height=0.36\columnwidth]{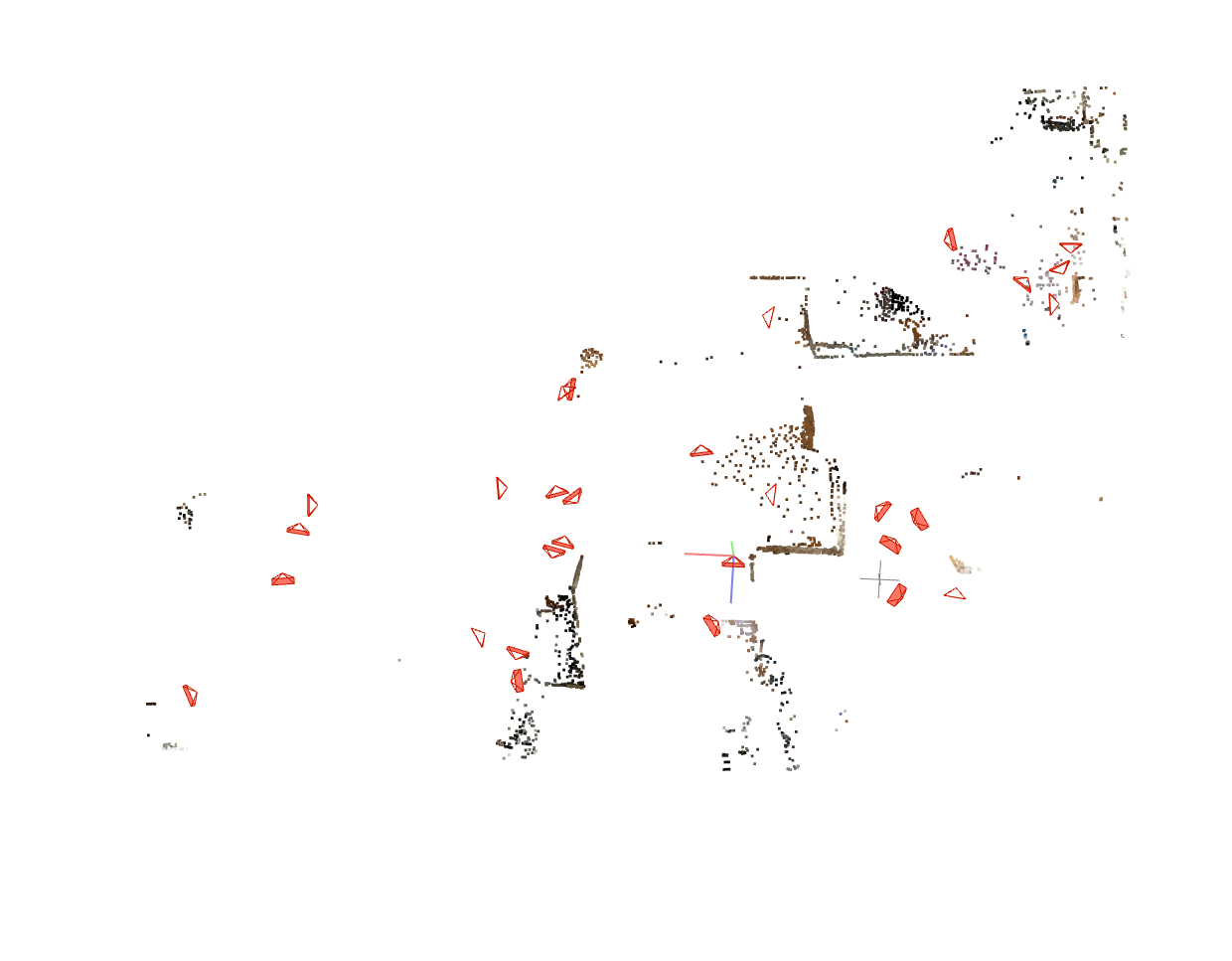} &
    \includegraphics[height=0.36\columnwidth]{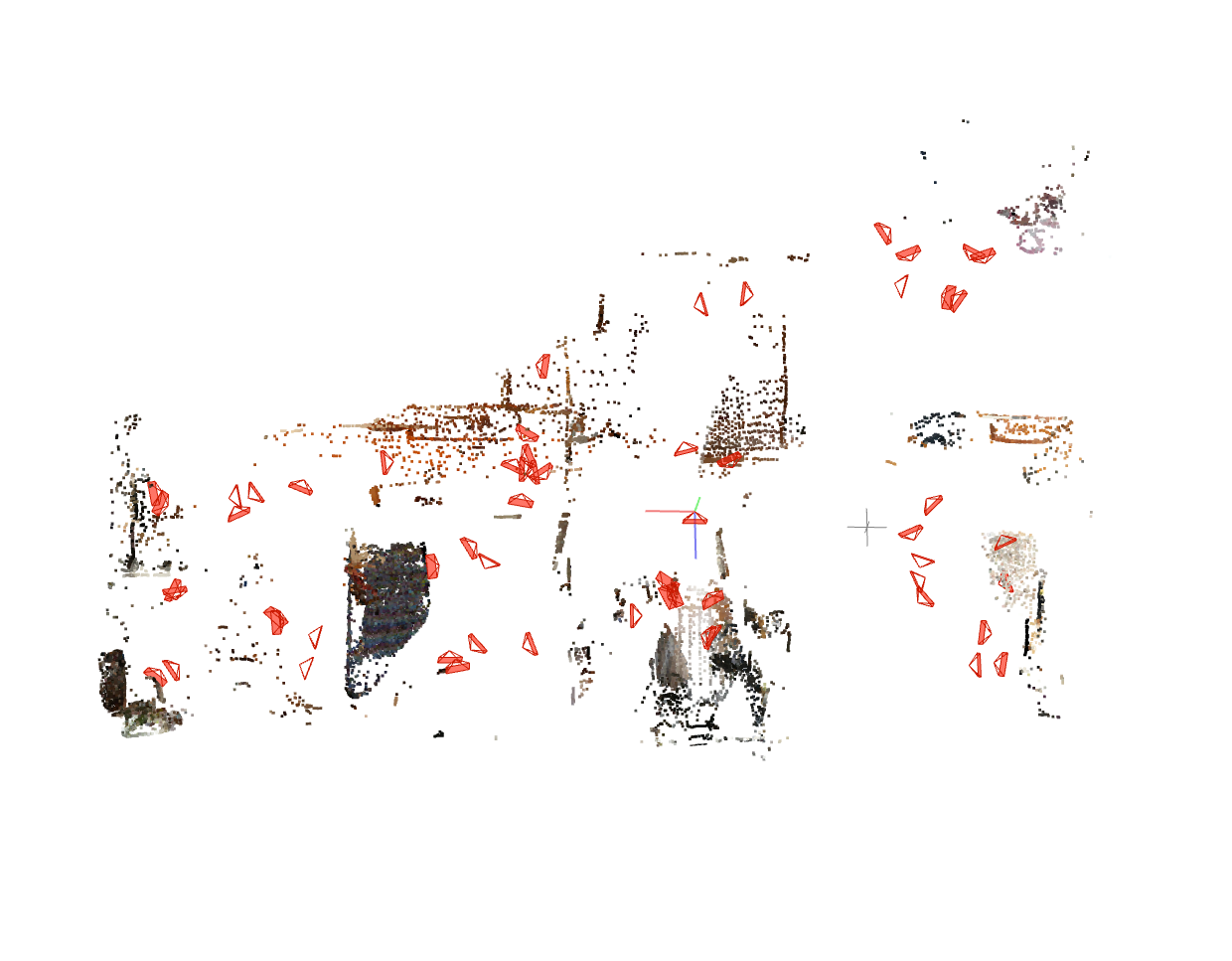} &
    \includegraphics[height=0.36\columnwidth]{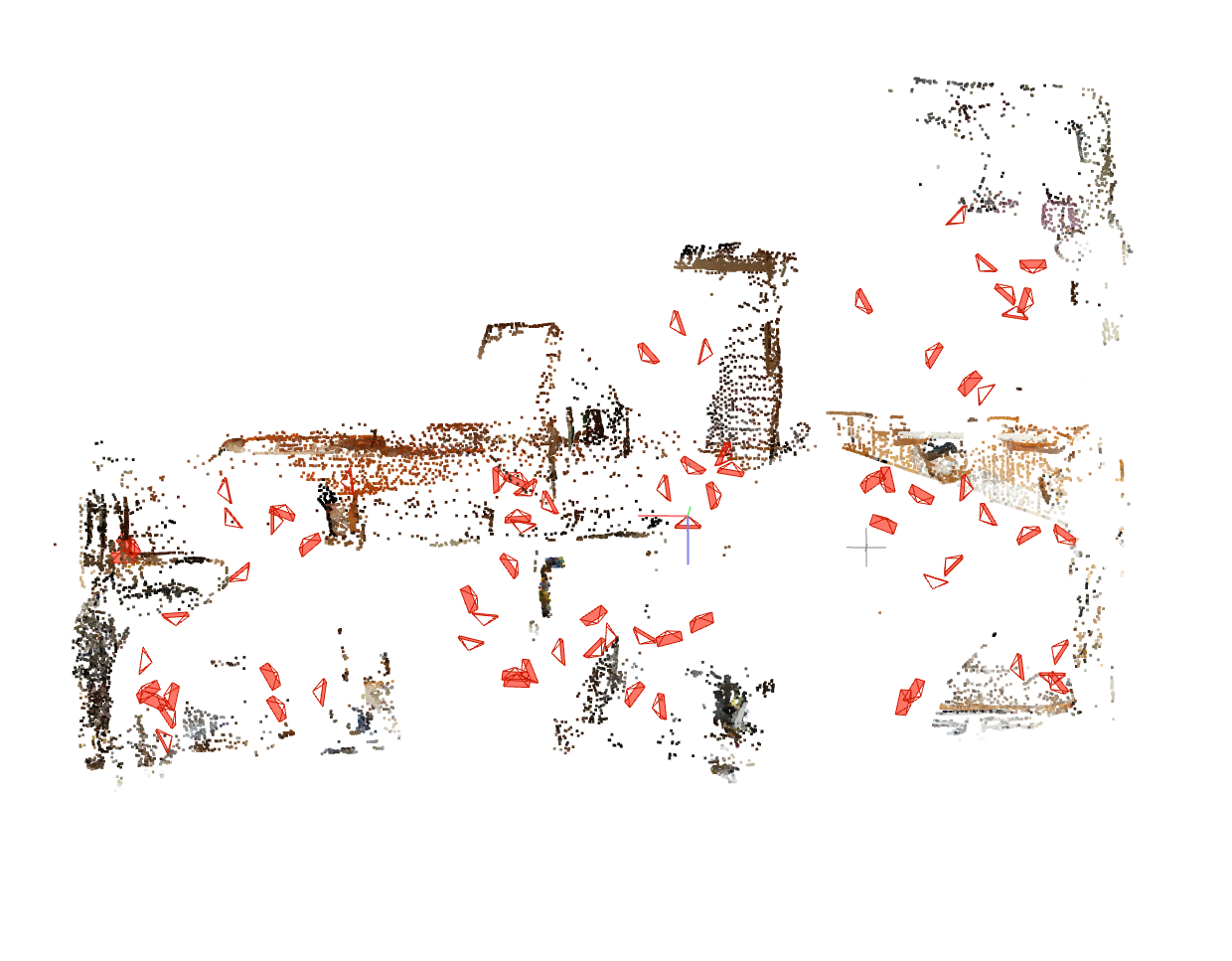} &
    \includegraphics[height=0.36\columnwidth]{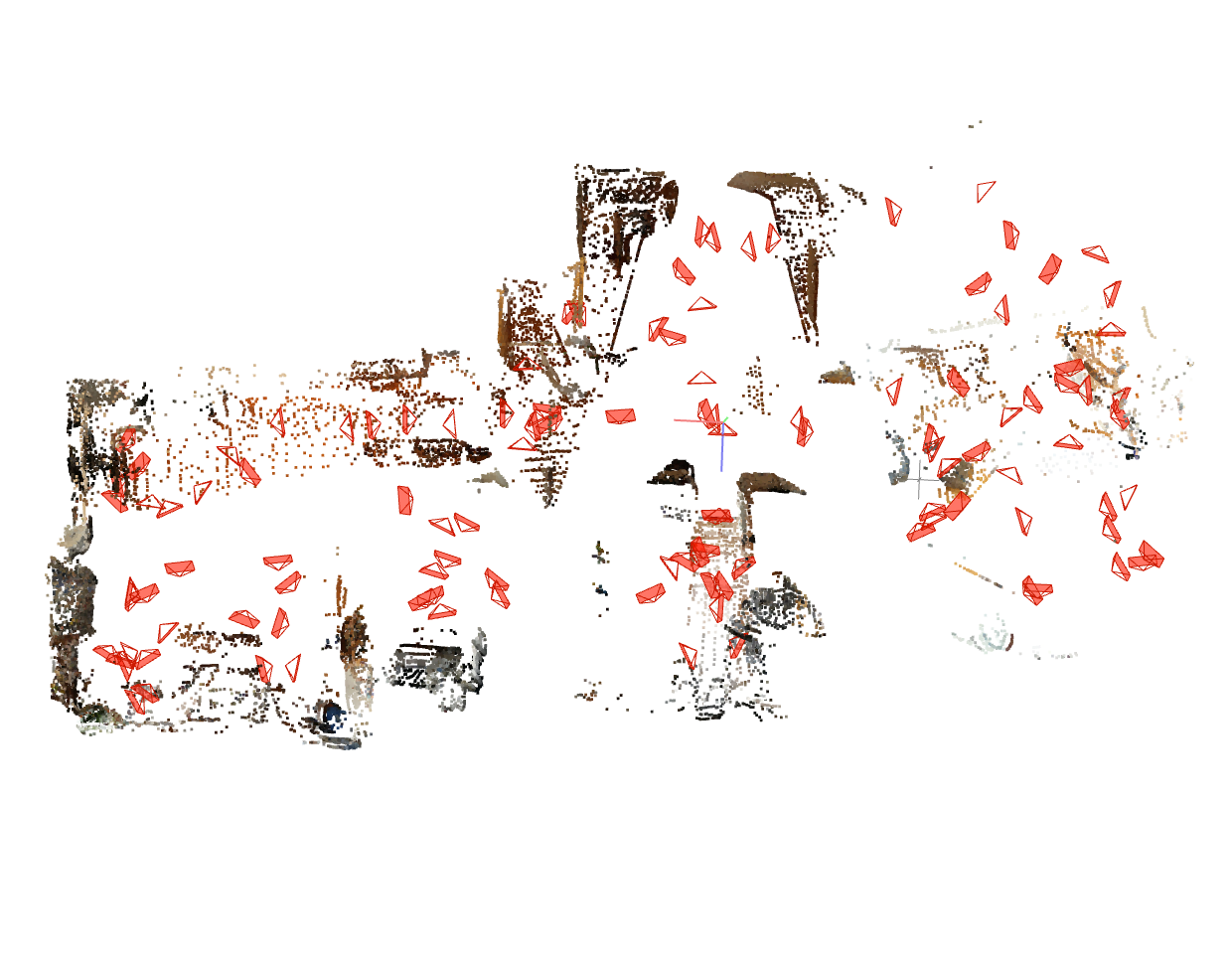} \\
     & 4.5/29.4/74.4 & 2.0/26.5/72.4 & 1.7/24.6/73.7 & 1.0/13.9/52.3\\
\end{tabular}
}
    \caption{
    Continuation of Figure~\ref{fig:low_overlap1}.
    }
\label{fig:low_overlap2}
\end{figure*}

\section{Detailed results}
In this section, we provide per-scene and per-category breakdowns for ETH3D~\cite{schops2017multi}, CO3DV2~\cite{reizenstein2021common}, and IMC2021~\cite{jin2021image}.

For ETH3D~\cite{schops2017multi}, per-scene results can be found in Table~\ref{tbl:eth3d_full}.
ETH3D features high-resolution images with millimeter-accurate ground truth across diverse indoor and outdoor environments.
The proposed method achieves the highest accuracy overall, benefiting from both the robustness of feedforward local estimation and the precision of classical global optimization.

For CO3D~\cite{reizenstein2021common}, per-category results can be found in Tables~\ref{tbl:co3d_full1},~\ref{tbl:co3d_full2}, and~\ref{tbl:co3d_full3}.
CO3D consists of object-centric video sequences that often exhibit low texture and limited multi-view overlap.
With 10 images (Table~\ref{tbl:co3d_full1}), $\pi^3$ with bundle adjustment generally achieves higher accuracy than our method, as the feedforward model can observe nearly the entire scene in a single pass, reducing the need for global optimization.
With 20 and 40 images, the proposed method matches or slightly exceeds $\pi^3$ + BA, as the larger scene coverage benefits from our global consistency enforcement.

For IMC2021~\cite{jin2021image}, per-scene results can be found in Table~\ref{tbl:imc_full}.
IMC2021 consists of unordered internet photo collections with significant illumination and viewpoint variation.
With only 5 images, $\pi^3$ + BA is slightly more accurate than the proposed method, though both significantly outperform classical methods that struggle with insufficient feature matches.
With 10 images, the proposed method achieves the best accuracy among both classical and feedforward approaches.
With 25 and full images, GLOMAP with SIFT features achieves the best accuracy as the denser sampling provides sufficient classical correspondences, while the proposed method remains competitive.

\begin{table*}[t]
    \centering
    \caption{Detailed results on ETH3D~\cite{schops2017multi}. Results marked with * are with ground truth calibration.}
    \vspace{-5px}
    \setlength{\tabcolsep}{3pt} 
    \resizebox{\textwidth}{!}{
    \begin{tabular}{l l c c c c c c c l c c c c c c c l c c c c c c c
    } \toprule 
&& \multicolumn{7}{c}{AUC@1} && \multicolumn{7}{c}{AUC@3} && \multicolumn{7}{c}{AUC@5} \\

        \cmidrule{3-9} \cmidrule{11-17} \cmidrule{19-25} 
        && \tiny{SIFT} & \tiny{AL+LG} & \tiny{$\pi^3$} & \tiny{$\pi^3$+BA} & \tiny{GLUEMAP$^\dagger$} & \tiny{GLUEMAP} & \tiny{GLUEMAP*} && \tiny{SIFT} & \tiny{AL+LG} & \tiny{$\pi^3$} & \tiny{$\pi^3$+BA} & \tiny{GLUEMAP$^\dagger$} & \tiny{GLUEMAP} & \tiny{GLUEMAP*} && \tiny{SIFT} & \tiny{AL+LG} & \tiny{$\pi^3$} & \tiny{$\pi^3$+BA} & \tiny{GLUEMAP$^\dagger$} & \tiny{GLUEMAP} & \tiny{GLUEMAP*} \\  \midrule
botanical\_garden & & 58.7 & \cellcolor{tabsecond}59.4 & 7.3 & 3.5 & 11.7 & 59.3 & \cellcolor{tabfirst}79.8 & & 84.5 & \cellcolor{tabsecond}84.8 & 21.1 & 6.6 & 49.9 & \cellcolor{tabsecond}84.8 & \cellcolor{tabfirst}93.2 & & 90.6 & 90.7 & 28.3 & 14.8 & 66.9 & \cellcolor{tabsecond}90.8 & \cellcolor{tabfirst}95.9 \\
boulders & & 3.8 & 4.0 & 17.0 & 42.3 & 18.8 & \cellcolor{tabsecond}47.6 & \cellcolor{tabfirst}82.5 & & 3.9 & 5.7 & 52.6 & 70.1 & 55.7 & \cellcolor{tabsecond}79.5 & \cellcolor{tabfirst}94.0 & & 4.2 & 8.2 & 67.6 & 78.9 & 71.1 & \cellcolor{tabsecond}87.1 & \cellcolor{tabfirst}96.4 \\
bridge & & 74.0 & 73.2 & 1.4 & 1.3 & 34.4 & \cellcolor{tabsecond}74.4 & \cellcolor{tabfirst}87.8 & & 90.9 & 90.4 & 3.3 & 11.3 & 71.3 & \cellcolor{tabsecond}91.0 & \cellcolor{tabfirst}95.7 & & 94.5 & 94.2 & 4.5 & 25.4 & 82.0 & \cellcolor{tabsecond}94.6 & \cellcolor{tabfirst}97.4 \\
courtyard & & 29.7 & 45.9 & 16.0 & 40.5 & 32.3 & \cellcolor{tabsecond}52.8 & \cellcolor{tabfirst}80.1 & & 42.2 & 76.4 & 51.4 & 73.6 & 69.1 & \cellcolor{tabsecond}82.1 & \cellcolor{tabfirst}93.4 & & 45.6 & 85.4 & 66.4 & 83.7 & 81.1 & \cellcolor{tabsecond}89.3 & \cellcolor{tabfirst}96.0 \\
delivery\_area & & 19.9 & \cellcolor{tabsecond}47.8 & 6.7 & 9.4 & 28.0 & 43.9 & \cellcolor{tabfirst}68.6 & & 25.5 & \cellcolor{tabsecond}69.9 & 32.7 & 34.4 & 60.3 & 68.9 & \cellcolor{tabfirst}78.5 & & 27.1 & \cellcolor{tabsecond}75.0 & 47.9 & 47.5 & 68.9 & 74.7 & \cellcolor{tabfirst}80.5 \\
door & & 79.6 & 70.7 & 21.9 & 47.8 & 21.8 & \cellcolor{tabsecond}80.2 & \cellcolor{tabfirst}86.9 & & 90.5 & 87.5 & 46.2 & 78.7 & 46.1 & \cellcolor{tabsecond}90.7 & \cellcolor{tabfirst}92.9 & & 92.7 & 90.9 & 64.7 & 85.6 & 64.6 & \cellcolor{tabsecond}92.8 & \cellcolor{tabfirst}94.1 \\
electro & & 47.3 & 44.4 & 9.6 & 12.8 & 18.2 & \cellcolor{tabsecond}50.8 & \cellcolor{tabfirst}79.5 & & 73.6 & 67.2 & 39.8 & 44.2 & 51.1 & \cellcolor{tabsecond}75.5 & \cellcolor{tabfirst}89.7 & & 81.2 & 73.8 & 56.5 & 60.3 & 64.2 & \cellcolor{tabsecond}82.6 & \cellcolor{tabfirst}92.1 \\
exhibition\_hall & & 25.2 & 8.7 & 1.6 & 1.5 & 3.7 & \cellcolor{tabsecond}26.9 & \cellcolor{tabfirst}82.2 & & 65.5 & 28.6 & 2.2 & 1.9 & 29.5 & \cellcolor{tabsecond}67.0 & \cellcolor{tabfirst}92.3 & & 77.7 & 38.7 & 3.2 & 2.6 & 47.7 & \cellcolor{tabsecond}78.7 & \cellcolor{tabfirst}95.2 \\
facade & & \cellcolor{tabsecond}71.4 & 68.8 & 19.7 & 21.7 & 21.8 & 70.4 & \cellcolor{tabfirst}88.8 & & \cellcolor{tabsecond}88.7 & 87.8 & 51.8 & 53.7 & 57.9 & \cellcolor{tabsecond}88.7 & \cellcolor{tabfirst}96.2 & & 93.0 & 92.5 & 65.5 & 67.4 & 72.7 & \cellcolor{tabsecond}93.1 & \cellcolor{tabfirst}97.7 \\
kicker & & 54.1 & 55.2 & 34.8 & \cellcolor{tabsecond}60.6 & 36.6 & 56.6 & \cellcolor{tabfirst}90.7 & & 77.1 & 81.6 & 71.7 & \cellcolor{tabsecond}85.2 & 71.6 & 82.7 & \cellcolor{tabfirst}96.9 & & 83.2 & 88.4 & 81.7 & \cellcolor{tabsecond}90.9 & 81.6 & 89.1 & \cellcolor{tabfirst}98.1 \\
lecture\_room & & 44.5 & \cellcolor{tabsecond}58.3 & 18.0 & 49.9 & 42.4 & 55.7 & \cellcolor{tabfirst}78.1 & & 62.8 & \cellcolor{tabsecond}80.2 & 51.0 & 76.3 & 75.0 & 79.1 & \cellcolor{tabfirst}91.7 & & 71.0 & \cellcolor{tabsecond}87.2 & 66.5 & 84.2 & 84.3 & 86.4 & \cellcolor{tabfirst}95.0 \\
living\_room & & \cellcolor{tabsecond}70.3 & 50.8 & 14.7 & 16.8 & 29.1 & 66.2 & \cellcolor{tabfirst}77.0 & & \cellcolor{tabsecond}88.4 & 66.8 & 48.2 & 50.1 & 66.9 & 86.7 & \cellcolor{tabfirst}91.4 & & \cellcolor{tabsecond}92.7 & 70.7 & 64.2 & 64.9 & 78.1 & 91.6 & \cellcolor{tabfirst}94.6 \\
lounge & & 34.0 & 28.1 & 20.3 & 34.0 & 18.9 & \cellcolor{tabsecond}37.1 & \cellcolor{tabfirst}42.3 & & 38.0 & 36.0 & \cellcolor{tabsecond}56.8 & \cellcolor{tabfirst}66.3 & 37.9 & 47.0 & 48.8 & & 38.8 & 37.6 & \cellcolor{tabsecond}72.8 & \cellcolor{tabfirst}76.7 & 44.6 & 49.9 & 51.1 \\
meadow & & 16.1 & 33.1 & 29.1 & 42.6 & 24.2 & \cellcolor{tabsecond}43.2 & \cellcolor{tabfirst}48.5 & & 31.0 & 75.7 & 69.1 & 74.8 & 65.6 & \cellcolor{tabsecond}76.7 & \cellcolor{tabfirst}79.4 & & 35.1 & 85.2 & 80.2 & 84.2 & 78.2 & \cellcolor{tabsecond}85.9 & \cellcolor{tabfirst}87.6 \\
observatory & & \cellcolor{tabsecond}43.0 & 21.2 & 5.5 & 33.4 & 9.4 & 42.8 & \cellcolor{tabfirst}68.1 & & \cellcolor{tabsecond}74.5 & 26.2 & 17.8 & 64.2 & 27.1 & 74.4 & \cellcolor{tabfirst}88.2 & & \cellcolor{tabsecond}84.0 & 27.2 & 30.1 & 76.2 & 43.1 & \cellcolor{tabsecond}84.0 & \cellcolor{tabfirst}92.9 \\
office & & 24.3 & 21.9 & 8.0 & \cellcolor{tabsecond}25.1 & 8.7 & 24.7 & \cellcolor{tabfirst}41.7 & & \cellcolor{tabsecond}42.6 & 36.9 & 18.4 & 41.6 & 20.6 & 41.4 & \cellcolor{tabfirst}54.4 & & 51.2 & 44.7 & 26.2 & \cellcolor{tabsecond}51.5 & 27.6 & 49.0 & \cellcolor{tabfirst}58.5 \\
old\_computer & & 12.9 & \cellcolor{tabsecond}31.0 & 2.1 & 18.0 & 23.3 & 29.6 & \cellcolor{tabfirst}50.6 & & 18.1 & 65.2 & 11.2 & 54.8 & 59.8 & \cellcolor{tabsecond}65.8 & \cellcolor{tabfirst}80.3 & & 19.7 & 77.6 & 26.0 & 69.2 & 73.7 & \cellcolor{tabsecond}78.1 & \cellcolor{tabfirst}87.8 \\
pipes & & 35.2 & 52.6 & 10.9 & 51.7 & 12.1 & \cellcolor{tabsecond}52.9 & \cellcolor{tabfirst}79.8 & & 41.9 & 81.3 & 32.8 & 80.8 & 37.0 & \cellcolor{tabsecond}81.9 & \cellcolor{tabfirst}92.9 & & 43.7 & 88.5 & 46.9 & 88.1 & 51.2 & \cellcolor{tabsecond}88.7 & \cellcolor{tabfirst}95.7 \\
playground & & 67.4 & 55.4 & 4.3 & 23.7 & 9.2 & \cellcolor{tabsecond}68.0 & \cellcolor{tabfirst}87.6 & & 88.2 & 83.1 & 16.1 & 53.0 & 38.6 & \cellcolor{tabsecond}88.6 & \cellcolor{tabfirst}95.8 & & 92.9 & 89.6 & 26.8 & 66.9 & 53.9 & \cellcolor{tabsecond}93.1 & \cellcolor{tabfirst}97.5 \\
relief & & 44.9 & 8.7 & 4.3 & 4.5 & 5.1 & \cellcolor{tabsecond}60.5 & \cellcolor{tabfirst}84.1 & & 59.9 & 11.3 & 17.1 & 12.0 & 22.8 & \cellcolor{tabsecond}85.7 & \cellcolor{tabfirst}94.4 & & 63.4 & 11.9 & 36.4 & 26.0 & 37.5 & \cellcolor{tabsecond}91.3 & \cellcolor{tabfirst}96.6 \\
relief\_2 & & 32.8 & 12.5 & 5.8 & 3.8 & 8.5 & \cellcolor{tabsecond}37.3 & \cellcolor{tabfirst}53.5 & & 44.9 & 13.9 & 18.2 & 13.6 & 26.9 & \cellcolor{tabfirst}62.5 & \cellcolor{tabsecond}61.9 & & 47.5 & 14.2 & 29.1 & 24.3 & 42.5 & \cellcolor{tabfirst}74.6 & \cellcolor{tabsecond}69.4 \\
statue & & \cellcolor{tabsecond}67.4 & 64.4 & 9.9 & 60.9 & 9.7 & 66.9 & \cellcolor{tabfirst}95.2 & & \cellcolor{tabsecond}89.1 & 88.1 & 19.2 & 87.0 & 19.6 & 89.0 & \cellcolor{tabfirst}98.4 & & \cellcolor{tabsecond}93.5 & 92.9 & 32.3 & 92.2 & 32.3 & 93.4 & \cellcolor{tabfirst}99.0 \\
terrace & & 55.3 & 47.7 & 10.7 & 50.6 & 27.7 & \cellcolor{tabsecond}60.5 & \cellcolor{tabfirst}89.6 & & 84.8 & 81.3 & 51.0 & 80.8 & 69.8 & \cellcolor{tabsecond}86.7 & \cellcolor{tabfirst}96.5 & & 90.9 & 88.8 & 68.1 & 88.2 & 80.8 & \cellcolor{tabsecond}92.0 & \cellcolor{tabfirst}97.9 \\
terrace\_2 & & \cellcolor{tabsecond}84.6 & 82.4 & 43.3 & 64.9 & 43.0 & 83.7 & \cellcolor{tabfirst}89.5 & & \cellcolor{tabsecond}94.2 & 93.3 & 68.4 & 84.1 & 68.1 & 93.9 & \cellcolor{tabfirst}96.4 & & \cellcolor{tabsecond}96.5 & 96.1 & 78.4 & 90.4 & 78.2 & 96.4 & \cellcolor{tabfirst}97.8 \\
terrains & & \cellcolor{tabfirst}44.3 & 26.3 & 7.5 & \cellcolor{tabsecond}42.8 & 8.4 & 33.6 & 36.7 & & \cellcolor{tabsecond}53.8 & 34.0 & 34.9 & \cellcolor{tabfirst}78.0 & 27.9 & 52.6 & \cellcolor{tabsecond}53.8 & & 55.9 & 36.0 & 52.7 & \cellcolor{tabfirst}86.5 & 41.6 & \cellcolor{tabsecond}62.7 & 60.8 \\
      \bottomrule
    \end{tabular}
    }
    \label{tbl:eth3d_full}
\end{table*}

\begin{table*}[t]
    \centering
    \caption{Results per-category on CO3D~\cite{reizenstein2021common} with 10 images per scene. The proposed method is on par with $\pi^3$ + BA.}
    \vspace{-5px}
    \setlength{\tabcolsep}{3pt} 
    \resizebox{\textwidth}{!}{
    \begin{tabular}{l l c c c c c c l c c c c c c l c c c c c c
    } \toprule 
&& \multicolumn{6}{c}{AUC@3} && \multicolumn{6}{c}{AUC@10} && \multicolumn{6}{c}{AUC@30} \\

        \cmidrule{3-8} \cmidrule{10-15} \cmidrule{17-22} 
        && \tiny{SIFT} & \tiny{AL+LG} & \tiny{$\pi^3$} & \tiny{$\pi^3$+BA} & \tiny{GLUEMAP$^\dagger$} & \tiny{GLUEMAP} && \tiny{SIFT} & \tiny{AL+LG} & \tiny{$\pi^3$} & \tiny{$\pi^3$+BA} & \tiny{GLUEMAP$^\dagger$} & \tiny{GLUEMAP} && \tiny{SIFT} & \tiny{AL+LG} & \tiny{$\pi^3$} & \tiny{$\pi^3$+BA} & \tiny{GLUEMAP$^\dagger$} & \tiny{GLUEMAP} \\  \midrule
apple (10) & & 26.0 & 22.7 & 49.9 & \cellcolor{tabfirst}61.1 & 47.6 & \cellcolor{tabsecond}57.4 & & 34.9 & 30.2 & 78.7 & \cellcolor{tabfirst}82.5 & 77.8 & \cellcolor{tabsecond}81.2 & & 39.6 & 35.0 & 90.8 & \cellcolor{tabfirst}91.6 & 90.2 & \cellcolor{tabsecond}91.3 \\
backpack (10) & & 28.6 & 22.2 & 49.8 & \cellcolor{tabfirst}56.6 & 47.3 & \cellcolor{tabsecond}56.1 & & 41.4 & 32.6 & 79.8 & \cellcolor{tabsecond}81.8 & 79.0 & \cellcolor{tabfirst}82.2 & & 48.1 & 38.8 & 92.3 & \cellcolor{tabsecond}92.7 & 92.1 & \cellcolor{tabfirst}93.0 \\
banana (10) & & 30.7 & 19.1 & 42.6 & \cellcolor{tabfirst}55.1 & 41.0 & \cellcolor{tabsecond}53.8 & & 44.0 & 27.3 & 73.2 & \cellcolor{tabfirst}77.5 & 72.2 & \cellcolor{tabsecond}77.1 & & 51.5 & 32.7 & 87.1 & \cellcolor{tabfirst}88.4 & 86.8 & \cellcolor{tabsecond}88.1 \\
baseballbat (10) & & 26.8 & 18.8 & 37.5 & \cellcolor{tabfirst}48.7 & 34.4 & \cellcolor{tabsecond}48.0 & & 41.3 & 27.4 & 68.3 & \cellcolor{tabsecond}73.0 & 67.7 & \cellcolor{tabfirst}73.3 & & 50.6 & 35.0 & 84.7 & \cellcolor{tabfirst}86.2 & 84.1 & \cellcolor{tabsecond}86.0 \\
baseballglove (10) & & 33.6 & 25.0 & 43.5 & \cellcolor{tabsecond}55.8 & 43.7 & \cellcolor{tabfirst}56.3 & & 45.9 & 36.6 & 72.6 & \cellcolor{tabsecond}77.2 & 72.9 & \cellcolor{tabfirst}77.7 & & 52.6 & 43.1 & 85.0 & \cellcolor{tabsecond}86.7 & 85.1 & \cellcolor{tabfirst}87.0 \\
bench (10) & & 26.9 & 23.6 & 69.2 & \cellcolor{tabfirst}77.3 & 68.4 & \cellcolor{tabsecond}76.0 & & 32.7 & 29.4 & 90.0 & \cellcolor{tabfirst}92.6 & 89.7 & \cellcolor{tabsecond}92.2 & & 35.3 & 32.6 & 96.6 & \cellcolor{tabfirst}97.5 & 96.5 & \cellcolor{tabsecond}97.4 \\
bicycle (10) & & 29.8 & 24.4 & 57.2 & \cellcolor{tabfirst}67.2 & 57.0 & \cellcolor{tabsecond}65.0 & & 39.5 & 33.5 & 84.4 & \cellcolor{tabsecond}87.1 & 84.5 & \cellcolor{tabfirst}87.2 & & 43.9 & 38.6 & 94.2 & \cellcolor{tabsecond}95.1 & 94.4 & \cellcolor{tabfirst}95.3 \\
bottle (10) & & 28.3 & 16.3 & 50.6 & \cellcolor{tabfirst}57.2 & 47.9 & \cellcolor{tabsecond}56.2 & & 40.1 & 22.3 & 78.2 & \cellcolor{tabsecond}80.0 & 77.7 & \cellcolor{tabfirst}80.3 & & 45.3 & 26.6 & 90.0 & \cellcolor{tabfirst}91.2 & 90.1 & \cellcolor{tabsecond}91.1 \\
bowl (10) & & 25.2 & 11.2 & 44.0 & \cellcolor{tabfirst}53.3 & 41.1 & \cellcolor{tabsecond}47.4 & & 40.5 & 14.1 & 75.5 & \cellcolor{tabfirst}77.3 & 73.9 & \cellcolor{tabsecond}75.9 & & 47.6 & 17.6 & \cellcolor{tabsecond}88.5 & \cellcolor{tabfirst}88.8 & 87.0 & 87.8 \\
broccoli (10) & & 25.2 & 17.7 & 38.9 & \cellcolor{tabsecond}48.3 & 37.5 & \cellcolor{tabfirst}51.3 & & 36.8 & 25.8 & 71.3 & \cellcolor{tabsecond}75.2 & 71.0 & \cellcolor{tabfirst}75.7 & & 43.8 & 32.0 & \cellcolor{tabsecond}88.4 & \cellcolor{tabfirst}89.9 & 87.3 & 87.8 \\
cake (10) & & 21.7 & 18.5 & 40.8 & \cellcolor{tabsecond}51.0 & 40.9 & \cellcolor{tabfirst}53.0 & & 31.6 & 27.0 & 71.0 & \cellcolor{tabsecond}75.0 & 72.4 & \cellcolor{tabfirst}77.5 & & 38.2 & 33.8 & 86.6 & \cellcolor{tabsecond}88.1 & \cellcolor{tabsecond}88.1 & \cellcolor{tabfirst}90.2 \\
car (10) & & 22.4 & 24.0 & 60.7 & 60.3 & \cellcolor{tabsecond}62.4 & \cellcolor{tabfirst}65.9 & & 28.2 & 29.4 & 83.4 & 81.5 & \cellcolor{tabsecond}84.0 & \cellcolor{tabfirst}84.9 & & 31.4 & 33.9 & 92.2 & 90.6 & \cellcolor{tabsecond}92.4 & \cellcolor{tabfirst}92.6 \\
carrot (10) & & 25.8 & 19.9 & 39.4 & \cellcolor{tabsecond}44.3 & 38.5 & \cellcolor{tabfirst}46.5 & & 39.1 & 30.7 & 72.5 & \cellcolor{tabsecond}72.7 & 72.0 & \cellcolor{tabfirst}75.0 & & 47.3 & 38.6 & \cellcolor{tabsecond}88.8 & 88.4 & 88.5 & \cellcolor{tabfirst}89.5 \\
cellphone (10) & & 21.6 & 19.2 & 33.6 & \cellcolor{tabfirst}40.1 & 34.2 & \cellcolor{tabsecond}39.1 & & 33.9 & 28.7 & 62.8 & \cellcolor{tabsecond}63.7 & 62.3 & \cellcolor{tabfirst}63.8 & & 43.2 & 35.1 & \cellcolor{tabfirst}79.9 & \cellcolor{tabsecond}79.4 & 78.4 & 78.9 \\
chair (10) & & 27.8 & 21.1 & 59.3 & \cellcolor{tabfirst}66.2 & 56.8 & \cellcolor{tabsecond}62.2 & & 36.8 & 28.8 & 85.6 & \cellcolor{tabfirst}87.7 & 84.8 & \cellcolor{tabsecond}86.5 & & 41.8 & 33.3 & 94.9 & \cellcolor{tabfirst}95.5 & 94.7 & \cellcolor{tabsecond}95.2 \\
cup (10) & & 20.0 & 23.8 & 48.3 & \cellcolor{tabfirst}55.3 & 46.1 & \cellcolor{tabsecond}53.5 & & 27.7 & 32.8 & 74.5 & \cellcolor{tabsecond}75.7 & 73.5 & \cellcolor{tabfirst}76.3 & & 31.6 & 37.7 & \cellcolor{tabfirst}86.5 & \cellcolor{tabfirst}86.5 & 85.3 & \cellcolor{tabsecond}86.3 \\
donut (10) & & 25.3 & 18.5 & 39.1 & \cellcolor{tabsecond}46.3 & 37.5 & \cellcolor{tabfirst}48.9 & & 38.8 & 26.6 & 75.8 & \cellcolor{tabsecond}77.1 & 73.4 & \cellcolor{tabfirst}77.9 & & 45.3 & 32.9 & 91.2 & \cellcolor{tabsecond}91.4 & 90.4 & \cellcolor{tabfirst}91.9 \\
hairdryer (10) & & 23.9 & 20.5 & 50.1 & \cellcolor{tabfirst}61.2 & 49.0 & \cellcolor{tabsecond}58.9 & & 32.9 & 30.5 & 80.9 & \cellcolor{tabfirst}85.1 & 80.4 & \cellcolor{tabsecond}84.5 & & 37.7 & 36.3 & 93.0 & \cellcolor{tabfirst}94.4 & 92.9 & \cellcolor{tabsecond}94.2 \\
handbag (10) & & 22.4 & 20.1 & 46.1 & \cellcolor{tabfirst}53.7 & 43.0 & \cellcolor{tabsecond}50.3 & & 30.1 & 29.1 & 75.9 & \cellcolor{tabfirst}78.9 & 74.3 & \cellcolor{tabsecond}77.8 & & 35.3 & 35.6 & 90.1 & \cellcolor{tabfirst}91.4 & 89.2 & \cellcolor{tabsecond}90.7 \\
hydrant (10) & & 32.4 & 21.1 & 61.7 & \cellcolor{tabfirst}70.1 & 59.8 & \cellcolor{tabsecond}67.7 & & 42.6 & 26.7 & 86.7 & \cellcolor{tabfirst}88.8 & 86.2 & \cellcolor{tabsecond}87.9 & & 46.8 & 30.0 & \cellcolor{tabsecond}95.1 & \cellcolor{tabfirst}95.4 & 95.0 & \cellcolor{tabsecond}95.1 \\
keyboard (10) & & 26.1 & 22.1 & 39.7 & \cellcolor{tabsecond}48.2 & 38.5 & \cellcolor{tabfirst}49.0 & & 39.2 & 31.5 & 69.6 & \cellcolor{tabfirst}73.4 & 69.5 & \cellcolor{tabsecond}73.2 & & 47.1 & 37.5 & 85.4 & \cellcolor{tabfirst}86.5 & 85.3 & \cellcolor{tabsecond}86.4 \\
laptop (10) & & 29.0 & 26.1 & 45.5 & \cellcolor{tabfirst}53.2 & 44.5 & \cellcolor{tabsecond}51.3 & & 42.2 & 37.5 & \cellcolor{tabsecond}72.6 & \cellcolor{tabfirst}74.3 & 72.1 & \cellcolor{tabfirst}74.3 & & 50.5 & 44.7 & \cellcolor{tabfirst}87.1 & 86.5 & 86.7 & \cellcolor{tabsecond}87.0 \\
microwave (10) & & 18.6 & 28.5 & 48.0 & 49.9 & \cellcolor{tabsecond}50.1 & \cellcolor{tabfirst}50.5 & & 22.4 & 36.4 & \cellcolor{tabsecond}74.0 & 73.2 & \cellcolor{tabfirst}74.2 & 73.1 & & 26.3 & 42.0 & \cellcolor{tabfirst}86.0 & 85.4 & \cellcolor{tabsecond}85.7 & 85.5 \\
motorcycle (10) & & 24.3 & 25.3 & 63.7 & \cellcolor{tabfirst}69.7 & 61.5 & \cellcolor{tabsecond}69.1 & & 31.7 & 34.8 & 88.2 & \cellcolor{tabsecond}89.3 & 87.7 & \cellcolor{tabfirst}90.3 & & 35.5 & 39.2 & 96.0 & \cellcolor{tabsecond}96.3 & 95.8 & \cellcolor{tabfirst}96.7 \\
mouse (10) & & 27.1 & 17.5 & 42.2 & \cellcolor{tabfirst}54.0 & 38.8 & \cellcolor{tabsecond}47.3 & & 38.8 & 27.1 & 76.3 & \cellcolor{tabfirst}81.2 & 74.1 & \cellcolor{tabsecond}78.0 & & 45.0 & 33.7 & 91.5 & \cellcolor{tabfirst}92.9 & 90.7 & \cellcolor{tabsecond}92.0 \\
orange (10) & & 23.2 & 20.2 & 42.9 & \cellcolor{tabfirst}49.1 & 42.5 & \cellcolor{tabsecond}49.0 & & 33.6 & 28.9 & \cellcolor{tabsecond}73.8 & 73.2 & 73.6 & \cellcolor{tabfirst}76.2 & & 39.6 & 33.6 & \cellcolor{tabsecond}88.2 & 86.8 & 87.7 & \cellcolor{tabfirst}88.6 \\
parkingmeter (10) & & 30.5 & 26.3 & \cellcolor{tabsecond}59.9 & 51.8 & 59.4 & \cellcolor{tabfirst}60.7 & & 44.3 & 38.3 & \cellcolor{tabfirst}84.3 & 78.7 & \cellcolor{tabsecond}83.8 & 83.3 & & 51.0 & 44.9 & \cellcolor{tabfirst}94.0 & 91.3 & \cellcolor{tabsecond}93.8 & 92.8 \\
pizza (10) & & 36.0 & 17.6 & 36.4 & \cellcolor{tabsecond}53.0 & 38.6 & \cellcolor{tabfirst}53.1 & & 52.6 & 28.3 & 69.1 & \cellcolor{tabfirst}74.8 & 69.8 & \cellcolor{tabsecond}74.4 & & 62.4 & 33.4 & \cellcolor{tabsecond}86.5 & \cellcolor{tabfirst}87.7 & 85.2 & 85.6 \\
plant (10) & & 30.0 & 16.7 & 52.7 & \cellcolor{tabfirst}64.6 & 51.2 & \cellcolor{tabsecond}59.9 & & 40.3 & 22.7 & 82.6 & \cellcolor{tabfirst}86.0 & 82.3 & \cellcolor{tabsecond}85.6 & & 44.3 & 26.1 & 93.9 & \cellcolor{tabsecond}94.6 & 93.9 & \cellcolor{tabfirst}95.0 \\
stopsign (10) & & 18.7 & 20.6 & 42.5 & \cellcolor{tabsecond}42.8 & 39.6 & \cellcolor{tabfirst}46.6 & & 27.4 & 29.9 & \cellcolor{tabfirst}78.0 & 71.1 & 76.0 & \cellcolor{tabsecond}77.9 & & 32.7 & 35.8 & \cellcolor{tabsecond}91.5 & 87.4 & 90.8 & \cellcolor{tabfirst}91.7 \\
teddybear (10) & & 25.9 & 26.8 & 53.1 & \cellcolor{tabfirst}58.2 & 50.7 & \cellcolor{tabsecond}57.7 & & 34.2 & 37.0 & 81.1 & \cellcolor{tabsecond}82.1 & 79.9 & \cellcolor{tabfirst}82.6 & & 39.1 & 42.6 & 92.4 & \cellcolor{tabsecond}92.8 & 92.0 & \cellcolor{tabfirst}93.2 \\
toaster (10) & & 20.9 & 17.4 & 57.4 & \cellcolor{tabfirst}66.1 & 55.2 & \cellcolor{tabsecond}60.9 & & 28.1 & 24.8 & 84.9 & \cellcolor{tabfirst}87.8 & 84.0 & \cellcolor{tabsecond}86.5 & & 32.0 & 30.1 & 94.8 & \cellcolor{tabfirst}95.8 & 94.5 & \cellcolor{tabsecond}95.4 \\
toilet (10) & & 28.3 & 31.6 & 39.0 & \cellcolor{tabsecond}41.7 & 39.9 & \cellcolor{tabfirst}42.5 & & 41.5 & 48.6 & 65.8 & \cellcolor{tabsecond}66.1 & \cellcolor{tabfirst}66.3 & 65.5 & & 51.6 & 61.3 & \cellcolor{tabsecond}81.9 & 81.3 & \cellcolor{tabfirst}82.0 & 81.4 \\
toybus (10) & & 22.9 & 18.4 & 48.8 & \cellcolor{tabsecond}51.1 & 47.2 & \cellcolor{tabfirst}54.1 & & 32.5 & 27.1 & \cellcolor{tabsecond}77.4 & 74.0 & 76.3 & \cellcolor{tabfirst}77.9 & & 37.4 & 32.8 & \cellcolor{tabsecond}89.0 & 87.4 & 88.8 & \cellcolor{tabfirst}89.5 \\
toyplane (10) & & 29.1 & 23.1 & 44.0 & \cellcolor{tabfirst}53.6 & 44.9 & \cellcolor{tabsecond}53.4 & & 40.1 & 34.5 & 72.5 & \cellcolor{tabsecond}73.6 & 72.6 & \cellcolor{tabfirst}74.9 & & 46.7 & 41.5 & \cellcolor{tabfirst}87.4 & 86.0 & 86.9 & \cellcolor{tabsecond}87.3 \\
toytrain (10) & & 17.5 & 14.1 & 44.4 & \cellcolor{tabfirst}51.6 & 44.8 & \cellcolor{tabsecond}48.7 & & 21.6 & 21.2 & 74.3 & \cellcolor{tabfirst}77.6 & 75.3 & \cellcolor{tabsecond}76.4 & & 23.9 & 25.8 & 87.6 & \cellcolor{tabfirst}88.7 & 87.9 & \cellcolor{tabsecond}88.4 \\
toytruck (10) & & 17.1 & 21.0 & 41.9 & \cellcolor{tabsecond}43.4 & 39.3 & \cellcolor{tabfirst}46.9 & & 24.7 & 29.5 & \cellcolor{tabsecond}72.1 & 71.2 & 70.6 & \cellcolor{tabfirst}73.9 & & 28.7 & 34.4 & \cellcolor{tabfirst}88.1 & 87.4 & 87.1 & \cellcolor{tabsecond}87.8 \\
tv (10) & & 36.3 & 16.8 & 50.6 & \cellcolor{tabsecond}53.4 & 51.0 & \cellcolor{tabfirst}55.2 & & 54.0 & 22.5 & 80.7 & \cellcolor{tabsecond}82.1 & 81.1 & \cellcolor{tabfirst}83.0 & & 60.4 & 38.7 & 93.1 & \cellcolor{tabfirst}93.5 & 92.7 & \cellcolor{tabsecond}93.4 \\
umbrella (10) & & 31.6 & 20.3 & 57.1 & \cellcolor{tabfirst}65.8 & 54.1 & \cellcolor{tabsecond}62.7 & & 43.5 & 29.5 & 84.8 & \cellcolor{tabfirst}87.7 & 83.9 & \cellcolor{tabsecond}86.8 & & 48.9 & 34.6 & 94.7 & \cellcolor{tabfirst}95.7 & 94.4 & \cellcolor{tabsecond}95.4 \\
vase (10) & & 21.9 & 14.9 & 52.5 & \cellcolor{tabfirst}59.3 & 50.9 & \cellcolor{tabsecond}57.0 & & 30.5 & 19.9 & 82.2 & \cellcolor{tabsecond}82.6 & 81.3 & \cellcolor{tabfirst}83.3 & & 35.1 & 23.8 & \cellcolor{tabsecond}93.8 & 93.1 & 93.4 & \cellcolor{tabfirst}94.0 \\
wineglass (10) & & 18.9 & 13.1 & 51.0 & \cellcolor{tabfirst}57.7 & 51.2 & \cellcolor{tabsecond}56.1 & & 26.5 & 16.7 & 77.3 & \cellcolor{tabfirst}79.0 & 77.0 & \cellcolor{tabsecond}78.4 & & 31.3 & 20.4 & \cellcolor{tabsecond}88.0 & \cellcolor{tabfirst}88.4 & 87.5 & 87.9 \\

      \bottomrule
    \end{tabular}
    }
    \label{tbl:co3d_full1}
\end{table*}

\begin{table*}[t]
    \centering
    \caption{Results per-category on CO3D~\cite{reizenstein2021common} with 20 images per scene. The proposed method is on par with $\pi^3$ + BA.}
    \vspace{-5px}
    \setlength{\tabcolsep}{3pt} 
    \resizebox{\textwidth}{!}{
    \begin{tabular}{l l c c c c c c l c c c c c c l c c c c c c
    } \toprule 
&& \multicolumn{6}{c}{AUC@3} && \multicolumn{6}{c}{AUC@10} && \multicolumn{6}{c}{AUC@30} \\

        \cmidrule{3-8} \cmidrule{10-15} \cmidrule{17-22} 
        && \tiny{SIFT} & \tiny{AL+LG} & \tiny{$\pi^3$} & \tiny{$\pi^3$+BA} & \tiny{GLUEMAP$^\dagger$} & \tiny{GLUEMAP} && \tiny{SIFT} & \tiny{AL+LG} & \tiny{$\pi^3$} & \tiny{$\pi^3$+BA} & \tiny{GLUEMAP$^\dagger$} & \tiny{GLUEMAP} && \tiny{SIFT} & \tiny{AL+LG} & \tiny{$\pi^3$} & \tiny{$\pi^3$+BA} & \tiny{GLUEMAP$^\dagger$} & \tiny{GLUEMAP} \\  \midrule
        apple (20) && 32.4 & 53.1 & 55.8 & \cellcolor{tabfirst}67.0 & 58.5 & \cellcolor{tabsecond}64.1 && 39.9 & 68.3 & 82.9 & \cellcolor{tabfirst}87.3 & 85.4 & \cellcolor{tabsecond}86.4 && 44.0 & 76.0 & 94.0 & \cellcolor{tabfirst}95.7 & \cellcolor{tabsecond}94.9 & 94.1 \\
apple (20) & & 32.4 & 53.1 & 52.8 & \cellcolor{tabsecond}67.0 & 57.4 & \cellcolor{tabfirst}69.3 & & 39.9 & 68.3 & 81.1 & \cellcolor{tabsecond}87.3 & 83.9 & \cellcolor{tabfirst}87.9 & & 44.0 & 76.0 & 93.2 & \cellcolor{tabfirst}95.7 & 94.2 & \cellcolor{tabsecond}95.6 \\
backpack (20) & & 31.0 & 38.5 & 44.9 & \cellcolor{tabsecond}53.6 & 46.4 & \cellcolor{tabfirst}54.5 & & 44.5 & 58.6 & 76.6 & \cellcolor{tabsecond}79.0 & 78.4 & \cellcolor{tabfirst}80.2 & & 53.4 & 72.5 & 91.3 & 91.2 & \cellcolor{tabsecond}92.1 & \cellcolor{tabfirst}92.4 \\
banana (20) & & 49.4 & 52.6 & 49.0 & \cellcolor{tabfirst}64.1 & 49.0 & \cellcolor{tabsecond}61.9 & & 63.9 & 65.7 & 77.3 & \cellcolor{tabfirst}83.0 & 77.4 & \cellcolor{tabsecond}81.5 & & 73.6 & 72.8 & 89.4 & \cellcolor{tabfirst}92.3 & 89.5 & \cellcolor{tabsecond}91.5 \\
baseballbat (20) & & 27.1 & 39.3 & 35.1 & \cellcolor{tabsecond}48.8 & 33.9 & \cellcolor{tabfirst}50.8 & & 43.5 & 55.2 & 66.7 & \cellcolor{tabfirst}72.1 & 66.4 & \cellcolor{tabsecond}71.4 & & 53.9 & 66.7 & 81.7 & \cellcolor{tabfirst}84.0 & 81.6 & \cellcolor{tabsecond}82.5 \\
baseballglove (20) & & 47.0 & 47.3 & 39.8 & \cellcolor{tabfirst}59.1 & 41.8 & \cellcolor{tabsecond}58.5 & & 55.9 & 60.9 & 67.0 & \cellcolor{tabfirst}75.1 & 67.8 & \cellcolor{tabsecond}74.8 & & 61.4 & 67.4 & 79.8 & \cellcolor{tabfirst}83.4 & 79.6 & \cellcolor{tabsecond}83.1 \\
bench (20) & & 30.7 & 51.3 & 71.0 & \cellcolor{tabfirst}81.8 & 67.0 & \cellcolor{tabsecond}76.6 & & 38.3 & 63.7 & 90.9 & \cellcolor{tabfirst}94.3 & 89.6 & \cellcolor{tabsecond}92.6 & & 41.1 & 67.6 & 96.9 & \cellcolor{tabfirst}98.1 & 96.5 & \cellcolor{tabsecond}97.5 \\
bicycle (20) & & 45.0 & 57.4 & 49.6 & \cellcolor{tabfirst}69.4 & 55.2 & \cellcolor{tabsecond}66.5 & & 58.3 & 76.5 & 79.6 & \cellcolor{tabfirst}87.4 & 82.0 & \cellcolor{tabsecond}86.0 & & 63.5 & 83.6 & 91.3 & \cellcolor{tabfirst}94.4 & 92.4 & \cellcolor{tabsecond}93.6 \\
bottle (20) & & 35.9 & 21.7 & 43.7 & \cellcolor{tabfirst}56.2 & 40.8 & \cellcolor{tabsecond}48.2 & & 50.2 & 31.7 & 77.2 & \cellcolor{tabfirst}80.7 & 76.7 & \cellcolor{tabsecond}78.7 & & 56.1 & 36.5 & 91.9 & \cellcolor{tabfirst}93.1 & 91.8 & \cellcolor{tabsecond}92.4 \\
bowl (20) & & \cellcolor{tabsecond}43.9 & 17.7 & 38.1 & \cellcolor{tabfirst}51.2 & 32.2 & 43.6 & & 66.6 & 28.4 & 73.5 & \cellcolor{tabfirst}76.4 & 71.0 & \cellcolor{tabsecond}74.8 & & 75.9 & 34.3 & 86.5 & \cellcolor{tabfirst}87.5 & 85.8 & \cellcolor{tabsecond}87.0 \\
broccoli (20) & & 35.6 & 28.4 & 38.2 & \cellcolor{tabfirst}53.6 & 36.7 & \cellcolor{tabsecond}46.9 & & 46.3 & 41.4 & 72.0 & \cellcolor{tabfirst}76.7 & 69.4 & \cellcolor{tabsecond}72.7 & & 50.9 & 48.3 & \cellcolor{tabsecond}89.1 & \cellcolor{tabfirst}90.7 & 85.0 & 85.7 \\
cake (20) & & 33.4 & 46.6 & 41.4 & \cellcolor{tabfirst}54.0 & 43.4 & \cellcolor{tabsecond}52.4 & & 45.4 & 65.6 & 70.2 & \cellcolor{tabsecond}75.4 & 73.7 & \cellcolor{tabfirst}78.7 & & 52.9 & 73.4 & 88.2 & \cellcolor{tabsecond}90.1 & 89.6 & \cellcolor{tabfirst}91.7 \\
car (20) & & 22.0 & 44.5 & 52.8 & \cellcolor{tabsecond}57.2 & 55.0 & \cellcolor{tabfirst}59.7 & & 31.9 & 61.2 & 79.2 & 79.3 & \cellcolor{tabsecond}81.1 & \cellcolor{tabfirst}82.7 & & 39.1 & 72.8 & 91.2 & 91.1 & \cellcolor{tabsecond}91.8 & \cellcolor{tabfirst}92.3 \\
carrot (20) & & 31.5 & 36.5 & 51.0 & \cellcolor{tabfirst}63.4 & 48.9 & \cellcolor{tabsecond}60.3 & & 42.0 & 50.1 & 82.1 & \cellcolor{tabfirst}86.7 & 80.6 & \cellcolor{tabsecond}84.3 & & 47.5 & 55.3 & 93.7 & \cellcolor{tabfirst}95.0 & 93.3 & \cellcolor{tabsecond}94.1 \\
cellphone (20) & & 18.1 & 23.6 & 24.3 & \cellcolor{tabfirst}36.0 & 27.0 & \cellcolor{tabsecond}33.2 & & 29.7 & 39.6 & 49.9 & \cellcolor{tabfirst}55.4 & 51.1 & \cellcolor{tabsecond}53.3 & & 40.0 & 51.5 & 66.2 & \cellcolor{tabfirst}70.1 & 66.8 & \cellcolor{tabsecond}67.9 \\
chair (20) & & 41.9 & 58.0 & 64.0 & \cellcolor{tabfirst}79.0 & 63.2 & \cellcolor{tabsecond}77.4 & & 50.9 & 68.6 & 88.1 & \cellcolor{tabfirst}92.6 & 87.9 & \cellcolor{tabsecond}92.4 & & 55.9 & 73.3 & 95.8 & \cellcolor{tabsecond}97.1 & 95.8 & \cellcolor{tabfirst}97.3 \\
cup (20) & & 29.8 & 27.3 & 46.8 & \cellcolor{tabfirst}64.2 & 47.3 & \cellcolor{tabsecond}55.6 & & 41.0 & 42.4 & 78.1 & \cellcolor{tabfirst}86.0 & 78.7 & \cellcolor{tabsecond}81.3 & & 45.4 & 52.4 & 91.2 & \cellcolor{tabfirst}94.9 & 91.6 & \cellcolor{tabsecond}92.4 \\
donut (20) & & 31.9 & 38.8 & 38.2 & \cellcolor{tabfirst}56.9 & 40.4 & \cellcolor{tabsecond}54.5 & & 43.9 & 56.5 & 76.3 & \cellcolor{tabfirst}81.2 & 77.5 & \cellcolor{tabsecond}80.8 & & 51.0 & 64.9 & 91.7 & \cellcolor{tabfirst}93.3 & 92.2 & \cellcolor{tabsecond}92.9 \\
hairdryer (20) & & 31.7 & 43.3 & 41.3 & \cellcolor{tabfirst}60.5 & 43.4 & \cellcolor{tabsecond}55.0 & & 46.5 & 67.6 & 77.6 & \cellcolor{tabfirst}85.7 & 77.2 & \cellcolor{tabsecond}83.9 & & 54.6 & 77.8 & 91.9 & \cellcolor{tabfirst}94.7 & 91.7 & \cellcolor{tabsecond}94.2 \\
handbag (20) & & 27.3 & 29.7 & 36.2 & \cellcolor{tabfirst}50.6 & 36.1 & \cellcolor{tabsecond}44.1 & & 39.1 & 44.9 & 69.2 & \cellcolor{tabfirst}74.9 & 68.4 & \cellcolor{tabsecond}73.2 & & 47.4 & 53.4 & 86.2 & \cellcolor{tabfirst}88.3 & 86.1 & \cellcolor{tabsecond}87.8 \\
hydrant (20) & & 53.8 & 58.8 & 63.1 & \cellcolor{tabfirst}77.1 & 63.1 & \cellcolor{tabsecond}76.3 & & 65.9 & 69.7 & 87.4 & \cellcolor{tabsecond}91.8 & 87.6 & \cellcolor{tabfirst}92.0 & & 70.0 & 73.9 & 95.3 & \cellcolor{tabsecond}97.0 & 95.6 & \cellcolor{tabfirst}97.2 \\
keyboard (20) & & 26.9 & 34.6 & 38.2 & \cellcolor{tabsecond}50.6 & 43.6 & \cellcolor{tabfirst}51.5 & & 38.8 & 48.9 & 68.2 & \cellcolor{tabfirst}74.0 & 70.1 & \cellcolor{tabsecond}73.7 & & 43.5 & 55.9 & 83.1 & \cellcolor{tabfirst}86.5 & 83.5 & \cellcolor{tabsecond}84.7 \\
laptop (20) & & 46.4 & 46.2 & 50.0 & \cellcolor{tabfirst}62.7 & 51.5 & \cellcolor{tabsecond}60.4 & & 56.9 & 64.1 & 77.7 & \cellcolor{tabfirst}81.8 & 78.9 & \cellcolor{tabsecond}81.4 & & 64.0 & 74.3 & 91.1 & \cellcolor{tabfirst}91.8 & \cellcolor{tabsecond}91.6 & 91.5 \\
microwave (20) & & 30.2 & 43.3 & 43.1 & \cellcolor{tabfirst}57.7 & 50.3 & \cellcolor{tabsecond}55.2 & & 39.2 & 56.4 & 75.9 & \cellcolor{tabsecond}80.3 & 80.1 & \cellcolor{tabfirst}81.3 & & 48.9 & 63.2 & 90.9 & 92.0 & \cellcolor{tabsecond}92.5 & \cellcolor{tabfirst}92.7 \\
motorcycle (20) & & 48.8 & 65.5 & 67.6 & \cellcolor{tabfirst}81.1 & 66.8 & \cellcolor{tabsecond}76.9 & & 57.3 & 79.3 & 89.7 & \cellcolor{tabfirst}93.9 & 89.5 & \cellcolor{tabsecond}92.6 & & 60.0 & 84.2 & 96.5 & \cellcolor{tabfirst}97.8 & 96.4 & \cellcolor{tabsecond}97.5 \\
mouse (20) & & 31.6 & 39.7 & 43.6 & \cellcolor{tabfirst}63.4 & 40.3 & \cellcolor{tabsecond}54.9 & & 43.9 & 55.4 & 80.0 & \cellcolor{tabfirst}85.5 & 77.4 & \cellcolor{tabsecond}82.7 & & 49.5 & 63.1 & 92.9 & \cellcolor{tabfirst}94.7 & 92.1 & \cellcolor{tabsecond}94.0 \\
orange (20) & & 34.8 & 32.2 & 37.8 & \cellcolor{tabfirst}51.9 & 41.2 & \cellcolor{tabsecond}51.5 & & 56.0 & 44.3 & 75.1 & \cellcolor{tabfirst}80.8 & 76.3 & \cellcolor{tabsecond}80.3 & & 64.8 & 49.3 & 91.0 & \cellcolor{tabfirst}93.2 & 91.7 & \cellcolor{tabsecond}92.9 \\
parkingmeter (20) & & 48.4 & 44.1 & 58.8 & \cellcolor{tabsecond}60.1 & 56.1 & \cellcolor{tabfirst}64.6 & & 68.4 & 65.9 & \cellcolor{tabsecond}85.3 & 84.2 & 84.1 & \cellcolor{tabfirst}86.2 & & 76.2 & 74.2 & \cellcolor{tabfirst}94.8 & 93.9 & 94.4 & \cellcolor{tabsecond}94.7 \\
pizza (20) & & 34.2 & 36.6 & 28.4 & \cellcolor{tabfirst}50.5 & 33.5 & \cellcolor{tabsecond}46.2 & & 52.3 & 51.8 & 63.0 & \cellcolor{tabfirst}72.1 & 65.4 & \cellcolor{tabsecond}69.5 & & 63.2 & 59.8 & 84.6 & \cellcolor{tabsecond}85.3 & 83.7 & \cellcolor{tabfirst}85.4 \\
plant (20) & & 52.3 & 50.4 & 61.0 & \cellcolor{tabfirst}71.4 & 61.2 & \cellcolor{tabsecond}71.2 & & 66.1 & 65.7 & 87.0 & \cellcolor{tabfirst}90.4 & \cellcolor{tabsecond}87.1 & \cellcolor{tabfirst}90.4 & & 70.8 & 72.3 & \cellcolor{tabsecond}95.6 & \cellcolor{tabfirst}96.7 & \cellcolor{tabsecond}95.6 & \cellcolor{tabfirst}96.7 \\
stopsign (20) & & 24.3 & 35.9 & 47.6 & \cellcolor{tabfirst}56.3 & 42.5 & \cellcolor{tabsecond}50.0 & & 36.0 & 52.0 & \cellcolor{tabsecond}81.3 & \cellcolor{tabfirst}85.0 & 78.4 & 79.6 & & 41.2 & 61.1 & \cellcolor{tabsecond}93.4 & \cellcolor{tabfirst}94.9 & 92.6 & 92.9 \\
teddybear (20) & & 29.8 & 46.2 & 56.0 & \cellcolor{tabfirst}62.9 & 50.2 & \cellcolor{tabsecond}58.2 & & 40.1 & 63.9 & 84.7 & \cellcolor{tabfirst}87.4 & 82.9 & \cellcolor{tabsecond}85.6 & & 44.4 & 72.8 & 94.6 & \cellcolor{tabfirst}95.7 & 94.0 & \cellcolor{tabsecond}95.0 \\
toaster (20) & & 29.3 & 46.3 & 57.1 & \cellcolor{tabfirst}71.2 & 58.5 & \cellcolor{tabsecond}67.4 & & 38.2 & 61.9 & 85.6 & \cellcolor{tabfirst}90.5 & 86.4 & \cellcolor{tabsecond}89.4 & & 41.6 & 68.8 & 95.1 & \cellcolor{tabfirst}96.8 & 95.4 & \cellcolor{tabsecond}96.4 \\
toilet (20) & & 33.1 & 33.2 & 31.7 & \cellcolor{tabfirst}43.9 & 34.1 & \cellcolor{tabsecond}39.0 & & 44.0 & 47.6 & 57.1 & \cellcolor{tabfirst}61.9 & 57.6 & \cellcolor{tabsecond}59.3 & & 52.7 & 57.0 & 71.8 & \cellcolor{tabsecond}72.4 & 72.3 & \cellcolor{tabfirst}72.9 \\
toybus (20) & & 29.0 & 34.3 & 41.1 & \cellcolor{tabfirst}53.0 & 42.8 & \cellcolor{tabsecond}50.1 & & 37.8 & 47.6 & 69.4 & \cellcolor{tabsecond}72.2 & 70.1 & \cellcolor{tabfirst}72.3 & & 43.4 & 56.7 & 82.2 & \cellcolor{tabsecond}82.8 & 82.7 & \cellcolor{tabfirst}83.3 \\
toyplane (20) & & 32.1 & 32.7 & 39.4 & \cellcolor{tabfirst}53.7 & 39.9 & \cellcolor{tabsecond}47.2 & & 46.7 & 50.6 & 66.5 & \cellcolor{tabfirst}72.4 & 66.7 & \cellcolor{tabsecond}68.5 & & 54.7 & 61.0 & \cellcolor{tabsecond}80.4 & \cellcolor{tabfirst}82.1 & 79.0 & 79.0 \\
toytrain (20) & & 22.3 & 34.4 & 40.4 & \cellcolor{tabsecond}48.6 & 41.6 & \cellcolor{tabfirst}48.8 & & 29.9 & 50.0 & 68.9 & \cellcolor{tabsecond}73.2 & 71.0 & \cellcolor{tabfirst}73.8 & & 34.0 & 57.5 & 81.6 & \cellcolor{tabsecond}85.1 & 84.0 & \cellcolor{tabfirst}85.6 \\
toytruck (20) & & 18.8 & 31.2 & 27.9 & \cellcolor{tabfirst}35.1 & 27.2 & \cellcolor{tabsecond}34.2 & & 29.3 & 44.5 & 60.7 & \cellcolor{tabfirst}63.7 & 59.8 & \cellcolor{tabsecond}62.3 & & 38.1 & 54.7 & \cellcolor{tabsecond}83.8 & \cellcolor{tabfirst}85.2 & 83.0 & 82.0 \\
tv (20) & & 47.9 & 42.5 & 49.0 & \cellcolor{tabsecond}55.5 & 50.3 & \cellcolor{tabfirst}57.4 & & 64.2 & 64.0 & 81.0 & \cellcolor{tabsecond}83.1 & 81.6 & \cellcolor{tabfirst}84.0 & & 70.6 & 79.3 & 93.5 & \cellcolor{tabsecond}93.9 & 93.6 & \cellcolor{tabfirst}94.4 \\
umbrella (20) & & 51.2 & 55.3 & 61.0 & \cellcolor{tabsecond}73.5 & 64.3 & \cellcolor{tabfirst}73.7 & & 64.6 & 67.2 & 86.5 & \cellcolor{tabfirst}90.7 & 88.0 & \cellcolor{tabsecond}90.6 & & 70.4 & 71.3 & 95.3 & \cellcolor{tabfirst}96.9 & 95.8 & \cellcolor{tabsecond}96.6 \\
vase (20) & & 38.0 & 38.3 & 55.3 & \cellcolor{tabfirst}68.3 & 55.9 & \cellcolor{tabsecond}66.6 & & 48.0 & 53.8 & 83.5 & \cellcolor{tabfirst}88.1 & 83.9 & \cellcolor{tabsecond}87.0 & & 52.0 & 60.4 & 94.4 & \cellcolor{tabfirst}96.0 & 94.5 & \cellcolor{tabsecond}95.6 \\
wineglass (20) & & 30.0 & 35.1 & 47.4 & \cellcolor{tabfirst}60.3 & 51.5 & \cellcolor{tabsecond}54.1 & & 41.4 & 48.5 & 75.8 & \cellcolor{tabfirst}82.4 & 78.2 & \cellcolor{tabsecond}78.5 & & 46.1 & 53.5 & 88.4 & \cellcolor{tabfirst}90.7 & \cellcolor{tabsecond}89.3 & 89.1 \\
      \bottomrule
    \end{tabular}
    }
    \label{tbl:co3d_full2}
\end{table*}

\begin{table*}[t]
    \centering
    \caption{Results per-category on CO3D~\cite{reizenstein2021common} with 40 images per scene. The proposed method is on par with $\pi^3$ + BA.}
    \vspace{-5px}
    \setlength{\tabcolsep}{3pt} 
    \resizebox{\textwidth}{!}{
    \begin{tabular}{l l c c c c c c l c c c c c c l c c c c c c
    } \toprule 
&& \multicolumn{6}{c}{AUC@3} && \multicolumn{6}{c}{AUC@10} && \multicolumn{6}{c}{AUC@30} \\

        \cmidrule{3-8} \cmidrule{10-15} \cmidrule{17-22} 
        && \tiny{SIFT} & \tiny{AL+LG} & \tiny{$\pi^3$} & \tiny{$\pi^3$+BA} & \tiny{GLUEMAP$^\dagger$} & \tiny{GLUEMAP} && \tiny{SIFT} & \tiny{AL+LG} & \tiny{$\pi^3$} & \tiny{$\pi^3$+BA} & \tiny{GLUEMAP$^\dagger$} & \tiny{GLUEMAP} && \tiny{SIFT} & \tiny{AL+LG} & \tiny{$\pi^3$} & \tiny{$\pi^3$+BA} & \tiny{GLUEMAP$^\dagger$} & \tiny{GLUEMAP} \\  \midrule
apple (40) & & 58.6 & 63.3 & 53.5 & \cellcolor{tabsecond}68.9 & 58.5 & \cellcolor{tabfirst}71.2 & & 70.2 & 79.6 & 80.9 & \cellcolor{tabsecond}87.2 & 84.1 & \cellcolor{tabfirst}88.4 & & 75.1 & 85.7 & 92.7 & \cellcolor{tabfirst}95.5 & \cellcolor{tabsecond}94.1 & \cellcolor{tabfirst}95.5 \\
backpack (40) & & 46.9 & 47.8 & 43.4 & \cellcolor{tabsecond}54.4 & 51.4 & \cellcolor{tabfirst}55.8 & & 66.4 & 75.3 & 76.2 & \cellcolor{tabsecond}78.8 & \cellcolor{tabfirst}80.9 & \cellcolor{tabfirst}80.9 & & 77.1 & 89.9 & 91.2 & 91.2 & \cellcolor{tabfirst}93.0 & \cellcolor{tabsecond}92.6 \\
banana (40) & & 59.0 & 63.7 & 47.8 & \cellcolor{tabsecond}65.0 & 51.2 & \cellcolor{tabfirst}65.5 & & 74.0 & 81.5 & 77.1 & \cellcolor{tabfirst}83.3 & 78.4 & \cellcolor{tabsecond}82.6 & & 81.8 & 90.7 & 89.6 & \cellcolor{tabfirst}92.4 & 90.3 & \cellcolor{tabsecond}91.8 \\
baseballbat (40) & & 46.6 & 45.0 & 37.0 & \cellcolor{tabfirst}52.3 & 36.0 & \cellcolor{tabsecond}51.1 & & 67.3 & 61.5 & 67.2 & \cellcolor{tabfirst}73.3 & 67.3 & \cellcolor{tabsecond}71.4 & & 78.4 & 77.5 & 82.1 & \cellcolor{tabfirst}84.5 & 82.3 & \cellcolor{tabsecond}82.6 \\
baseballglove (40) & & \cellcolor{tabfirst}63.1 & 51.6 & 40.3 & \cellcolor{tabsecond}59.2 & 40.6 & 55.4 & & \cellcolor{tabfirst}76.4 & 65.8 & 67.3 & \cellcolor{tabsecond}75.2 & 66.1 & 73.9 & & 83.2 & 73.2 & 80.9 & \cellcolor{tabsecond}83.7 & 80.0 & \cellcolor{tabfirst}84.3 \\
bench (40) & & 62.0 & 78.4 & 69.7 & \cellcolor{tabfirst}83.9 & 71.0 & \cellcolor{tabsecond}80.7 & & 74.3 & 89.0 & 90.3 & \cellcolor{tabfirst}94.9 & 90.9 & \cellcolor{tabsecond}93.8 & & 78.4 & 92.2 & 96.7 & \cellcolor{tabfirst}98.3 & 96.9 & \cellcolor{tabsecond}97.9 \\
bicycle (40) & & 55.0 & 65.3 & 56.5 & \cellcolor{tabfirst}66.6 & 56.8 & \cellcolor{tabsecond}66.2 & & 73.5 & \cellcolor{tabfirst}86.4 & 83.1 & 85.9 & 83.5 & \cellcolor{tabsecond}86.0 & & 80.4 & \cellcolor{tabfirst}94.3 & 93.2 & 93.8 & 93.4 & \cellcolor{tabsecond}94.2 \\
bottle (40) & & 32.3 & 37.8 & 43.5 & \cellcolor{tabfirst}55.2 & 43.2 & \cellcolor{tabsecond}52.0 & & 42.6 & 56.1 & 77.6 & \cellcolor{tabfirst}80.5 & \cellcolor{tabsecond}78.0 & \cellcolor{tabfirst}80.5 & & 46.8 & 64.7 & 91.8 & \cellcolor{tabfirst}93.0 & 92.2 & \cellcolor{tabsecond}92.9 \\
bowl (40) & & \cellcolor{tabsecond}49.8 & 22.4 & 37.0 & \cellcolor{tabfirst}51.8 & 31.6 & 46.3 & & \cellcolor{tabsecond}73.6 & 34.2 & 73.2 & \cellcolor{tabfirst}76.0 & 71.7 & \cellcolor{tabfirst}76.0 & & 83.6 & 41.3 & 86.6 & \cellcolor{tabsecond}87.6 & 86.5 & \cellcolor{tabfirst}87.7 \\
broccoli (40) & & 44.6 & \cellcolor{tabsecond}55.9 & 38.8 & 54.5 & 39.0 & \cellcolor{tabfirst}56.0 & & 57.5 & 73.2 & 72.1 & \cellcolor{tabfirst}77.1 & 72.2 & \cellcolor{tabsecond}77.0 & & 62.5 & 83.9 & \cellcolor{tabsecond}89.0 & \cellcolor{tabfirst}90.7 & 87.6 & 88.4 \\
cake (40) & & 51.4 & 54.1 & 41.1 & \cellcolor{tabsecond}54.4 & 41.9 & \cellcolor{tabfirst}55.7 & & 71.4 & \cellcolor{tabfirst}79.4 & 69.7 & \cellcolor{tabsecond}75.3 & 73.0 & \cellcolor{tabfirst}79.4 & & 81.6 & \cellcolor{tabfirst}91.5 & 87.1 & \cellcolor{tabsecond}89.8 & 88.9 & \cellcolor{tabfirst}91.5 \\
car (40) & & 38.9 & \cellcolor{tabfirst}61.2 & 51.8 & 58.8 & 55.4 & \cellcolor{tabsecond}60.9 & & 51.6 & 80.5 & 77.6 & \cellcolor{tabsecond}81.1 & 79.6 & \cellcolor{tabfirst}81.2 & & 58.8 & \cellcolor{tabsecond}90.2 & 88.9 & \cellcolor{tabfirst}91.6 & 89.6 & 89.7 \\
carrot (40) & & 43.6 & 58.8 & 52.1 & \cellcolor{tabfirst}63.9 & 51.2 & \cellcolor{tabsecond}63.3 & & 57.9 & 73.5 & 82.3 & \cellcolor{tabfirst}86.8 & 81.8 & \cellcolor{tabsecond}85.7 & & 64.4 & 79.1 & 93.8 & \cellcolor{tabfirst}95.2 & 93.7 & \cellcolor{tabsecond}94.7 \\
cellphone (40) & & 29.2 & 30.2 & 23.0 & \cellcolor{tabfirst}37.0 & 25.3 & \cellcolor{tabsecond}32.2 & & 48.5 & 48.3 & 49.5 & \cellcolor{tabfirst}56.2 & 50.5 & \cellcolor{tabsecond}53.2 & & 61.3 & 60.6 & 65.9 & \cellcolor{tabfirst}70.3 & 66.1 & \cellcolor{tabsecond}67.9 \\
chair (40) & & 68.1 & 69.5 & 65.9 & \cellcolor{tabsecond}79.7 & 68.7 & \cellcolor{tabfirst}80.7 & & 81.1 & 85.3 & 88.9 & \cellcolor{tabsecond}92.8 & 89.8 & \cellcolor{tabfirst}93.6 & & 85.4 & 92.0 & 96.1 & \cellcolor{tabsecond}97.2 & 96.5 & \cellcolor{tabfirst}97.7 \\
cup (40) & & 46.2 & 44.7 & 47.5 & \cellcolor{tabfirst}64.8 & 47.1 & \cellcolor{tabsecond}55.1 & & 61.4 & 65.3 & 79.1 & \cellcolor{tabfirst}86.3 & 79.5 & \cellcolor{tabsecond}81.6 & & 66.9 & 77.9 & 91.8 & \cellcolor{tabfirst}95.0 & 92.3 & \cellcolor{tabsecond}92.7 \\
donut (40) & & 49.3 & 52.8 & 39.5 & \cellcolor{tabfirst}59.1 & 43.1 & \cellcolor{tabsecond}54.2 & & 73.3 & 77.4 & 77.4 & \cellcolor{tabfirst}82.1 & 79.1 & \cellcolor{tabsecond}81.2 & & 85.0 & 88.6 & 91.9 & \cellcolor{tabfirst}93.5 & 92.6 & \cellcolor{tabsecond}93.0 \\
hairdryer (40) & & 50.9 & \cellcolor{tabsecond}55.7 & 42.0 & \cellcolor{tabfirst}61.0 & 45.7 & 54.5 & & 70.0 & 80.0 & 78.3 & \cellcolor{tabfirst}86.0 & 79.8 & \cellcolor{tabsecond}84.0 & & 77.3 & 91.6 & 92.1 & \cellcolor{tabfirst}94.8 & 92.8 & \cellcolor{tabsecond}94.3 \\
handbag (40) & & 36.4 & 40.2 & 36.3 & \cellcolor{tabfirst}49.5 & 36.2 & \cellcolor{tabsecond}45.6 & & 51.7 & 60.5 & 69.6 & \cellcolor{tabfirst}76.3 & 70.4 & \cellcolor{tabsecond}73.8 & & 60.7 & 71.8 & 86.8 & \cellcolor{tabfirst}90.6 & 87.9 & \cellcolor{tabsecond}88.6 \\
hydrant (40) & & 71.3 & 68.0 & 64.7 & \cellcolor{tabsecond}78.0 & 66.2 & \cellcolor{tabfirst}78.5 & & 85.5 & 79.4 & 88.1 & \cellcolor{tabsecond}92.1 & 88.9 & \cellcolor{tabfirst}92.8 & & 90.7 & 83.7 & 95.8 & \cellcolor{tabsecond}97.2 & 96.1 & \cellcolor{tabfirst}97.5 \\
keyboard (40) & & 39.5 & 44.0 & 42.5 & \cellcolor{tabfirst}51.1 & 42.8 & \cellcolor{tabsecond}50.8 & & 56.9 & 64.1 & 70.6 & \cellcolor{tabfirst}74.4 & 70.6 & \cellcolor{tabsecond}73.3 & & 64.8 & 75.1 & 84.1 & \cellcolor{tabfirst}86.8 & 84.0 & \cellcolor{tabsecond}84.8 \\
laptop (40) & & 56.4 & 59.6 & 49.1 & \cellcolor{tabsecond}62.0 & 53.6 & \cellcolor{tabfirst}63.4 & & 72.5 & 78.8 & 77.1 & \cellcolor{tabsecond}81.7 & 79.4 & \cellcolor{tabfirst}82.1 & & 81.5 & 88.8 & 90.6 & \cellcolor{tabfirst}92.0 & \cellcolor{tabsecond}91.7 & \cellcolor{tabsecond}91.7 \\
microwave (40) & & 45.2 & 50.9 & 46.4 & \cellcolor{tabfirst}60.2 & 47.9 & \cellcolor{tabsecond}58.1 & & 55.8 & 65.2 & 78.3 & \cellcolor{tabsecond}81.6 & 79.2 & \cellcolor{tabfirst}82.1 & & 62.1 & 72.8 & 91.9 & \cellcolor{tabsecond}92.6 & 92.4 & \cellcolor{tabfirst}93.0 \\
motorcycle (40) & & 75.2 & \cellcolor{tabsecond}81.3 & 68.7 & \cellcolor{tabfirst}82.0 & 69.2 & 81.1 & & 87.7 & 93.8 & 90.0 & \cellcolor{tabfirst}94.2 & 90.3 & \cellcolor{tabsecond}93.9 & & 91.8 & \cellcolor{tabsecond}97.8 & 96.6 & \cellcolor{tabfirst}97.9 & 96.7 & \cellcolor{tabfirst}97.9 \\
mouse (40) & & 56.6 & 55.3 & 43.5 & \cellcolor{tabfirst}61.8 & 37.9 & \cellcolor{tabsecond}60.7 & & 73.1 & 72.7 & 79.2 & \cellcolor{tabsecond}85.1 & 77.0 & \cellcolor{tabfirst}85.2 & & 80.4 & 81.0 & 92.7 & \cellcolor{tabsecond}94.4 & 92.0 & \cellcolor{tabfirst}94.8 \\
orange (40) & & 37.8 & 41.3 & 39.9 & \cellcolor{tabfirst}53.2 & 43.5 & \cellcolor{tabsecond}53.1 & & 58.2 & 66.2 & 76.1 & \cellcolor{tabsecond}81.7 & 78.5 & \cellcolor{tabfirst}81.8 & & 66.7 & 79.5 & 91.4 & \cellcolor{tabfirst}93.6 & 92.3 & \cellcolor{tabsecond}93.4 \\
parkingmeter (40) & & \cellcolor{tabsecond}66.1 & \cellcolor{tabsecond}66.1 & 59.2 & 63.7 & 57.2 & \cellcolor{tabfirst}67.6 & & 83.6 & 83.4 & \cellcolor{tabsecond}85.7 & 85.0 & 83.6 & \cellcolor{tabfirst}87.9 & & 90.5 & 90.4 & \cellcolor{tabsecond}94.9 & 94.1 & 94.2 & \cellcolor{tabfirst}95.4 \\
pizza (40) & & \cellcolor{tabsecond}47.4 & 45.3 & 29.7 & \cellcolor{tabfirst}50.8 & 32.7 & 44.8 & & \cellcolor{tabsecond}70.6 & 65.6 & 64.4 & \cellcolor{tabfirst}74.5 & 65.2 & 70.1 & & 83.3 & 75.9 & 85.1 & \cellcolor{tabfirst}90.0 & 85.0 & \cellcolor{tabsecond}86.4 \\
plant (40) & & 67.1 & 66.0 & 58.3 & \cellcolor{tabfirst}77.6 & 60.0 & \cellcolor{tabsecond}74.8 & & 80.7 & 82.8 & 86.2 & \cellcolor{tabfirst}92.8 & 86.3 & \cellcolor{tabsecond}91.4 & & 85.3 & 88.1 & 95.3 & \cellcolor{tabfirst}97.5 & 95.3 & \cellcolor{tabsecond}97.0 \\
stopsign (40) & & 48.6 & 55.5 & 47.7 & \cellcolor{tabfirst}60.9 & 41.1 & \cellcolor{tabsecond}57.2 & & 72.5 & 80.1 & 81.4 & \cellcolor{tabfirst}87.0 & 76.2 & \cellcolor{tabsecond}83.8 & & 82.7 & 88.8 & 93.5 & \cellcolor{tabfirst}95.6 & 91.8 & \cellcolor{tabsecond}94.4 \\
teddybear (40) & & 46.2 & 47.2 & 56.6 & \cellcolor{tabfirst}66.7 & 52.3 & \cellcolor{tabsecond}58.1 & & 63.4 & 65.1 & 84.7 & \cellcolor{tabfirst}88.4 & 83.6 & \cellcolor{tabsecond}85.5 & & 72.1 & 76.8 & 94.6 & \cellcolor{tabfirst}96.0 & 94.2 & \cellcolor{tabsecond}94.9 \\
toaster (40) & & 43.2 & 65.2 & 60.9 & \cellcolor{tabsecond}71.6 & 61.1 & \cellcolor{tabfirst}73.1 & & 54.8 & 81.0 & 86.8 & \cellcolor{tabsecond}90.5 & 87.2 & \cellcolor{tabfirst}91.1 & & 58.8 & 86.4 & 95.4 & \cellcolor{tabsecond}96.8 & 95.6 & \cellcolor{tabfirst}97.0 \\
toilet (40) & & \cellcolor{tabsecond}39.5 & 35.5 & 29.7 & \cellcolor{tabfirst}43.5 & 31.8 & 37.6 & & 52.3 & 54.3 & 54.7 & \cellcolor{tabfirst}62.0 & 55.7 & \cellcolor{tabsecond}57.3 & & 61.1 & 65.4 & 69.7 & \cellcolor{tabfirst}72.7 & 70.0 & \cellcolor{tabsecond}71.0 \\
toybus (40) & & 46.1 & \cellcolor{tabsecond}51.7 & 42.8 & \cellcolor{tabfirst}53.7 & 43.1 & 50.1 & & 57.9 & 70.7 & 69.8 & \cellcolor{tabfirst}72.1 & 69.4 & \cellcolor{tabsecond}72.0 & & 63.9 & 81.0 & \cellcolor{tabsecond}82.4 & \cellcolor{tabfirst}83.1 & \cellcolor{tabsecond}82.4 & \cellcolor{tabfirst}83.1 \\
toyplane (40) & & 44.6 & 43.9 & 37.3 & \cellcolor{tabfirst}53.8 & 39.4 & \cellcolor{tabsecond}49.3 & & 60.9 & 64.9 & 65.9 & \cellcolor{tabfirst}72.5 & 66.7 & \cellcolor{tabsecond}69.0 & & 69.8 & 75.5 & \cellcolor{tabsecond}80.8 & \cellcolor{tabfirst}81.9 & 79.8 & 78.9 \\
toytrain (40) & & 35.6 & 47.8 & 41.1 & \cellcolor{tabfirst}52.4 & 43.3 & \cellcolor{tabsecond}52.3 & & 47.1 & 70.2 & 70.0 & \cellcolor{tabsecond}75.1 & 73.1 & \cellcolor{tabfirst}77.8 & & 52.5 & 80.8 & 83.7 & 86.0 & \cellcolor{tabsecond}86.3 & \cellcolor{tabfirst}88.0 \\
toytruck (40) & & 31.9 & 35.8 & 26.5 & \cellcolor{tabfirst}37.3 & 26.7 & \cellcolor{tabsecond}35.9 & & 49.4 & 54.6 & 62.2 & \cellcolor{tabsecond}63.0 & 60.3 & \cellcolor{tabfirst}63.4 & & 59.0 & 65.2 & \cellcolor{tabsecond}84.2 & \cellcolor{tabfirst}85.0 & 83.3 & 83.6 \\
tv (40) & & 51.4 & 49.0 & 51.5 & \cellcolor{tabsecond}57.8 & 53.2 & \cellcolor{tabfirst}58.2 & & 67.0 & 76.1 & 81.6 & \cellcolor{tabfirst}84.0 & 82.4 & \cellcolor{tabsecond}83.6 & & 73.1 & 87.6 & 93.5 & \cellcolor{tabfirst}94.2 & \cellcolor{tabsecond}93.8 & \cellcolor{tabfirst}94.2 \\
umbrella (40) & & 68.4 & 57.6 & 62.7 & \cellcolor{tabfirst}74.4 & 65.2 & \cellcolor{tabsecond}73.3 & & 83.2 & 71.1 & 86.6 & \cellcolor{tabfirst}91.0 & 87.9 & \cellcolor{tabsecond}90.2 & & 88.6 & 79.1 & 95.2 & \cellcolor{tabfirst}97.0 & 95.7 & \cellcolor{tabsecond}96.5 \\
vase (40) & & 57.2 & 56.8 & 57.3 & \cellcolor{tabfirst}69.3 & 57.2 & \cellcolor{tabsecond}66.4 & & 71.2 & 73.8 & 83.3 & \cellcolor{tabfirst}88.7 & 84.1 & \cellcolor{tabsecond}87.1 & & 77.8 & 81.0 & 94.2 & \cellcolor{tabfirst}96.2 & 94.6 & \cellcolor{tabsecond}95.6 \\
wineglass (40) & & 43.3 & 45.9 & 50.2 & \cellcolor{tabfirst}60.3 & 55.5 & \cellcolor{tabsecond}58.2 & & 56.3 & 63.3 & 77.1 & \cellcolor{tabfirst}81.6 & 80.5 & \cellcolor{tabsecond}81.2 & & 61.2 & 70.3 & 89.0 & \cellcolor{tabsecond}90.4 & \cellcolor{tabsecond}90.4 & \cellcolor{tabfirst}90.6 \\

      \bottomrule
    \end{tabular}
    }
    \label{tbl:co3d_full3}
\end{table*}

\begin{table*}[t]
    \centering
    \caption{Per-dataset results on IMC2021~\cite{jin2021image}.}
    \vspace{-5px}
    \setlength{\tabcolsep}{3pt} 
    \resizebox{\textwidth}{!}{
    \begin{tabular}{l l c c c c c c l c c c c c c l c c c c c c
    } \toprule 
&& \multicolumn{6}{c}{AUC@3} && \multicolumn{6}{c}{AUC@5} && \multicolumn{6}{c}{AUC@10} \\

        \cmidrule{3-8} \cmidrule{10-15} \cmidrule{17-22} 
        && \tiny{SIFT} & \tiny{AL+LG} & \tiny{$\pi^3$} & \tiny{$\pi^3$+BA} & \tiny{GLUEMAP$^\dagger$} & \tiny{GLUEMAP} && \tiny{SIFT} & \tiny{AL+LG} & \tiny{$\pi^3$} & \tiny{$\pi^3$+BA} & \tiny{GLUEMAP$^\dagger$} & \tiny{GLUEMAP} && \tiny{SIFT} & \tiny{AL+LG} & \tiny{$\pi^3$} & \tiny{$\pi^3$+BA} & \tiny{GLUEMAP$^\dagger$} & \tiny{GLUEMAP} \\  \midrule
\multicolumn{22}{c}{Bag 5} \\
\midrule
british\_museum & & 31.6 & \cellcolor{tabsecond}41.5 & 40.9 & \cellcolor{tabfirst}41.9 & 40.9 & 39.7 & & 38.6 & 52.0 & \cellcolor{tabsecond}52.7 & 52.3 & \cellcolor{tabfirst}52.8 & 50.2 & & 51.0 & 66.4 & \cellcolor{tabsecond}67.8 & 67.7 & \cellcolor{tabfirst}67.9 & 65.1 \\
florence\_cathedral\_side & & 58.1 & 61.4 & 57.5 & \cellcolor{tabfirst}68.1 & 57.7 & \cellcolor{tabsecond}67.7 & & 67.5 & 71.6 & 70.0 & \cellcolor{tabfirst}77.5 & 70.1 & \cellcolor{tabsecond}77.2 & & 77.8 & 81.8 & 82.4 & \cellcolor{tabsecond}86.7 & 82.5 & \cellcolor{tabfirst}86.8 \\
lincoln\_memorial\_statue & & 40.6 & 52.5 & 61.1 & \cellcolor{tabfirst}68.8 & 61.1 & \cellcolor{tabsecond}68.1 & & 48.3 & 62.2 & 73.8 & \cellcolor{tabfirst}79.4 & 73.8 & \cellcolor{tabsecond}78.9 & & 57.4 & 71.9 & 85.6 & \cellcolor{tabfirst}89.1 & 85.6 & \cellcolor{tabsecond}88.8 \\
london\_bridge & & 34.9 & 51.2 & 42.0 & \cellcolor{tabfirst}52.3 & 41.8 & \cellcolor{tabsecond}51.8 & & 41.9 & 62.5 & 54.7 & \cellcolor{tabfirst}64.7 & 54.3 & \cellcolor{tabsecond}64.5 & & 50.6 & \cellcolor{tabsecond}75.2 & 71.9 & \cellcolor{tabfirst}78.1 & 71.6 & \cellcolor{tabfirst}78.1 \\
milan\_cathedral & & 41.3 & 41.3 & 40.2 & \cellcolor{tabsecond}43.4 & 40.1 & \cellcolor{tabfirst}46.7 & & 51.2 & 52.2 & 53.1 & \cellcolor{tabsecond}55.1 & 53.0 & \cellcolor{tabfirst}58.3 & & 63.7 & 67.3 & 69.8 & \cellcolor{tabsecond}70.5 & 69.6 & \cellcolor{tabfirst}73.2 \\
mount\_rushmore & & 30.1 & 31.9 & 32.7 & \cellcolor{tabsecond}42.9 & 32.7 & \cellcolor{tabfirst}43.8 & & 35.2 & 38.2 & 40.5 & \cellcolor{tabsecond}52.6 & 40.5 & \cellcolor{tabfirst}53.0 & & 42.6 & 48.1 & 53.8 & \cellcolor{tabfirst}66.0 & 53.7 & \cellcolor{tabsecond}65.8 \\
piazza\_san\_marco & & 27.1 & 46.7 & 52.6 & \cellcolor{tabsecond}54.2 & 53.2 & \cellcolor{tabfirst}54.3 & & 30.9 & 56.2 & 65.7 & 65.3 & \cellcolor{tabsecond}66.2 & \cellcolor{tabfirst}66.5 & & 35.7 & 67.7 & \cellcolor{tabsecond}79.8 & 78.0 & \cellcolor{tabfirst}80.1 & \cellcolor{tabsecond}79.8 \\
sagrada\_familia & & 47.4 & 55.1 & 38.8 & \cellcolor{tabfirst}60.0 & 38.6 & \cellcolor{tabsecond}58.1 & & 57.0 & 66.4 & 51.2 & \cellcolor{tabfirst}71.2 & 51.0 & \cellcolor{tabsecond}69.4 & & 67.8 & 78.5 & 66.7 & \cellcolor{tabfirst}82.9 & 66.5 & \cellcolor{tabsecond}81.5 \\
st\_pauls\_cathedral & & 45.4 & 54.2 & 50.1 & \cellcolor{tabsecond}54.3 & 50.1 & \cellcolor{tabfirst}58.6 & & 54.8 & 65.5 & 64.0 & \cellcolor{tabsecond}66.1 & 63.8 & \cellcolor{tabfirst}69.7 & & 66.1 & 78.2 & 78.9 & \cellcolor{tabsecond}79.3 & 78.8 & \cellcolor{tabfirst}81.4 \\
\midrule
\multicolumn{22}{c}{Bag 10} \\
\midrule
british\_museum & & 35.8 & \cellcolor{tabfirst}42.2 & 35.9 & \cellcolor{tabsecond}40.7 & 35.9 & \cellcolor{tabfirst}42.2 & & 47.3 & \cellcolor{tabfirst}54.8 & 49.8 & 52.8 & 49.8 & \cellcolor{tabsecond}54.2 & & 63.0 & \cellcolor{tabfirst}70.6 & 67.2 & 69.3 & 67.1 & \cellcolor{tabsecond}70.1 \\
florence\_cathedral\_side & & \cellcolor{tabsecond}72.6 & 66.7 & 50.2 & 71.6 & 50.1 & \cellcolor{tabfirst}74.7 & & \cellcolor{tabsecond}81.1 & 76.8 & 64.0 & 80.2 & 64.0 & \cellcolor{tabfirst}82.9 & & \cellcolor{tabsecond}88.9 & 85.7 & 78.4 & 87.8 & 78.4 & \cellcolor{tabfirst}90.1 \\
lincoln\_memorial\_statue & & 61.0 & 62.2 & 57.0 & \cellcolor{tabsecond}74.1 & 57.3 & \cellcolor{tabfirst}75.4 & & 72.0 & 73.2 & 71.3 & \cellcolor{tabsecond}83.0 & 71.5 & \cellcolor{tabfirst}84.2 & & 82.1 & 82.9 & 84.4 & \cellcolor{tabsecond}90.4 & 84.5 & \cellcolor{tabfirst}91.5 \\
london\_bridge & & 48.8 & \cellcolor{tabsecond}54.4 & 29.8 & 53.8 & 29.8 & \cellcolor{tabfirst}55.9 & & 58.6 & \cellcolor{tabsecond}65.9 & 44.1 & 65.5 & 44.2 & \cellcolor{tabfirst}67.7 & & 68.1 & 76.7 & 63.6 & \cellcolor{tabsecond}77.3 & 64.3 & \cellcolor{tabfirst}79.5 \\
milan\_cathedral & & \cellcolor{tabsecond}41.1 & 35.4 & 33.5 & 36.7 & 33.4 & \cellcolor{tabfirst}46.7 & & \cellcolor{tabsecond}54.5 & 48.1 & 47.6 & 49.6 & 47.5 & \cellcolor{tabfirst}60.0 & & \cellcolor{tabsecond}70.7 & 64.9 & 65.9 & 66.0 & 65.8 & \cellcolor{tabfirst}75.6 \\
mount\_rushmore & & 33.2 & 28.3 & 24.3 & \cellcolor{tabsecond}36.2 & 24.4 & \cellcolor{tabfirst}40.6 & & 42.0 & 36.7 & 32.9 & \cellcolor{tabsecond}47.1 & 32.9 & \cellcolor{tabfirst}52.2 & & 54.3 & 49.3 & 47.8 & \cellcolor{tabsecond}62.1 & 47.8 & \cellcolor{tabfirst}67.8 \\
piazza\_san\_marco & & 43.3 & 49.5 & 47.4 & \cellcolor{tabfirst}58.0 & 47.6 & \cellcolor{tabsecond}56.9 & & 52.8 & 61.3 & 62.0 & \cellcolor{tabsecond}69.8 & 62.1 & \cellcolor{tabfirst}70.4 & & 63.0 & 73.1 & 77.6 & \cellcolor{tabsecond}81.5 & 77.7 & \cellcolor{tabfirst}83.5 \\
sagrada\_familia & & 56.8 & \cellcolor{tabsecond}60.7 & 35.8 & 60.1 & 35.9 & \cellcolor{tabfirst}65.3 & & 68.8 & 71.8 & 50.2 & \cellcolor{tabsecond}72.0 & 50.4 & \cellcolor{tabfirst}76.4 & & 80.7 & 82.3 & 67.5 & \cellcolor{tabsecond}83.2 & 67.6 & \cellcolor{tabfirst}86.9 \\
st\_pauls\_cathedral & & 58.5 & \cellcolor{tabsecond}58.9 & 43.2 & 55.3 & 43.4 & \cellcolor{tabfirst}64.3 & & 70.6 & \cellcolor{tabsecond}71.3 & 58.6 & 67.5 & 58.9 & \cellcolor{tabfirst}75.7 & & 81.8 & \cellcolor{tabsecond}82.8 & 75.4 & 80.0 & 75.5 & \cellcolor{tabfirst}86.0 \\
\midrule
\multicolumn{22}{c}{Bag 25} \\
\midrule
british\_museum & & \cellcolor{tabfirst}42.8 & 41.4 & 31.3 & 40.9 & 31.3 & \cellcolor{tabsecond}41.9 & & \cellcolor{tabfirst}56.2 & 55.0 & 46.1 & 54.1 & 46.1 & \cellcolor{tabsecond}55.1 & & \cellcolor{tabfirst}72.6 & \cellcolor{tabsecond}71.6 & 64.5 & 70.8 & 64.5 & 71.1 \\
florence\_cathedral\_side & & \cellcolor{tabfirst}79.2 & 69.6 & 47.1 & 73.3 & 47.2 & \cellcolor{tabsecond}75.1 & & \cellcolor{tabfirst}86.0 & 77.3 & 61.6 & 81.7 & 61.7 & \cellcolor{tabsecond}83.1 & & \cellcolor{tabfirst}92.1 & 84.6 & 76.8 & 89.2 & 76.8 & \cellcolor{tabsecond}90.1 \\
lincoln\_memorial\_statue & & 78.5 & 58.7 & 56.3 & \cellcolor{tabsecond}80.6 & 56.5 & \cellcolor{tabfirst}82.5 & & 85.5 & 66.7 & 70.9 & \cellcolor{tabsecond}87.9 & 71.0 & \cellcolor{tabfirst}89.0 & & 91.2 & 74.5 & 84.3 & \cellcolor{tabsecond}93.8 & 84.3 & \cellcolor{tabfirst}94.4 \\
london\_bridge & & \cellcolor{tabfirst}68.0 & 59.3 & 26.9 & 51.0 & 26.7 & \cellcolor{tabsecond}62.8 & & \cellcolor{tabfirst}76.5 & 68.6 & 42.7 & 63.1 & 43.0 & \cellcolor{tabsecond}72.8 & & \cellcolor{tabfirst}84.1 & 77.1 & 63.1 & 75.4 & 64.0 & \cellcolor{tabsecond}82.0 \\
milan\_cathedral & & \cellcolor{tabsecond}49.6 & 42.7 & 27.6 & 38.8 & 27.5 & \cellcolor{tabfirst}55.2 & & \cellcolor{tabsecond}62.7 & 56.0 & 42.5 & 52.2 & 42.4 & \cellcolor{tabfirst}67.7 & & \cellcolor{tabsecond}77.1 & 71.8 & 62.4 & 69.3 & 62.3 & \cellcolor{tabfirst}81.2 \\
mount\_rushmore & & \cellcolor{tabfirst}53.4 & 31.5 & 21.0 & 43.6 & 21.0 & \cellcolor{tabsecond}52.6 & & \cellcolor{tabfirst}65.3 & 40.5 & 30.3 & 55.9 & 30.3 & \cellcolor{tabsecond}64.6 & & \cellcolor{tabfirst}78.4 & 52.9 & 46.7 & 71.0 & 46.7 & \cellcolor{tabsecond}78.1 \\
piazza\_san\_marco & & \cellcolor{tabsecond}66.4 & 56.6 & 45.2 & \cellcolor{tabfirst}67.1 & 45.2 & 58.8 & & \cellcolor{tabsecond}76.2 & 67.6 & 60.3 & \cellcolor{tabfirst}77.8 & 60.4 & 72.0 & & \cellcolor{tabsecond}85.0 & 77.9 & 76.6 & \cellcolor{tabfirst}87.7 & 76.7 & 84.4 \\
sagrada\_familia & & \cellcolor{tabfirst}72.5 & 66.7 & 34.2 & 69.6 & 34.3 & \cellcolor{tabsecond}71.8 & & \cellcolor{tabfirst}81.9 & 76.5 & 48.8 & 80.0 & 48.9 & \cellcolor{tabsecond}81.0 & & \cellcolor{tabfirst}90.1 & 85.6 & 66.8 & 89.1 & 66.9 & \cellcolor{tabsecond}89.4 \\
st\_pauls\_cathedral & & \cellcolor{tabsecond}69.0 & 65.4 & 40.0 & 55.9 & 40.3 & \cellcolor{tabfirst}71.2 & & \cellcolor{tabsecond}78.7 & 75.5 & 56.8 & 69.2 & 57.1 & \cellcolor{tabfirst}80.3 & & \cellcolor{tabsecond}87.3 & 84.6 & 74.4 & 82.2 & 74.6 & \cellcolor{tabfirst}88.5 \\
\midrule
\multicolumn{22}{c}{Bag full} \\
\midrule
british\_museum & & \cellcolor{tabfirst}54.5 & 41.3 & 29.8 & 4.4 & 29.2 & \cellcolor{tabsecond}50.5 & & \cellcolor{tabfirst}66.8 & 53.5 & 45.1 & 8.2 & 44.1 & \cellcolor{tabsecond}63.0 & & \cellcolor{tabfirst}79.9 & 69.1 & 64.2 & 17.4 & 63.1 & \cellcolor{tabsecond}76.4 \\
florence\_cathedral\_side & & \cellcolor{tabfirst}89.0 & 81.1 & 47.1 & 73.3 & 46.2 & \cellcolor{tabsecond}83.6 & & \cellcolor{tabfirst}92.9 & 86.8 & 61.3 & 80.9 & 60.9 & \cellcolor{tabsecond}89.0 & & \cellcolor{tabfirst}96.2 & 92.1 & 76.5 & 88.0 & 76.2 & \cellcolor{tabsecond}93.7 \\
lincoln\_memorial\_statue & & \cellcolor{tabfirst}85.7 & 72.4 & 56.1 & 84.6 & 55.1 & \cellcolor{tabsecond}85.1 & & 89.0 & 79.4 & 70.6 & \cellcolor{tabsecond}90.1 & 70.3 & \cellcolor{tabfirst}90.7 & & 92.2 & 86.3 & 84.0 & \cellcolor{tabsecond}94.9 & 84.0 & \cellcolor{tabfirst}95.2 \\
london\_bridge & & \cellcolor{tabsecond}74.8 & 60.9 & 18.5 & 20.6 & 35.2 & \cellcolor{tabfirst}79.3 & & \cellcolor{tabsecond}81.4 & 70.4 & 29.1 & 31.5 & 52.4 & \cellcolor{tabfirst}85.8 & & \cellcolor{tabsecond}87.2 & 80.0 & 45.6 & 48.2 & 71.6 & \cellcolor{tabfirst}91.6 \\
milan\_cathedral & & \cellcolor{tabfirst}73.9 & 56.7 & 27.7 & 47.5 & 25.6 & \cellcolor{tabsecond}68.2 & & \cellcolor{tabfirst}81.8 & 68.0 & 43.2 & 60.3 & 41.1 & \cellcolor{tabsecond}77.5 & & \cellcolor{tabfirst}89.4 & 79.7 & 63.3 & 75.0 & 61.9 & \cellcolor{tabsecond}87.0 \\
mount\_rushmore & & \cellcolor{tabfirst}76.1 & 47.3 & 18.8 & 52.5 & 18.2 & \cellcolor{tabsecond}62.8 & & \cellcolor{tabfirst}83.3 & 59.3 & 27.6 & 63.4 & 27.9 & \cellcolor{tabsecond}74.0 & & \cellcolor{tabfirst}89.6 & 74.6 & 44.0 & 77.2 & 44.7 & \cellcolor{tabsecond}84.5 \\
piazza\_san\_marco & & \cellcolor{tabfirst}77.6 & 67.0 & 46.7 & 4.5 & 47.2 & \cellcolor{tabsecond}72.5 & & \cellcolor{tabfirst}85.5 & 77.8 & 62.1 & 9.4 & 62.9 & \cellcolor{tabsecond}82.1 & & \cellcolor{tabfirst}92.4 & 87.3 & 78.1 & 21.2 & 78.8 & \cellcolor{tabsecond}90.4 \\
sagrada\_familia & & \cellcolor{tabfirst}83.2 & \cellcolor{tabsecond}77.0 & 31.9 & 73.3 & 32.6 & 76.2 & & \cellcolor{tabfirst}88.5 & \cellcolor{tabsecond}84.5 & 46.8 & 82.7 & 47.7 & 84.2 & & \cellcolor{tabfirst}93.1 & 91.2 & 65.2 & 90.6 & 66.2 & \cellcolor{tabsecond}91.3 \\
st\_pauls\_cathedral & & \cellcolor{tabsecond}77.8 & 61.6 & 40.4 & 51.9 & 40.9 & \cellcolor{tabfirst}78.5 & & \cellcolor{tabsecond}84.6 & 72.0 & 57.2 & 66.4 & 57.7 & \cellcolor{tabfirst}85.5 & & \cellcolor{tabsecond}90.7 & 82.5 & 74.8 & 80.5 & 75.1 & \cellcolor{tabfirst}91.7 \\

      \bottomrule
    \end{tabular}
    }
    \label{tbl:imc_full}
\end{table*}

\clearpage

\end{document}